\RequirePackage{fix-cm}
\documentclass[twocolumn]{svjour3}          
\smartqed  
\usepackage{graphicx}
\usepackage{multirow}
\usepackage{amssymb}
\usepackage{amsmath}
\usepackage{color}
\usepackage{enumitem}
\usepackage{threeparttable}
\usepackage[ruled, vlined]{algorithm2e}
\usepackage{subfigure}
\usepackage{gensymb} 
\usepackage{stfloats}
\usepackage{mathptmx}      
\usepackage[pagebackref=false,breaklinks=true,letterpaper=true,colorlinks,bookmarks=false,citecolor=blue]{hyperref}
\usepackage{natbib}
\usepackage{bbding}
\usepackage{float}

\newcommand{\bl}{\color{black}}
\begin{document}

\title{PageNet: Towards End-to-End Weakly Supervised Page-Level Handwritten Chinese Text Recognition}


\author{Dezhi Peng$^ 1$ \and
        Lianwen Jin$^ {1,4,5*}$ \and 
        Yuliang Liu$^ 2$ \and 
        Canjie Luo$^ 1$ \and 
        Songxuan Lai$^ 3$
}

\authorrunning{Dezhi Peng et al.} 

\institute{
	$ ^1$South China University of Technology, Guangzhou, China\\
	$ ^2$Huazhong University of Science and Technology, Wuhan, China\\
	$ ^3$Huawei Cloud Computing Technologies, Shenzhen, China\\
	$ ^4$Pazhou Laboratory (Huangpu), Guangzhou, China\\
	$ ^5$Peng Cheng Laboratory, Shenzhen, China\\
	$ ^*$Corresponding author
}

\date{Received: date / Accepted: date}

\maketitle

\begin{abstract}
Handwritten Chinese text recognition (HCTR) has been an active research topic for decades. However, most previous studies solely focus on the recognition of cropped text line images, ignoring the error caused by text line detection in real-world applications. Although some approaches aimed at page-level text recognition have been proposed in recent years, they either are limited to simple layouts or require very detailed annotations including expensive line-level and even character-level bounding boxes. To this end, we propose PageNet for end-to-end weakly supervised page-level HCTR. PageNet detects and recognizes characters and predicts the reading order between them, which is more robust and flexible when dealing with complex layouts including multi-directional and curved text lines. Utilizing the proposed weakly supervised learning framework, PageNet requires only transcripts to be annotated {\bl for real data}; however, it can still output detection and recognition results at both the character and line levels, avoiding the labor and cost of labeling bounding boxes of characters and text lines. Extensive experiments conducted on five datasets demonstrate the superiority of PageNet over existing weakly supervised and fully supervised page-level methods.
These experimental results may spark further research beyond the realms of existing methods based on connectionist temporal classification or attention. 
The source code is available at \href{https://github.com/shannanyinxiang/PageNet}{https://github.com/shannanyinxiang/PageNet}.
\end{abstract}

\section{Introduction}
\label{intro}
Handwritten Chinese text recognition (HCTR) has been studied for decades \citep{A_Graves_A_Novel,Q_Wang_Handwritten,X_Zhou_Handwritten,D_Keysers_Multi_Language,X_Zhang_Drawing}. However, most previous studies \citep{F_Yin_ICDAR13,Q_Wang_Handwritten,S_Wang_Deep,Y_Wu_Improving,D_Peng_A_Fast,T_Su_Off-line,J_Du_Deep,Z_Wang_A_Comprehensive,Z_Wang_Writer,R_Messina_Segmentation,Y_Wu_Handwritten,C_Xie_High,Y_Xiu_A_Handwritten,Z_Xie_Aggregation,Z_Wang_Weakly,Z_Zhu_Attention,C_Luo_Separating,J_Rod_Label,M_Jaderberg_Reading} assume that text line detection is provided by annotations and only focus on the recognition of cropped text line images. Although the accuracy of these line-level methods seems to be sufficient when combined with language models, they are limited to the one-dimensional distribution of characters and are significantly affected by the accuracy of text line detection in real-world applications. Therefore, handwritten text recognition at page level has important industrial value and has recently attracted remarkable research interest.
One category of page-level methods \citep{Y_Huang_Adversarial,J_Chung_A_Computationally,M_Carbonell_End-to-End,H_Yang_Recognition,Z_Xie_Weakly,W_Ma_Joint,B_Moysset_Full,C_Wigington_Start,C_Tensmeyer_Training,H_Yang_Dense,Y_Liu_Exploring,W_Feng_Residual,Z_Liu_Bottom} segments text regions from the full page and recognizes the text regions, while the others \citep{M_Yousef_OrigamiNet,T_Bluche_Joint,T_Bluche_Scan} address page-level text recognition in a segmentation-free or implicit-segmentation fashion, utilizing connectionist temporal classification (CTC) \citep{G_Alex_Connectionist} or attention mechanism combined with multi-dimensional long short-term memory. 

However, existing page-level methods have several limitations. First, most of them \citep{Y_Huang_Adversarial,J_Chung_A_Computationally,M_Carbonell_End-to-End,W_Ma_Joint,H_Yang_Dense,B_Moysset_Full} cannot be trained in a weakly supervised manner, i.e., using only line-level or page-level transcripts. Extra annotations, such as bounding boxes of text lines or characters, are necessary, but it is costly to annotate them. The methods proposed by \citep{L_Xing_Convolutional,Y_Baek_Character} can produce character bounding boxes without using corresponding annotations, but still require bounding box annotations of text lines. Some studies \citep{C_Wigington_Start,C_Tensmeyer_Training} allow a part of the training data to be annotated with only transcripts; however, expensive detection annotations are still required in the remaining training data.
Although the method proposed by \citep{Z_Xie_Weakly} can be trained under weak supervision, it is limited to a specific layout and segments pages into text lines through vertical projection. Moreover, the methods proposed by \citep{M_Yousef_OrigamiNet,T_Bluche_Joint,T_Bluche_Scan} are trained with transcripts but cannot explicitly output bounding boxes of characters or text lines. 
Second, the reading order problem, which should be a very important issue for precisely understanding sentences, has rarely been discussed in previous literature. Retrospectively, most line-level and page-level methods output recognition results from left to right. There are also page-level methods \citep{J_Chung_A_Computationally,W_Ma_Joint} that simply cluster detected words or characters into text lines based on specifically designed rules which cannot be generalized to other layouts. However, the reading order in the real world is significantly more complex, such as traditional Chinese texts that are read from top to bottom and curved text lines that are difficult to detect and recognize. 
Third, most previous approaches are not end-to-end trainable, which somewhat undermines accuracy and efficiency. Some studies \citep{J_Chung_A_Computationally,B_Moysset_Full} separately train two models to localize and recognize text lines; however, it may cause localization errors to propagate to the recognition part. The Start-Follow-Read model \citep{C_Wigington_Start} consists of three sequentially executed sub-networks, resulting in the inefficiency of the entire process. 
Finally, most previous methods are not designed for Chinese texts and thus do not perform well. Although a few approaches have been proposed for Chinese documents, they are limited to specific layouts \citep{H_Yang_Recognition,Z_Xie_Weakly,H_Yang_Dense} or require detailed annotations \citep{W_Ma_Joint}.

To address the limitations mentioned above, we propose a novel method named PageNet for end-to-end weakly supervised page-level HCTR. PageNet performs page-level text recognition from a new perspective, i.e., detecting and recognizing characters and predicting the reading order between them. Three novel components are proposed for PageNet. The detection and recognition module detects and recognizes each character on a page. The reading order module determines the linking relationship between characters and whether a character is the start/end of a line. Finally, the graph-based decoding algorithm outputs detection and recognition results at both the character and line levels. Each component is seamlessly integrated into a unified network, which makes it end-to-end trainable with high efficiency. Generally, expensive bounding box annotations of characters and text lines are required to train such a network. To this end, a novel weakly supervised learning framework, consisting of matching, updating, and optimization, is proposed to make PageNet trainable under weak supervision. Bounding box annotations are no longer required, and only line-level transcripts need to be annotated {\bl for real data}.

To the best of our knowledge, PageNet is the first method to solve page-level HCTR under weak supervision. Although no bounding box annotation is provided {\bl for real data}, our model can still produce rich information that contains detection and recognition results at both the character and line levels. Therefore, our method can avoid the high cost of annotating the bounding boxes. Moreover, weakly annotated data is easy to obtain from the Internet, which makes data collection almost free. A comparison of the required annotations versus the model output is presented in Table \ref{Tab_Intro_Ano_Out}. Compared with existing page-level methods, our method requires fewer annotations but outputs more information. To the best of our knowledge, PageNet is also the first method to solve the reading order problem in page-level HCTR. The reading order problem involves determining the order in which characters are read. By utilizing the proposed reading order module and graph-based decoding algorithm, our model can handle arbitrarily curved and multi-directional texts. In addition, although the model is designed for Chinese texts, it can also process multilingual texts including Chinese and English. 

\begin{table}[tb]
	\centering
	\tiny
	\caption{Comparison of the required annotations versus the model output of existing page-level methods (L: line-level; W: word-level; C: character-level)}
	\label{Tab_Intro_Ano_Out}
	\resizebox{\hsize}{!}{
	\begin{tabular}{|c|c|c|c|c|c|c|}
		\hline
		\multirow{3}*{Method} & \multicolumn{3}{|c|}{Annotations} & \multicolumn{3}{|c|}{Outputs} \\
		\cline{2-7}
		~ & \multicolumn{2}{|c|}{Detection} & \multirow{2}*{Transcript} & \multicolumn{2}{|c|}{Detection} & \multirow{2}*{Transcript} \\
		\cline{2-3}
		\cline{5-6}
		~ & L & W or C & ~ & L & W or C & ~ \\
		\hline
		\citet{T_Bluche_Joint} & \multicolumn{1}{c}{} & \multicolumn{1}{c}{} & \Checkmark & \multicolumn{1}{c}{} & \multicolumn{1}{c}{} & \Checkmark \\
		\citet{M_Yousef_OrigamiNet} & \multicolumn{1}{c}{} & \multicolumn{1}{c}{} & \Checkmark & \multicolumn{1}{c}{} & \multicolumn{1}{c}{} & \Checkmark \\
		\citet{C_Wigington_Start} & \multicolumn{1}{c}{} & \multicolumn{1}{c}{} & \Checkmark & \multicolumn{1}{c}{\Checkmark} & \multicolumn{1}{c}{} & \Checkmark \\
		\citet{Y_Huang_Adversarial} & \multicolumn{1}{c}{\Checkmark} & \multicolumn{1}{c}{} & \Checkmark & \multicolumn{1}{c}{\Checkmark} & \multicolumn{1}{c}{} &  \Checkmark \\
		\citet{W_Ma_Joint} & \multicolumn{1}{c}{\Checkmark} & \multicolumn{1}{c}{\Checkmark} & \Checkmark & \multicolumn{1}{c}{\Checkmark} & \multicolumn{1}{c}{\Checkmark} & \Checkmark \\
		\textbf{Ours} & \multicolumn{1}{c}{} & \multicolumn{1}{c}{} & {\color{red} \CheckmarkBold} & \multicolumn{1}{c}{\color{red} \CheckmarkBold} & \multicolumn{1}{c}{\color{red} \CheckmarkBold} & {\color{red} \CheckmarkBold} \\
		\hline
	\end{tabular}}
\end{table}

To verify the effectiveness of our method, extensive experiments are conducted on five datasets, namely CASIA-HWDB \citep{C_Liu_CASIA}, ICDAR2013 \citep{F_Yin_ICDAR13}, MTHv2 \citep{W_Ma_Joint}, SCUT-HCCDoc \citep{H_Zhang_SCUT-HCCDoc}, and JS-SCUT PrintCC. Because our model is weakly supervised, we further propose two evaluation metrics, termed accurate rate* (AR*) and correct rate* (CR*), for the situation in which only line-level transcripts are given. The experimental results show that PageNet outperforms other weakly supervised page-level methods. Compared with fully supervised approaches, PageNet can also achieve competitive or better performance. Moreover, PageNet performs better than existing line-level methods for HCTR that directly recognize cropped text line images. 

In summary, the main contributions of this paper are: 

\begin{itemize}[topsep=0pt,partopsep=0pt]
	\item We propose a novel method named PageNet for end-to-end weakly supervised page-level HCTR. PageNet solves page-level text recognition from a new perspective, namely, detecting and recognizing characters and predicting the reading order.
	\item A novel weakly supervised learning framework, consisting of matching, updating, and optimization, is proposed to make PageNet trainable {\bl with only line-level transcripts annotated for real data.} Nevertheless, it can output detection and recognition results at both the character and line levels. Therefore, the cost of manual annotation can be significantly reduced.
	\item To the best of our knowledge, PageNet is the first method to address the reading order problem in page-level HCTR. The model can handle pages with multi-directional reading order and arbitrarily curved text lines.
	\item Extensive experiments on five benchmarks demonstrate the superiority of PageNet, indicating that it may be a remarkable step towards a new effective approach to the page-level HCTR problem.
\end{itemize}
\section{Related Work}
\label{Sec_Related}
\subsection{Line-level Handwritten Chinese Text Recognition}
The methods for line-level HCTR aim to recognize text line images, which can be divided into two categories: segmentation-based and segmentation-free methods.

Segmentation-based methods address this problem based on oversegmentation or deep detection networks. The strategy using oversegmentation first obtains consecutive oversegments and then searches for the optimal segmentation-recognition path by integrating classifier outputs, geometric context, and linguistic context \citep{Q_Wang_Handwritten}. \citet{S_Wang_Deep} improved the oversegmentation method using deep knowledge training and heterogeneous convolutional neural networks.
Furthermore, based on oversegmentation methods, \citet{Y_Wu_Improving} explored neural network language models and \citet{Z_Wang_Weakly} proposed a weakly supervised learning method.
However, it is difficult for these methods to recognize touching and overlapping characters. Therefore, with the prevalence of deep detection networks, \citet{D_Peng_A_Fast} proposed a segmentation and recognition module to detect and recognize characters in an end-to-end manner. 

In addition, there are methods that solve line-level HCTR from a segmentation-free perspective. The methods proposed by \citep{T_Su_Off-line,J_Du_Deep,Z_Wang_A_Comprehensive} adopted systems based on hidden Markov model. \citet{Z_Wang_Writer} further introduced writer adaptation to this type of approach. Combining long short-term memory recurrent neural network (LSTM-RNN) and CTC \citep{G_Alex_Connectionist} is another framework. \citet{R_Messina_Segmentation} used multi-dimensional LSTM-RNN to resolve line-level HCTR. \citet{Y_Wu_Handwritten} proposed a separable multi-dimensional LSTM-RNN and achieved a significant improvement compared with previous LSTM-RNN-based methods. In addition to the methods focusing on the network architecture, \citet{C_Xie_High} explored data preprocessing and augmentation pipelines and achieved state-of-the-art results. The attention mechanism can also be used for line-level HCTR. \citet{Y_Xiu_A_Handwritten} explored the attention-based decoder and proposed a multi-level multimodal fusion network to incorporate both the visual and linguistic semantic information. 

Furthermore, to utilize both segmentation-based and segmentation-free methods, \citet{Z_Zhu_Attention} proposed to combine these two kinds of approaches using a convolutional combination strategy.

In contrast to these line-level methods, the proposed PageNet model recognizes the text directly from the full page in an end-to-end fashion.
\subsection{Page-level Handwritten Text Recognition}
\label{sec_related_work_pageHTR}
The goal of page-level handwritten text recognition is to recognize handwritten text from the full page. One category of methods detects text regions and then recognizes them. \citet{J_Chung_A_Computationally} developed two separate components for text localization and recognition. \citet{M_Carbonell_End-to-End} proposed an end-to-end text detection and transcription framework wherein the two components are jointly trained. \citet{Y_Huang_Adversarial} further improved the end-to-end framework using an adversarial feature enhancing network. \citet{B_Moysset_Full} proposed to regress the left-side triplets rather than the coordinates of bounding boxes and determine the end of a line using a recognizer with an extra end-of-line label. Some methods for scene text spotting, such as Mask TextSpotter \citep{P_Lyu_Mask,M_L_Mask} and FOTS \citep{X_Liu_FOTS}, can also be applied to page-level handwritten text recognition. For Chinese text, \citet{W_Ma_Joint} presented a historical document processing system that simultaneously performs layout analysis, character detection, and character recognition. \citet{H_Yang_Dense} proposed a recognition-guided detector for tight Chinese character detection in historical documents.\\
\indent However, detection annotations, such as bounding boxes of text lines or characters, must be provided for the aforementioned methods. Therefore, some studies have focused on weakly supervised page-level handwritten text recognition that requires only transcripts for model training. 
\citet{Z_Xie_Weakly} proposed a method for weakly supervised character detection in historical documents. However, this method is limited to a specific layout and cannot be generalized to unconstrained situations. \citet{C_Wigington_Start} proposed the Start-Follow-Read model that requires only a small proportion of data to be fully annotated and the remaining data to be weakly annotated. \citet{C_Tensmeyer_Training} further designed a novel alignment algorithm and enabled methods such as Start-Follow-Read to be trained using transcripts without line breaks. Combining multi-dimensional LSTM-RNN with the attention mechanism is another way to solve weakly supervised page-level handwritten text recognition. Following this idea, \citet{T_Bluche_Scan} and \citet{T_Bluche_Joint} proposed methods for transcribing paragraphs. Furthermore, OrigamiNet \citep{M_Yousef_OrigamiNet} demonstrates that CTC can also be used for page-level text recognition by implicitly unfolding the 2-dimensional input signal to 1-dimensional.\\
\indent Compared with existing methods, the proposed PageNet is trained without bounding box annotations {\bl for real data} but outputs detection and recognition results at both the character and line levels. PageNet is also the first method to solve the reading order problem in page-level HCTR, which makes the model more robust and flexible.

\begin{figure*}[ht]
	\centering
	\includegraphics[width=2.0\columnwidth]{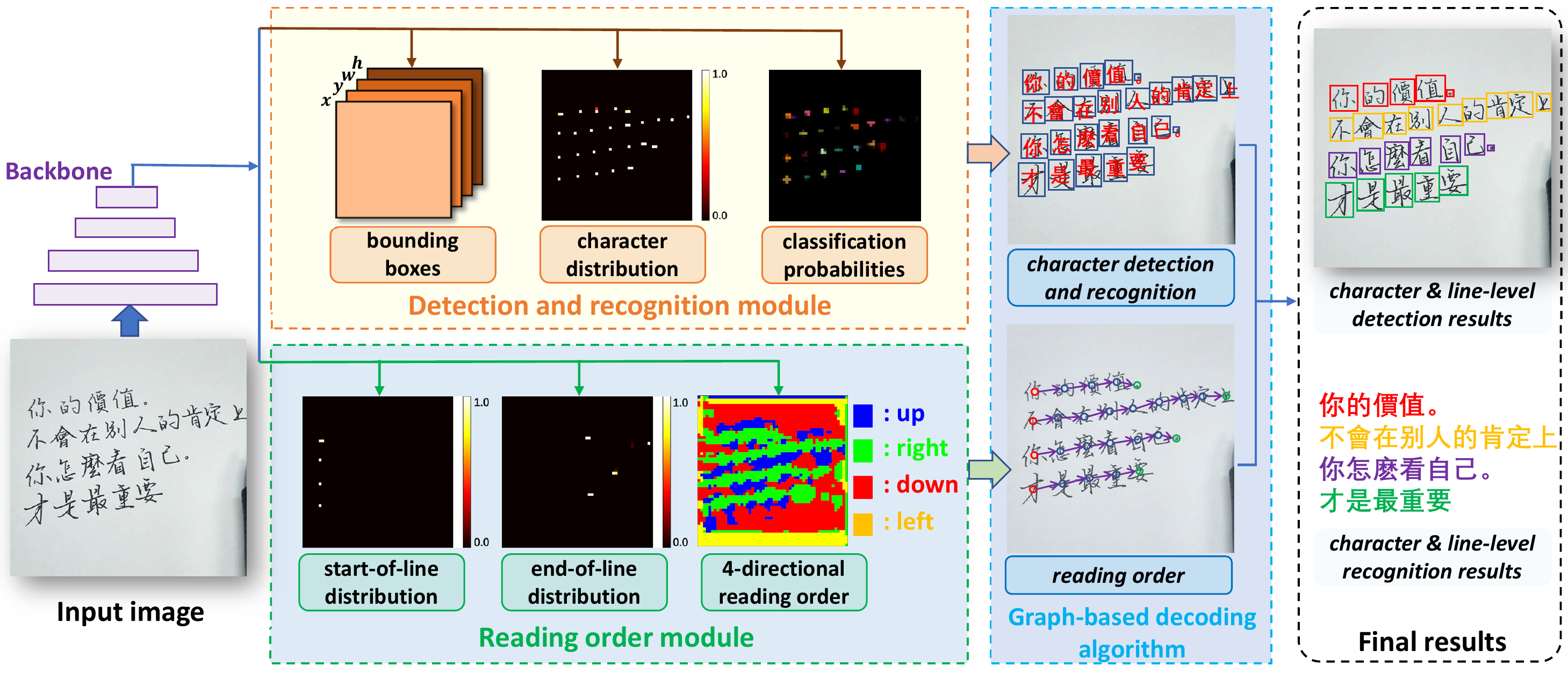}
	{\bl \caption{Overall architecture of PageNet. Based on the features extracted by the backbone network, the detection and recognition module predicts the character detection and recognition results, while the reading order module predicts the reading order between characters. Combining these two predictions, the graph-based decoding algorithm outputs the final results containing detection and recognition results at both the character and line levels.}\label{Fig_Model_Overall}}	
\end{figure*}

\section{Methodology}
\label{Sec_Model}
Most existing methods solve page-level text recognition following a top-down pipeline, i.e., text line detection and recognition. However, curved text lines have become a major challenge for such methods, and the reading order problem has rarely been investigated. Moreover, unlike other languages, Chinese characters are the basic elements that directly form sentences. Therefore, following a bottom-up pipeline, we propose PageNet for end-to-end weakly supervised page-level HCTR.
	
PageNet performs page-level text recognition from a new perspective, i.e., detecting and recognizing characters and predicting the reading order between them, which enables it to handle pages with multi-directional reading order and arbitrarily curved text lines. As shown in Fig. \ref{Fig_Model_Overall}, PageNet consists of four parts: (1) the backbone network for feature extraction, (2) the detection and recognition module for character detection and recognition, (3) the reading order module for predicting the reading order between characters, and (4) the graph-based decoding algorithm that outputs the final results containing detection and recognition results at both the character and line levels. 
The detailed network architecture of PageNet is shown in Fig. \ref{Fig_Exp_Arch}.
{\bl The components for character detection, character recognition, and reading order are integrated into a single network that is end-to-end optimized.}
	
Manual annotations, including expensive line-level and character-level bounding boxes, are required by most previous methods. To this end, a novel weakly supervised learning framework (Fig. \ref{Fig_WSL_Overview}) is proposed to make PageNet trainable with only line-level transcripts annotated {\bl for real data}, thereby avoiding the labor and cost of labeling bounding boxes of characters and text lines.

\subsection{Backbone Network}
Given an image with height $H$ and width $W$, the backbone network extracts high-level feature maps of shape $\frac{W}{16} \times \frac{H}{16} \times 512$. In the following, we denote $\frac{W}{16}$ as $W_g$ and $\frac{H}{16}$ as $H_g$ for convenience.

\subsection{Detection and Recognition Module}
\label{Sec_DRM}

\begin{figure}[t]
	\centering 
	\includegraphics[width=1.0\columnwidth]{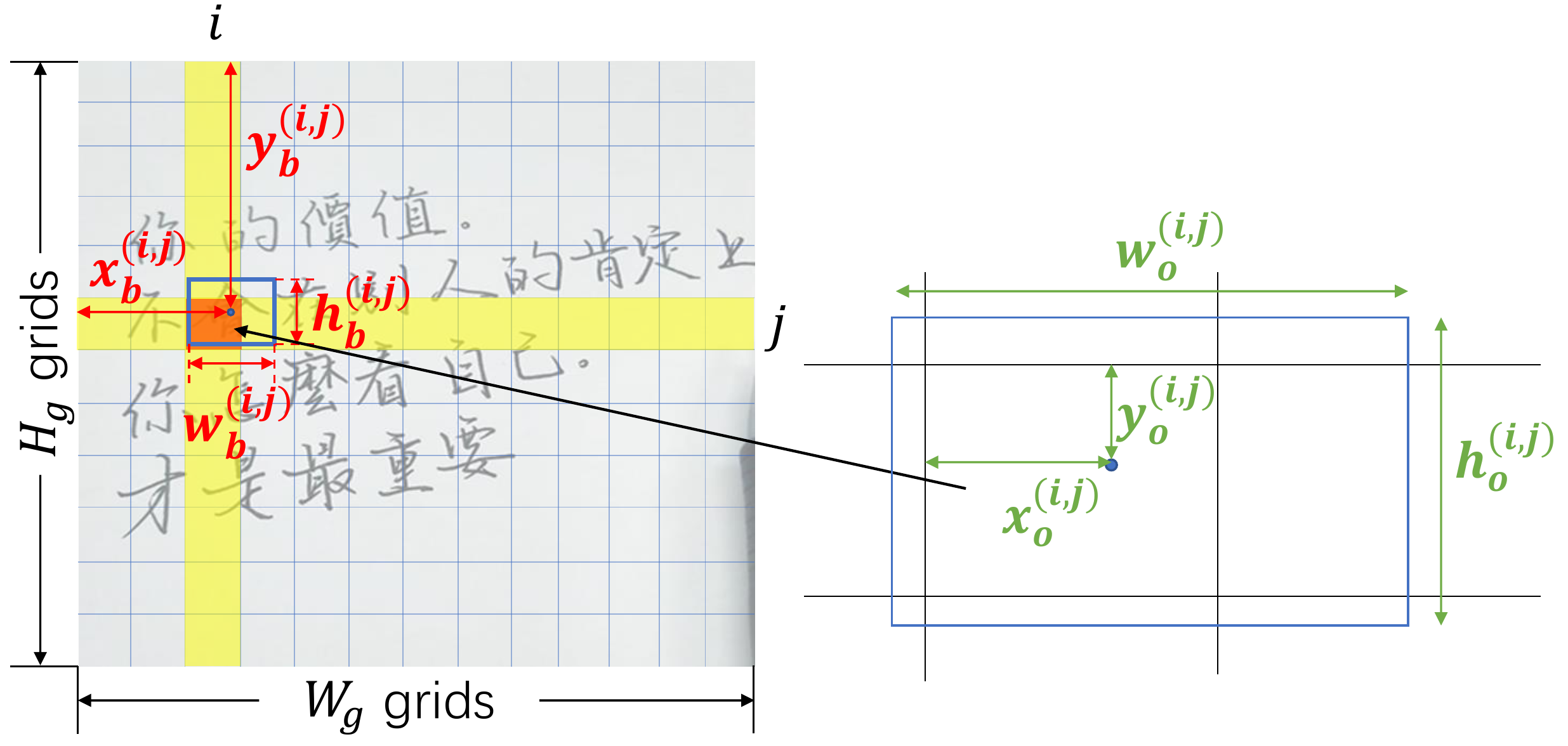}
	\caption{Relationship between the output $O_{{\bl box}}$ of {\bl CharBox branch} and the coordinates $B_{{\bl box}}$ of bounding boxes.}
	\label{Fig_Model_SRM_Mechanism}
\end{figure}

Following the successful decoupled three-branch design of our previous work \citep{D_Peng_A_Fast}, the detection and recognition module is proposed for character detection and recognition, which consists of {\bl character bounding box (CharBox), character distribution (CharDis), and character classification (CharCls) branches.}
We first apply $W_g \times H_g$ grids to the input image, as shown in the left part of Fig. \ref{Fig_Model_SRM_Mechanism}, and denote the grid at the $i$-th column and $j$-th row as $G^{(i,j)}$. Then, the function of each branch is as follows:

\noindent\textbf{\bl CharBox Branch} outputs $O_{{\bl box}}$ of shape $W_g \times H_g \times 4$. Fig. \ref{Fig_Model_SRM_Mechanism} and Eq. (\ref{Equ_final_bb}) show the relationship between $O_{{\bl box}}^{(i,j)}=({x}_o^{(i,j)}, {y}_o^{(i,j)}, {w}_o^{(i,j)}, {h}_o^{(i,j)})$ and the coordinate $B_{{\bl box}}^{(i,j)}=({x}_b^{(i,j)}, {y}_b^{(i,j)}, {w}_b^{(i,j)}, {h}_b^{(i,j)})$ of the bounding box for grid $G^{(i,j)}$.
\begin{equation}
	\bl
	\small
	\label{Equ_final_bb}
	\begin{aligned}
		x_b^{(i,j)} & = & (i - 1 + {x}_o^{(i,j)}) / {W_g} \times W,\\
		y_b^{(i,j)} & = & (j - 1 + {y}_o^{(i,j)}) / {H_g} \times H,\\
		w_b^{(i,j)} & = & {w}_o^{(i,j)}, \\
		h_b^{(i,j)} & = & {h}_o^{(i,j)}. \\
	\end{aligned}
\end{equation}

\noindent\textbf{\bl CharDis Branch} produces character distribution $O_{\bl dis}$ of shape $W_g \times H_g$, where $O_{{\bl dis}}^{(i,j)}$ is the confidence that grid $G^{(i,j)}$ contains characters.
	
\noindent\textbf{\bl CharCls Branch} generates $O_{cls}$ of shape $W_g \times H_g \times N_{cls}$, where $O_{cls}^{(i,j)}$ contains the classification probabilities of $N_{cls}$ categories for grid $G^{(i,j)}$.

\begin{figure}[b]
	\centering
	\includegraphics[width=1.02\columnwidth]{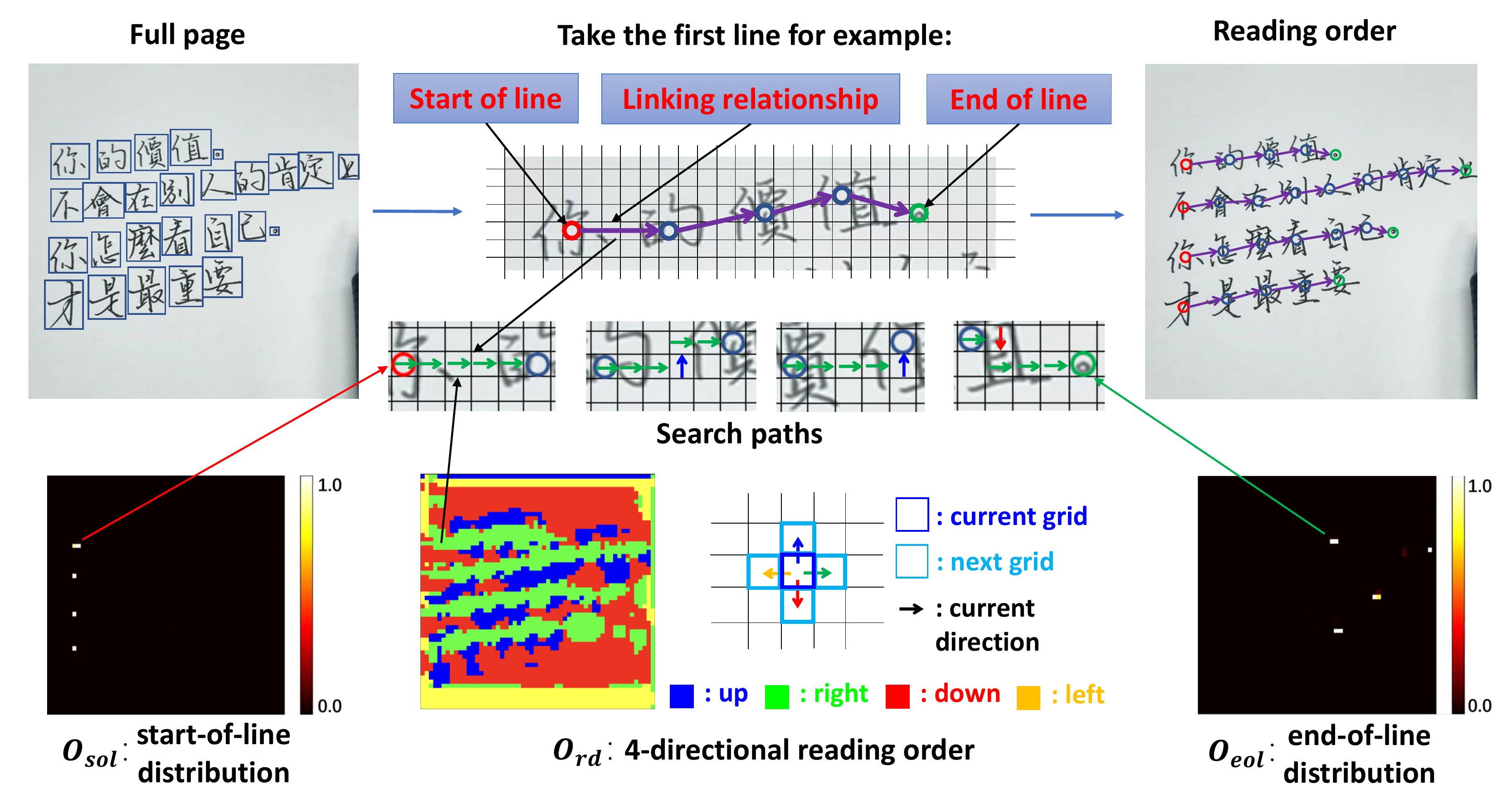}
	\caption{Reading order problem is solved by making three predictions: (1) $O_{sol}$: start-of-line distribution, (2) $O_{rd}$: 4-directional reading order prediction, and (3) $O_{eol}$: end-of-line distribution.}
	\label{Fig_Model_ROM}
\end{figure}

\subsection{Reading Order Module}

For a line-level recognizer, it is natural to arrange the recognized characters from left to right. However, the situation becomes significantly more complicated when characters can be arbitrarily distributed along two dimensions. The reading order problem has rarely been studied in previous literature, especially on the task of HCTR. However, this problem is important for building a flexible and robust page-level recognizer. Therefore, we propose the reading order module to solve this problem.

\subsubsection{Problem Definition}
Given an unordered set of characters, the reading order problem is to determine the order in which characters are read. We only investigate the reading order at line level rather than page level by rearranging the characters into multiple line-level transcripts. This means that the characters in one line-level transcript are sorted according to the reading order, but we do not consider the page-level reading order between different line-level transcripts. When a page contains simple layouts, such as only one paragraph, the page-level reading order can be easily determined using location information. However, when a page contains complex layouts, it is usually difficult to determine the page-level reading order. Different people may read the text lines in different orders. Moreover, most datasets only provide the transcript of each line separately.

\subsubsection{Our Solution}
\label{Model_ROM_ProDe}
Most previous methods simply solve the reading order problem by detecting text lines and recognizing them from left to right \citep{B_Moysset_Full,Y_Huang_Adversarial,P_Lyu_Mask,X_Liu_FOTS}. However, these methods have difficulty handling multi-directional and curved text lines. To solve these issues, as illustrated in Fig. \ref{Fig_Model_ROM}, we decompose the reading order problem into three steps: (1) starting at the start-of-line, (2) finding the next character according to the linking relationship between characters, and (3) stopping at the end-of-line. The linking relationship is further decomposed into the movements from the current grid to its neighbor step by step. This pipeline makes it possible to deal with text lines with arbitrary directions and curves. 

Specifically, we solve the reading order problem by making three predictions, namely the start-of-line distribution $O_{sol}$, the 4-directional reading order prediction $O_{rd}$, and the end-of-line distribution $O_{eol}$. The detailed network architecture of the reading order module is shown in Fig. \ref{Fig_Exp_Arch}(c). Both $O_{sol}$ and $O_{eol}$ are of shape $W_g \times H_g$, where $O_{sol}^{(i,j)}$ and $O_{eol}^{(i,j)}$ are the confidence that the character in grid $G^{(i,j)}$ is the start-of-line and the end-of-line, respectively.
The 4-directional reading order prediction $O_{rd}$ is of shape $W_g \times H_g \times 4$, where $O_{rd}^{(i,j)}$ are the probabilities of the four directions for grid $G^{(i,j)}$. The four predefined directions are up, right, down, and left, respectively. If the direction of grid $G^{(i,j)}$ is right, then the next grid is on its right, i.e., the next grid is grid $G^{(i+1,j)}$. The other three directions can be defined similarly. Thus, from a character, we can find the next character in the reading order by iteratively moving from a grid to the next according to the direction with maximum probability until arriving at a new character, as illustrated in the visualization of search paths in Fig. \ref{Fig_Model_ROM}.

\begin{figure}[b]
	\centering 
	\includegraphics[width=1.0\columnwidth]{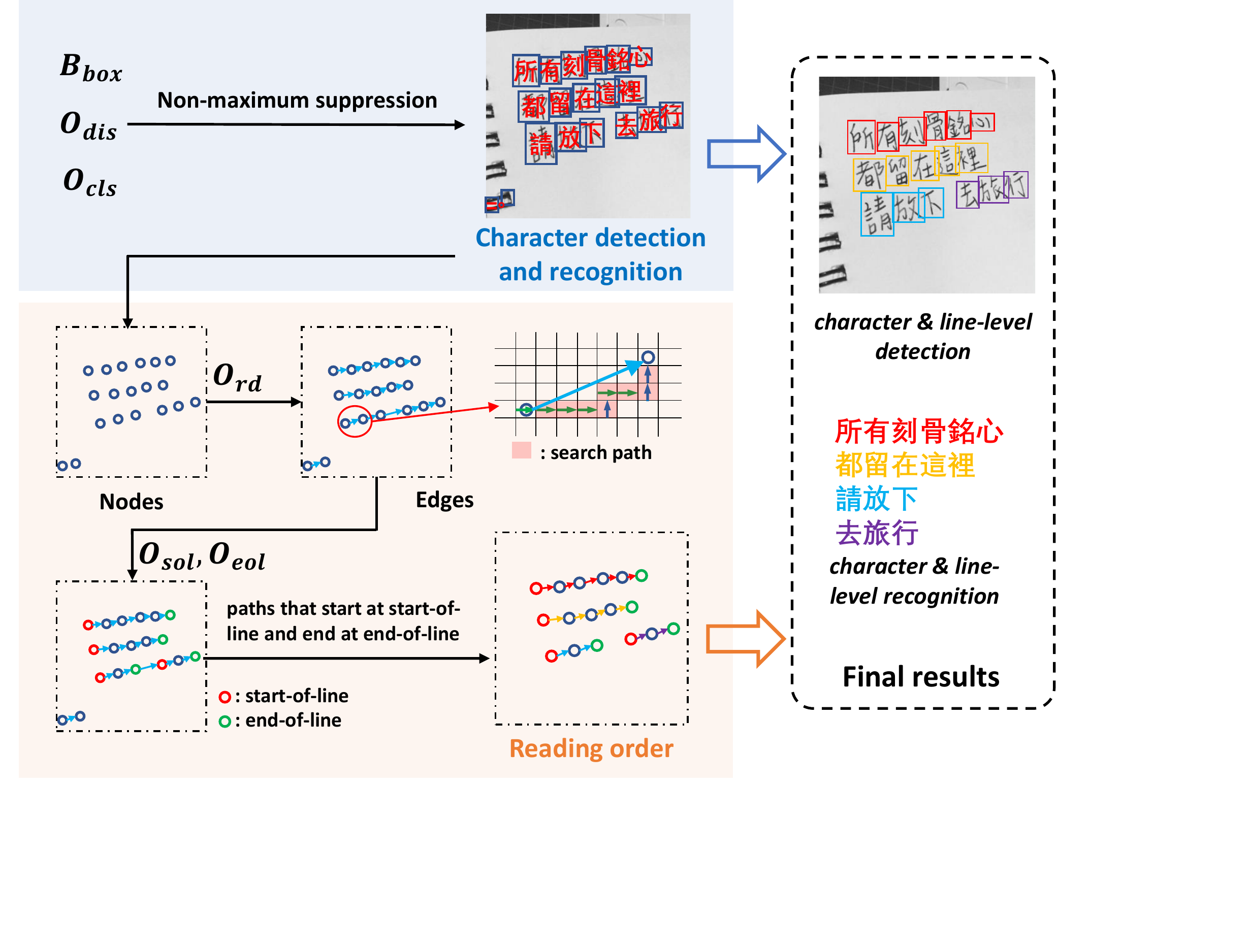}
	\caption{Pipeline of the graph-based decoding algorithm. Based on the outputs from the detection and recognition module and the reading order module, the graph-based decoding algorithm produces the final results.}
	\label{Fig_Model_GDA}
\end{figure}

\subsection{Graph-based Decoding Algorithm}
\label{Sec_GDA}

Based on the predictions from the detection and recognition module and the reading order module, we propose a novel graph-based decoding algorithm to produce the final output containing detection and recognition results at both the character and line levels by viewing characters and reading order as a graph. 
	
As shown in Fig. \ref{Fig_Model_GDA}, the graph-based decoding algorithm consists of the following three steps: (1) the character detection and recognition results are derived from the outputs of the detection and recognition module, (2) the reading order is generated based on the outputs of the reading order module, and (3) the final results, which contain detection and recognition results at both the character and line levels, are obtained by combining the reading order and the character detection and recognition results.

\subsubsection{Character Detection and Recognition}
\label{sec_char_seg_rec}
The detection and recognition module predicts the coordinates $B_{{\bl box}}$ of bounding boxes, the character distribution $O_{{\bl dis}}$, and the classification probabilities $O_{cls}$. We use non-maximum suppression (NMS) \citep{A_Neubeck_Efficient} to remove redundant bounding boxes and obtain the character detection and recognition results, as shown in the blue part of Fig. \ref{Fig_Model_GDA}. The character detection and recognition results contain multiple characters with their bounding boxes and categories.

\subsubsection{Reading Order}
\label{Model_GDA_RO}

The pipeline for generating the reading order is illustrated in the orange part of Fig. \ref{Fig_Model_GDA}. The three steps are as follows.
	
\noindent\textbf{Nodes.} Each character detection and recognition result is viewed as a node. Therefore, each node corresponds to a grid in which the bounding box and category of the related character are predicted.
	
\noindent\textbf{Edges.} Based on the 4-directional reading order prediction $O_{rd}$, we find the next node of every node. Starting at the corresponding grid of one node, we move into the neighboring grid step by step according to the direction with the maximum probability in $O_{rd}$. If a grid with a corresponding node is reached, the next node is successfully found. However, if the search path exceeds the boundary of the grids or is stuck in a cycle, the next node does not exist. 
	
\noindent\textbf{Reading Order.} We distinguish whether a node is the start-of-line or the end-of-line according to the start-of-line distribution $O_{sol}$ and the end-of-line distribution $O_{eol}$. Then, the reading order is represented by the paths that start at the start-of-line and end at the end-of-line.

\subsubsection{Final Results}
\label{sec_final_results}
In the reading order, each path represents a text line, and each node corresponds to a character. After reorganizing the character detection and recognition results according to the reading order, the final results are obtained as:
\begin{equation}
	\label{Equ_final_results}
	R=\{R^{(p)} | 1 \leq p \leq N_{ln}\},
\end{equation}
where $N_{ln}$ is the number of text lines. $R^{(p)}$ contains the categories and bounding boxes of the characters in the $p$-th line, as shown in Eq. (\ref{Equ_R_l}).
\begin{equation}
	\label{Equ_R_l}
	R^{(p)}\!=\!\{(c^{(p,m)}\!,x^{(p,m)}\!, y^{(p,m)}\!, w^{(p,m)}\!, h^{(p,m)}) | 1\! \leq\! m\! \leq\! N_{ch}^{(p)}\},
\end{equation}
where $N_{ch}^{(p)}$ is the number of characters in the $p$-th line. $c^{(p,m)}$ and $(x^{(p,m)}, y^{(p,m)}, w^{(p,m)}, h^{(p,m)})$ are the category and bounding box of the $m$-th character in the $p$-th line, respectively, which are defined as:
\begin{gather}
	c^{(p,m)} = \mathop{\arg\max}\nolimits_{1 \leq c \leq N_{cls}} O_{cls}^{(\alpha^{(p, m)},\beta^{(p,m)},c)}, \\
	(x^{(p,m)}, y^{(p,m)}, w^{(p,m)}, h^{(p,m)}) = B_{{\bl box}}^{(\alpha^{(p,m)}, \beta^{(p,m)})},
\end{gather}
where $(\alpha^{(p,m)}, \beta^{(p,m)})$ is the coordinate of the grid corresponding to the $m$-th node of the $p$-th path.

\begin{figure*}[t]
	\centering
	\includegraphics[width=2.0\columnwidth]{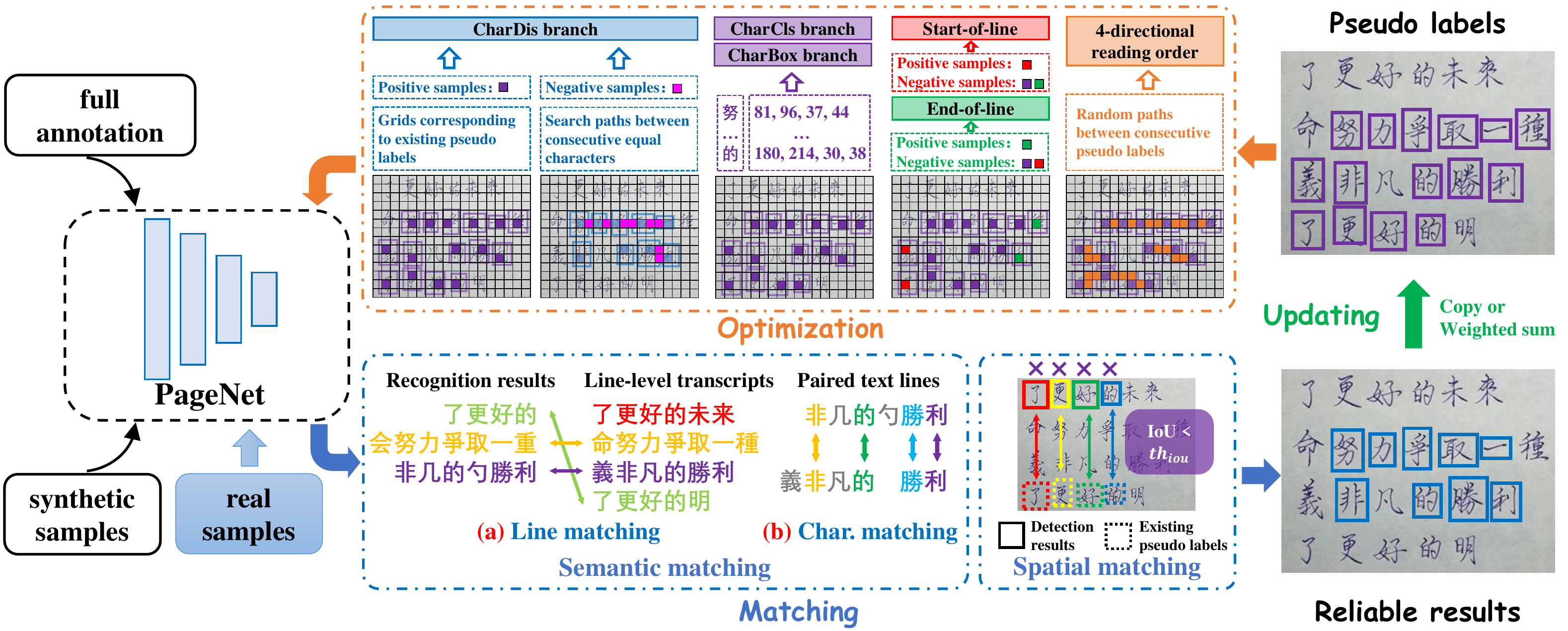}
	{\bl \caption{Overall framework of weakly supervised learning.}\label{Fig_WSL_Overview}}
\end{figure*}

\subsection{Weakly Supervised Learning}
\label{Sec_Weakly}
Normally, training PageNet requires full annotations, including the bounding boxes and categories of characters. However, the cost of annotating the bounding box and category of each character on a page is significantly higher than only annotating the transcript of each line. Moreover, line-level transcripts can be obtained almost without cost if the pages are from published books or historical documents.
There is also a large amount of weakly annotated data on the Internet that has not been made full use of. 
Therefore, in this section, we present a weakly supervised learning framework, which consists of matching, updating, and optimization, to make PageNet trainable with only line-level transcripts {\bl annotated for real data.} The character-level and line-level bounding boxes not only no longer need to be labeled, but can even be automatically annotated through the proposed weakly supervised learning framework.

\subsubsection{Overview}
\label{sec_wsl_overview}
The overall framework of weakly supervised learning is shown in Fig. \ref{Fig_WSL_Overview}. The training data consists of synthetic and real samples. As described in Sec. \ref{Sec_Dataset}, the synthetic samples have full annotations; therefore, the model can be normally optimized. However, the annotations of the real samples only contain line-level transcripts. Thus, three steps are designed for real samples. (1) \textit{Matching:} match the results of PageNet with the line-level transcripts in the annotations to find reliable results. (2) \textit{Updating:} Use the reliable results to update pseudo-labels. The pseudo-labels are the bounding boxes of the characters in the transcript annotations. (3) \textit{Optimization:} Calculate the losses using the updated pseudo-labels to optimize the parameters. Because not all the characters have corresponding pseudo-labels, it is challenging to effectively train the model in such a situation.

\subsubsection{Definition of Symbols}
\label{sec_def_symbol}
For convenience, the symbols are defined as follows.
\begin{itemize}[itemsep=0pt, topsep=0pt, leftmargin=12pt]
	\item Given a real image, PageNet predicts the results $R$ (Eq. (\ref{Equ_final_results})){\bl , where $c^{(p, m)}$ and $(x^{(p,m)}, y^{(p,m)}, w^{(p,m)}, h^{(p,m)})$ are the category and bounding box of the $m$-th character in the $p$-th line, respectively, as specified in Sec. \ref{sec_final_results}.} Based on the predicted results, we further define the recognition result of the $p$-th line as $L^{(p)} = \{c^{(p,m)} | 1 \leq m \leq N_{ch}^{(p)}\}$.
	\item $B_{sco}^{(p,m)}$ is defined as the score of the bounding box of the $m$-th character in the $p$-th line .
	\item $P_{sch}^{(p,m)}$ is defined as the coordinates of the grids in the search path that starts from the $m$-th character of the $p$-th line.
	\item The line-level transcript annotations are denoted as $A = \{A^{(q)}| 1 \leq q \leq \hat{N}_{ln}\}$, where $A^{(q)} = \{c_{gt}^{(q,n)} | 1 \leq n \leq \hat{N}_{ch}^{(q)}\}$ is the transcript of the $q$-th line and $\hat{N}_{ln}$ is the total number of lines. Moreover, $c_{gt}^{(q,n)}$ is the category of the $n$-th character in the $q$-th line, and $\hat{N}_{ch}^{(q)}$ is the number of characters in the $q$-th line.
	\item The pseudo-label of character $c_{gt}^{(q,n)}$ is the coordinate of its bounding box $A_{ps}^{(q,n)} = (x_{ps}^{(q,n)}, y_{ps}^{(q,n)}, w_{ps}^{(q,n)}, h_{ps}^{(q,n)})$. We further denote the score of $A_{ps}^{(q,n)}$ as $\gamma^{(q,n)}$.
\end{itemize}

\subsubsection{Matching}
\label{Sec_Matching}
Given the results $R$ and the line-level transcripts $A$, the aim of matching is to find reliable character-level results and their corresponding ground-truth characters. Specifically, the matching algorithm consists of semantic matching and spatial matching.
	
\noindent\textbf{Semantic Matching.} In general, correctly recognized characters also have accurate bounding boxes. Based on this observation, semantic matching aims to identify correctly recognized characters in the results. As shown in Fig. \ref{Fig_WSL_Overview}, semantic matching is composed of two steps as follows. 
\begin{algorithm}[tb]
	\footnotesize
	\caption{Line Matching}
	\label{Alg_Line_Matching}
	\LinesNumbered
	\KwIn{{recognition results $L$}, transcripts $A$}
	\KwOut{line matches $M_l$}
	\For{$p=1$ to $N_{ln}$ and $q=1$ to $\hat{N}_{ln}$}{
		$M_{AR}^{(p, q)} \leftarrow$ AR between $L^{(p)}$ and $A^{(q)}$\;
	}
	Sort $M_{AR}$ from large to small, yielding the sorted indices $S_{AR} = \{(i^{(k)}, j^{(k)}) | M_{AR}^{(i^{(k+1)}, j^{(k+1)})} \leq M_{AR}^{(i^{(k)}, j^{(k)})}\}$\;
	\For{$k=1$ to $N_{ln} \times \hat{N}_{ln}$}{
		\If{$M_{AR}^{(i^{(k)}, j^{(k)})} \geq th_{AR}$}{
			\If{both $L^{(i^{(k)})}$ and $A^{(j^{(k)})}$ are not matched}{
				add $(i^{(k)}, j^{(k)})$ to $M_l$\;
			}
		}
	}
\end{algorithm}

\noindent(1) \textit{Line matching:} We match the line-level transcripts $A$ and the recognition results $L$ by the accurate rate (AR) \citep{Q_Wang_Handwritten} using Algorithm \ref{Alg_Line_Matching}. The algorithm first calculates the AR between every pair of line-level transcripts and recognition results, and then obtains matching pairs $M_l$ in the descending order of all the calculated ARs, where the threshold $th_{AR}$ is used to filter out poor recognition results. {\bl Specifically, $(p, q) \in M_l$ indicates that the recognition result $L^{(p)}$ is matched to the line-level transcript $A^{(q)}$.} 

\noindent(2) \textit{Character matching:} The characters in each pair of lines are matched according to the edit distance using Algorithm \ref{Alg_Character_Matching}, where ``$E$", ``$S$", ``$I$", and ``$D$" denote ``equal", ``substitution", ``insertion", and ``deletion", respectively. {\bl The algorithm outputs the character matches $M_c$ and consecutive equals $M_{ce}$. Specifically, $(p,m,q,n) \in M_c$ indicates that the character $c^{(p,m)}$ in the results is matched to the character $c_{gt}^{(q, n)}$ in the annotations, and $M_{ce}$ contains the indices of the character in the results where two consecutive states of computing edit distance are ``equal.''}

\noindent\textbf{Spatial Matching.} If several same or similar sentences occur in the line-level transcripts $A$, it is possible that one line in the recognition results $L$ is matched to any one of these sentences. This type of matching ambiguity cannot be solved using semantic matching. Therefore, spatial matching is proposed to address this issue. If the character $c^{(p, m)}$ in the result $R$ is matched to the character $c_{gt}^{(q, n)}$ in the annotations $A$, we calculate the intersection over union (IoU) between bounding boxes $(x^{(p,m)}, y^{(p,m)}, w^{(p,m)}, h^{(p,m)})$ and $A_{ps}^{(q,n)}$. If the IoU is lower than the threshold $th_{IoU}$, the matching pair $(p, m, q, n)$ is removed from the character matches $M_c$.

\begin{algorithm}[tb]
	\footnotesize
	\caption{Character Matching}
	\label{Alg_Character_Matching}
	\LinesNumbered
	\KwIn{line matches $M_l$, {recognition results $L$}, transcripts $A$}
	\KwOut{character matches $M_c$, consecutive equals $M_{ce}$}
	\ForEach{$(p, q) \in M_l$}{
		compute edit distance between $L^{(p)}$ and  $A^{(q)}$\;
		$\xi = \{\xi^{(1)}, \xi^{(2)}, ... | \xi^{(j)} \in \{``E", ``S", ``I", ``D"\}\}$ are the matching states of $L^{(p)}$\;
		\For{$j=1$ to $|\xi|$}{
			\If{$\xi^{(j)}$ is $``E"$}{
				$\xi^{(j)}$ corresponds to $c^{(p,m)} = c^{(q,n)}_{gt}$\;
				add $(p, m, q, n)$ to $M_c$\;
				\If{$\xi^{(j+1)}$ is $``E"$ or $j=|\xi|$}{
					add $(p, m)$ to $M_{ce}$\;
				}
			}
		}
	}
\end{algorithm}

\subsubsection{Updating}
After matching, the pseudo-labels $A_{ps}$ are updated using Algorithm \ref{Alg_Update}, where a pseudo-label is either copied from the predicted bounding box of the matched character or updated as the weighted sum of the existing pseudo-label and the predicted bounding box of the matched character. Specifically, the weight $\lambda$ is calculated based on the scores of the predicted bounding box and existing pseudo-label. 
Because the predicted bounding boxes or pseudo-labels with low scores are usually inaccurate and thus should have a much lower influence on the updated pseudo-labels, the exponential function and scale factor $\epsilon$ are used to enlarge the gap between $B^{(p, m)}_{sco}$ and $\gamma^{(q,n)}$ when calculating the weight $\lambda$.
\begin{algorithm}[t]
	\footnotesize
	\caption{Pseudo-label Update}
	\label{Alg_Update}
	\LinesNumbered
	\KwIn{character matches $M_c$, predicted results $R$, pseudo-labels $A_{ps}$, scores of bounding boxes $B_{sco}$, scores of pseudo-labels $\gamma$}
	\KwOut{updated pseudo-label $A_{ps}$}
	\ForEach{$(p, m, q, n)$ $\in$ $M_c$}{
		\eIf{ $A_{ps}^{(q,n)}$ does not exist}{
			$A_{ps}^{(q,n)} = (x^{(p,m)}, y^{(p,m)}, w^{(p,m)}, h^{(p,m)})$\;
			$\gamma^{(q,n)} = B_{sco}^{(p,m)}$\;
		}
		{
			$\lambda = e^{\epsilon \times \gamma^{(q,n)}} / (e^{\epsilon \times \gamma^{(q,n)}} + e^{\epsilon \times B_{sco}^{(p,m)}})$\;
			$A_{ps}^{(q,n)} = \lambda \times A_{ps}^{(q,n)} + (1 - \lambda) \times (x^{(p,m)}, y^{(p,m)}, w^{(p,m)}, h^{(p,m)})$\;
			$\gamma^{(q,n)} = \lambda \times \gamma^{(q,n)} + (1 - \lambda) \times B_{sco}^{(p,m)}$\;
		}
	}
	
\end{algorithm}

\subsubsection{Optimization}
\label{Sec_Method_WSL_Optim}
Because the pseudo-labels $A_{ps}$ may not contain the bounding box of every character in the line-level transcripts $A$, it becomes challenging to effectively optimize the network. In the following, we introduce how the losses are calculated for each part of the model when the pseudo-labels are incomplete.

First, we define {\bl $S_c$ as the mapping relationship between the grids and existing pseudo-labels, which is given by}
\begin{equation}
	\label{Equ_Pos_Character}
	\small
	\begin{aligned}
		S_c\! =\! &\{(i\!, j\!, q\!, n)|\exists A_{ps}^{(q,n)}, 
		(\lceil \frac{x_{ps}^{(q,n)} \times W_g}{W} \rceil, \lceil \frac{y_{ps}^{(q,n)} \times H_g}{H} \rceil) = (i,j)\},
	\end{aligned}
\end{equation}
where $(i,j,q,n) \in S_c$ means pseudo-label $A_{ps}^{(q, n)}$ exists and corresponds to grid $G^{(i, j)}$.

\noindent\textbf{\bl CharDis Branch.} For the {\bl CharDis branch}, it is easy to find positive samples ($S_c$) from the existing pseudo-labels. However, determining negative samples becomes a difficult problem if not all pseudo-labels exist. In Eq. (\ref{Equ_Neg_Conf}), we view the grids in the search paths that begin at consecutive equal characters as negative samples. Because the characters at two ends of these search paths are matched as ``equal'' consecutively, there is no character in these grids.
Therefore, the loss of the {\bl CharDis branch} is calculated using Eq. (\ref{Equ_Loss_Conf}).
\begin{gather}
	S_d^n =\{(i, j)|\exists (p, m) \in M_{ce}, (i, j) \in P^{(p,m)}_{sch} \}, \label{Equ_Neg_Conf}\\
	L_{{\bl dis}}\! =\! -\! \frac{1}{2|S_c|}\! \sum_{(i, j, q, n)\! \in S_c}\! log(O_{{\bl dis}}^{(i,j)})\! -\!
	\frac{1}{2|S_d^n|}\! \sum_{(i, j)\! \in\! S_d^n}\! log(1\! -\! O_{{\bl dis}}^{(i,j)}). \label{Equ_Loss_Conf}
\end{gather}

\noindent\textbf{\bl CharBox Branch.} First, each existing pseudo-label $A_{ps}^{(q,n)}$ is transformed back to $O_{ps}^{(q,n)}$ using Eq. (\ref{Equ_final_bb}) inversely. {\bl The loss of the CharBox branch is then calculated as the mean square error between every $O_{ps}^{(q,n)}$ generated from the existing pseudo-labels and its corresponding output $O_{box}^{(i, j)}$ of the CharBox branch:}
\begin{equation}
	\small
	\begin{aligned}
	L_{{\bl box}} = \frac{1}{|S_c|} \sum_{(i,j, q, n) \in S_c} (O_{{\bl box}}^{(i,j)} - O_{ps}^{(q,n)}) W_{{\bl box}} (O_{{\bl box}}^{(i,j)} - O_{ps}^{(q,n)})^T,
	\end{aligned}
\end{equation}
\noindent where $W_{{\bl box}}$ is a diagonal matrix. The elements on the diagonal are the weights
$(\delta_x, \delta_y, \delta_w, \delta_h)$ that are set to $(1, 1, 0.1, 0.1)$. 

\noindent\textbf{\bl CharCls Branch.} {\bl For each existing pseudo label $A_{ps}^{(q,n)}$, we can obtain the character classification probabilities $O_{cls}^{(i,j)}$ at its corresponding grid $G^{(i, j)}$ (Eq. \ref{Equ_Pos_Character}) and the ground-truth character category $c_{gt}^{(q,n)}$. Thus, the loss of the {\bl CharCls branch} is calculated as the cross entropy loss between them:}
\begin{equation}
	\small
	L_{cls} = - \frac{1}{|S_c|} \sum\nolimits_{(i, j, q, n) \in S_c} log(O_{cls}^{(i, j, c^{(q,n)}_{gt})}).
\end{equation}

\noindent\textbf{Start-of-Line.} For the start-of-line distribution $O_{sol}$, if the pseudo-label of the first character in a line exists, its corresponding grid is selected as a positive sample. The grids corresponding to other existing pseudo-labels are regarded as negative samples. Therefore, the loss is calculated as
\begin{gather}
S_s^{p} = \{(i, j)|\exists(i, j, q, n) \in S_c, n=1\}, \\
S_s^{n} = \{(i, j) | \exists(q, n), (i, j, q, n) \in S_c\} - S_s^{p}, \label{Equ_S_s_n}\\
L_{sol}\! =\! -\! \frac{1}{2|S_s^{p}|}\! \sum_{(i, j)\! \in\! S_s^{p}}\! log(O_{sol}^{(i,j)})\! -\! \frac{1}{2|S_s^{n}|}\! \sum_{(i, j)\! \in\! S_s^{n}}\! log(1\! -\! O_{sol}^{(i,j)}), 
\end{gather}
where the subtraction in Eq. (\ref{Equ_S_s_n}) means removing the elements in $S_s^{p}$ from $\{(i, j) | \exists(q, n), (i, j, q, n) \in S_c\}$.

\noindent\textbf{End-of-Line.} The loss $L_{eol}$ of the end-of-line distribution $O_{eol}$ can be obtained similarly to $L_{sol}$, by viewing the grids corresponding to the pseudo-labels of the last characters of lines as positive samples and those corresponding to other pseudo-labels as negative samples.

\begin{algorithm}[t]
	\footnotesize
	\caption{Paths Generating}
	\label{Alg_O_RD}
	\LinesNumbered
	\KwIn{pseudo-labels $A_{ps}$}   
	\KwOut{paths $S_{rd}$}
	$D_{rd} = \{(0, -1), (1, 0), (0, 1), (-1, 0)\}$\;
	\For{$q=1$ to $\hat{N}_{ln}$ and $n=1$ to $\hat{N}_{ch}^{(q)}$}{
		\If{$A_{ps}^{(q,n)}$ exists and $A_{ps}^{(q,n+1)}$ exists}{
			$(i, j) = (\lceil \frac{x_{ps}^{(q,n)} \times W_g}{W} \rceil, \lceil \frac{y_{ps}^{(q,n)} \times H_g}{H} \rceil)$\;
			$(s, t) = (\lceil \frac{x_{ps}^{(q,n+1)} \times W_g}{W} \rceil, \lceil \frac{y_{ps}^{(q,n+1)} \times H_g}{H} \rceil)$\;
			$\zeta = |s - i| + |t - j|$\;
			initialize a $\zeta \times 1$ vector $\vec x$ with $sgn(s - i)$\;
			initialize a $\zeta \times 1$ vector $\vec y$ with 0\;
			randomly set $|t-j|$ elements in $\vec y$ to $sgn(t-j)$, and set the elements at the same indices in $\vec x$ to 0\;
			\For{$k=1$ to $\zeta$}{
				$d \leftarrow$ the index of $(\vec x^{(k)}, \vec y^{(k)})$ in $D_{rd}$\;
				add $(i, j, d)$ to $S_{rd}$\;
				$(i, j) = (i, j) + (\vec x^{(k)}, \vec y^{(k)})$\;
			}
		}
	}
\end{algorithm}

\noindent \textbf{4-Directional Reading Order.} For the 4-directional reading order predictions $O_{rd}$, we randomly generate paths {\bl $S_{rd}$} between the grids corresponding to consecutive pseudo-labels using Algorithm \ref{Alg_O_RD}, {\bl where $(i,j,d) \in S_{rd}$ indicates that grid $G^{(i,j)}$ with $d$ direction is in the paths.} Then the loss is given by
\begin{equation}
	L_{rd} = - \frac{1}{|S_{rd}|} \sum\nolimits_{(i, j, d) \in S_{rd}} log(O_{rd}^{(i, j, d)}).
\end{equation}
\noindent \textbf{Total Loss.} The total loss $L_{total}$ is given by Eq. (\ref{L}). The model parameters are optimized to minimize the loss.
\begin{equation}
	\label{L}
	L_{total} = L_{{\bl box}} + L_{cls} + L_{{\bl dis}} + L_{sol} + L_{eol} + L_{rd},
\end{equation}
{\bl where the total loss is the simple sum of all the loss terms. Although the performance may be improved when the weighting factors for the loss terms are carefully tuned on the target dataset, we formulate our method in a more generic manner.}

\section{Experiments}
\label{Sec_Exp}
\subsection{Dataset}
\label{Sec_Dataset}
\textbf{CASIA-HWDB} \citep{C_Liu_CASIA} is a large-scale Chinese handwriting database. We use two offline databases, namely \textbf{CASIA-HWDB1.0-1.2} and \textbf{CASIA-HWDB2.0-2.2}. CASIA-HWDB1.0-1.2 contains 3,895,135 isolated character samples. CASIA-HWDB2.0-2.2 contains 5,091 pages.

\noindent\textbf{ICDAR2013} \citep{F_Yin_ICDAR13} includes a page-level dataset (\textbf{ICDAR13}) and a single character dataset (\textbf{ICDAR13-SC}). There are 300 pages in ICDAR13 and 224,419 character samples in ICDAR13-SC. The number of character categories is 7,356 when conducting experiments on CASIA-HWDB and ICDAR2013.

\noindent\textbf{MTHv2} \citep{W_Ma_Joint} contains 3,199 pages of historical documents, including 2,399 pages for training and 800 pages for testing. There are 6,762 categories of characters in MTHv2.

\noindent\textbf{SCUT-HCCDoc} \citep{H_Zhang_SCUT-HCCDoc} contains 12,253 camera-captured documents with 6,109 categories of characters. The training and testing sets contain 9,801 images and 2,452 images, respectively.

\noindent\textbf{JS-SCUT PrintCC} is an in-house dataset that consists of 398 scanned images of printed documents. The images are divided into 348 for training and 50 for testing. 
{\bl There are 2,652 character classes in the dataset.}

\begin{figure}[t]
	\centering 
	\subfigure[CASIA-SR]{
		\begin{minipage}[t]{0.5\columnwidth}
			\centering
			\includegraphics[height=0.6\columnwidth]{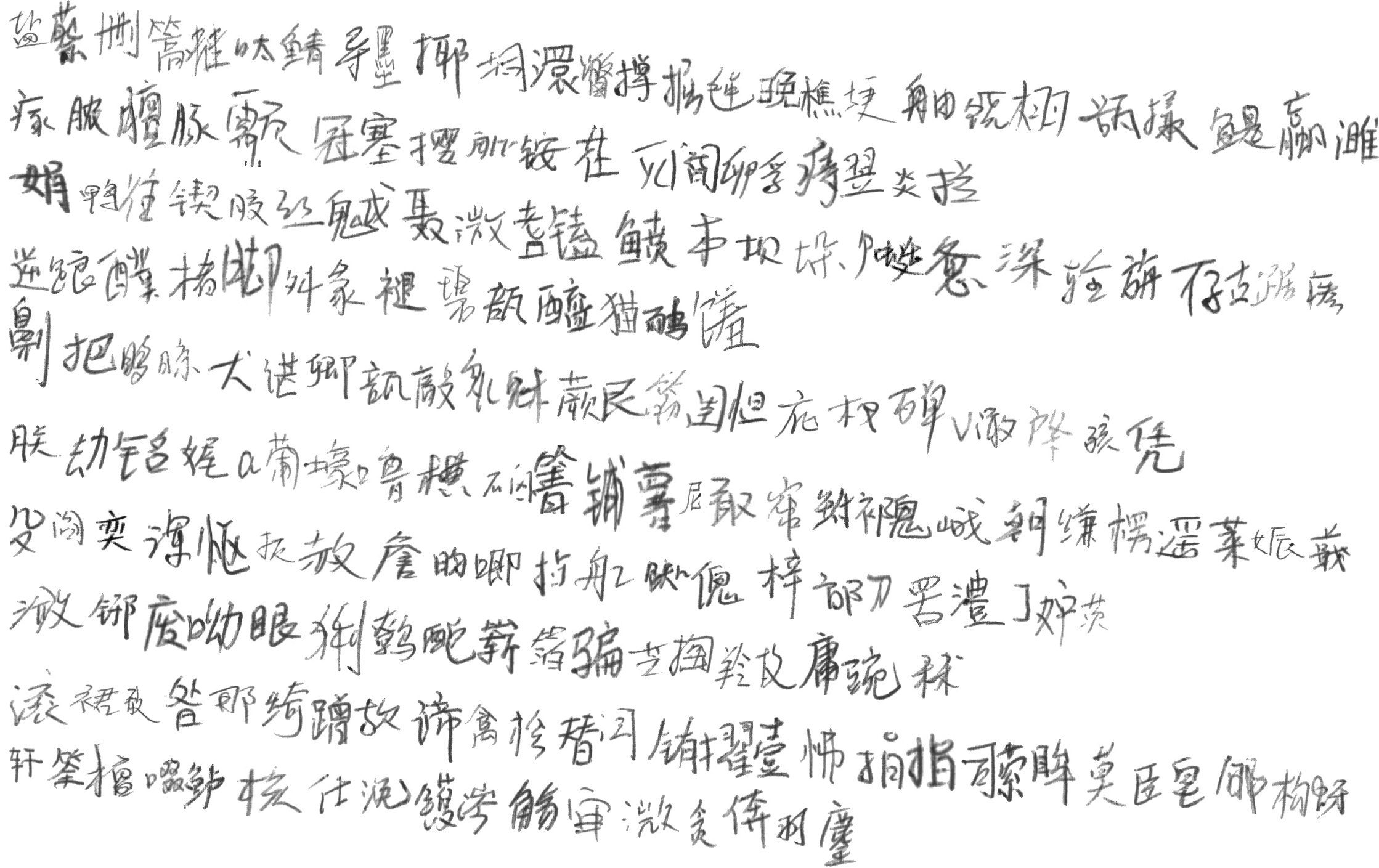}
	\end{minipage}}%
	\subfigure[MTH-SFB]{
		\begin{minipage}[t]{0.5\columnwidth}
			\centering
			\includegraphics[height=0.6\columnwidth]{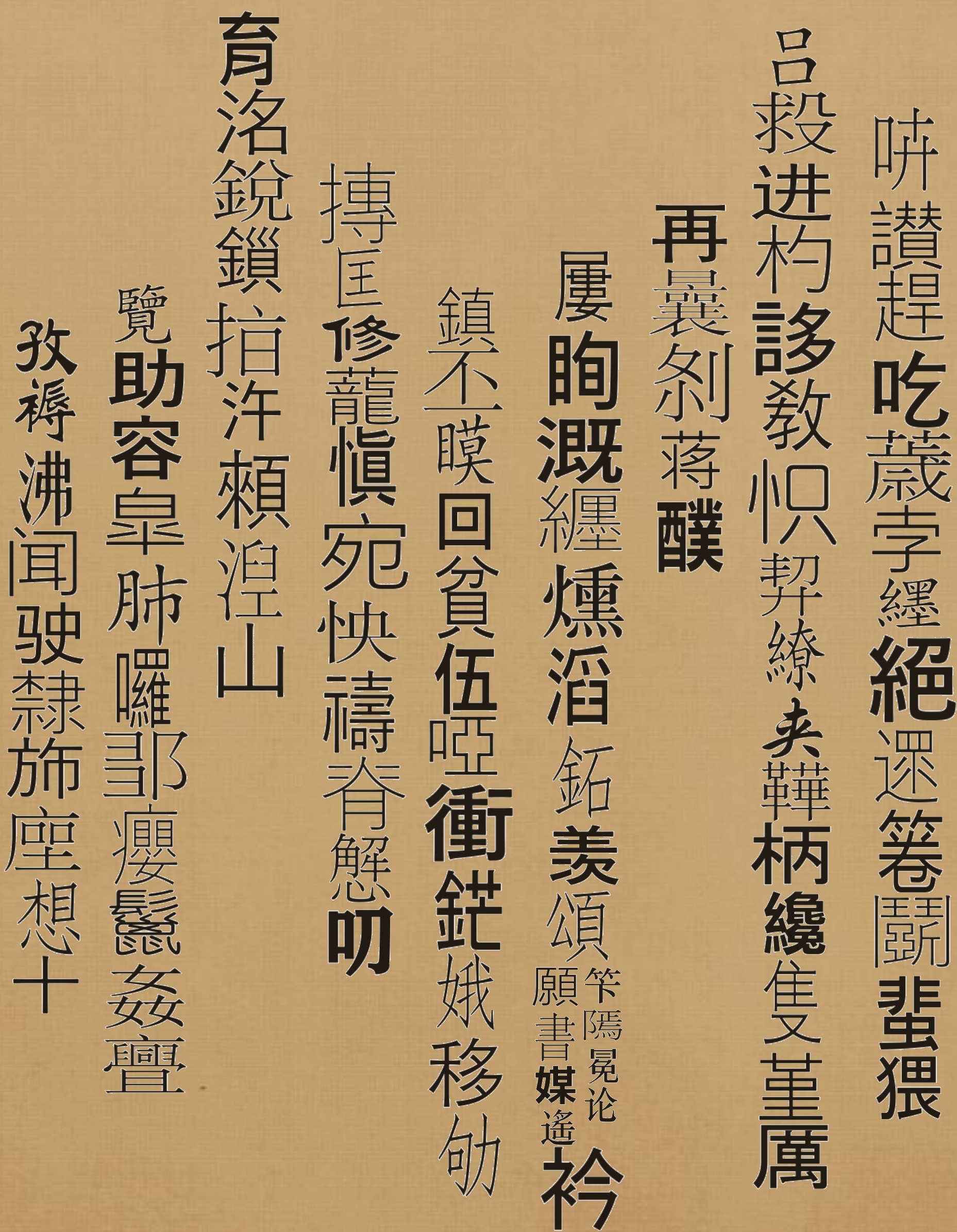}
	\end{minipage}}
	\subfigure[HCCDoc-SFB]{
		\begin{minipage}[t]{0.3\columnwidth}
			\centering
			\includegraphics[height=0.8\columnwidth]{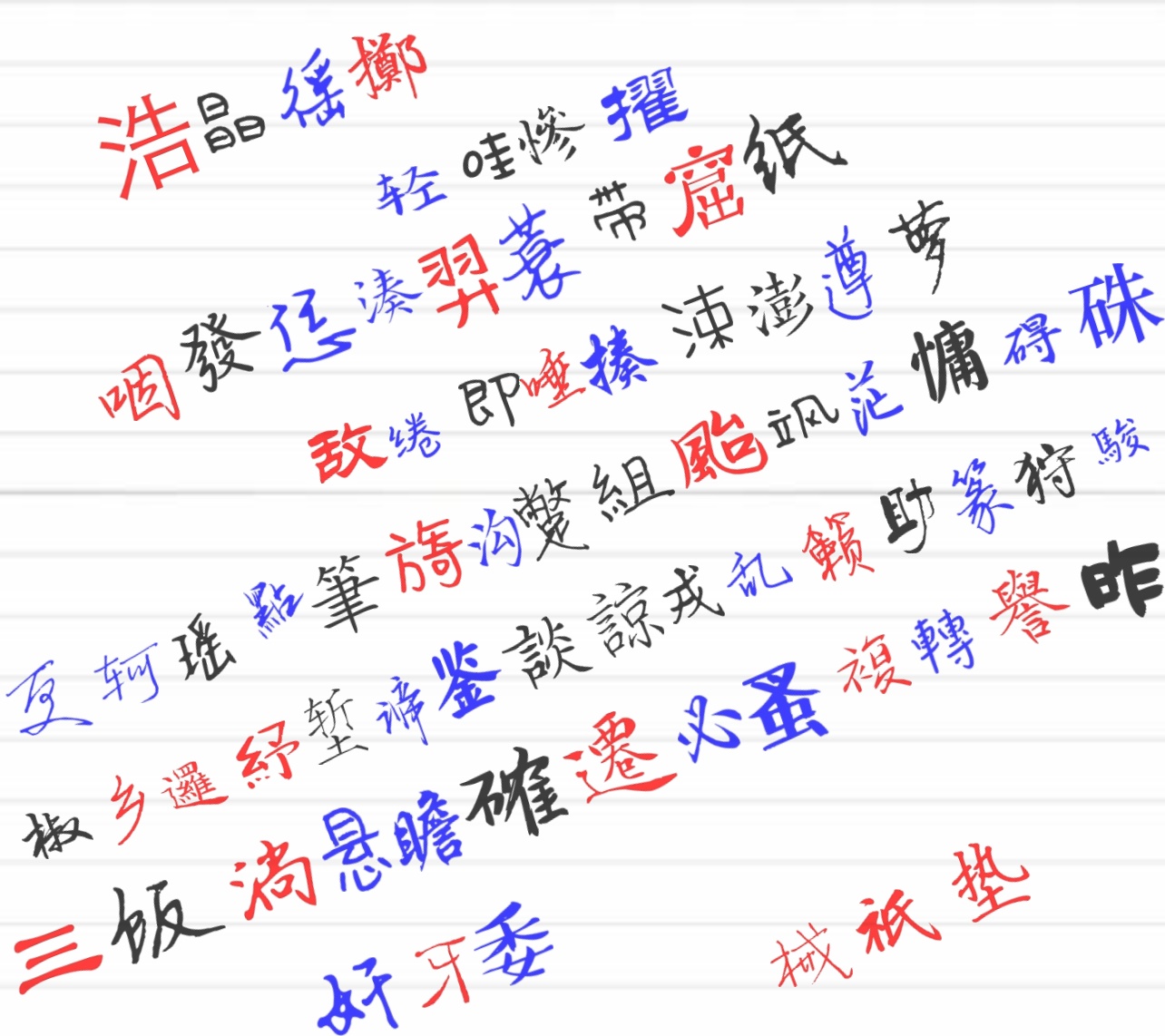}
	\end{minipage}}%
	\subfigure[JS-SF]{
		\begin{minipage}[t]{0.7\columnwidth}
			\centering
			\includegraphics[height=0.34\columnwidth]{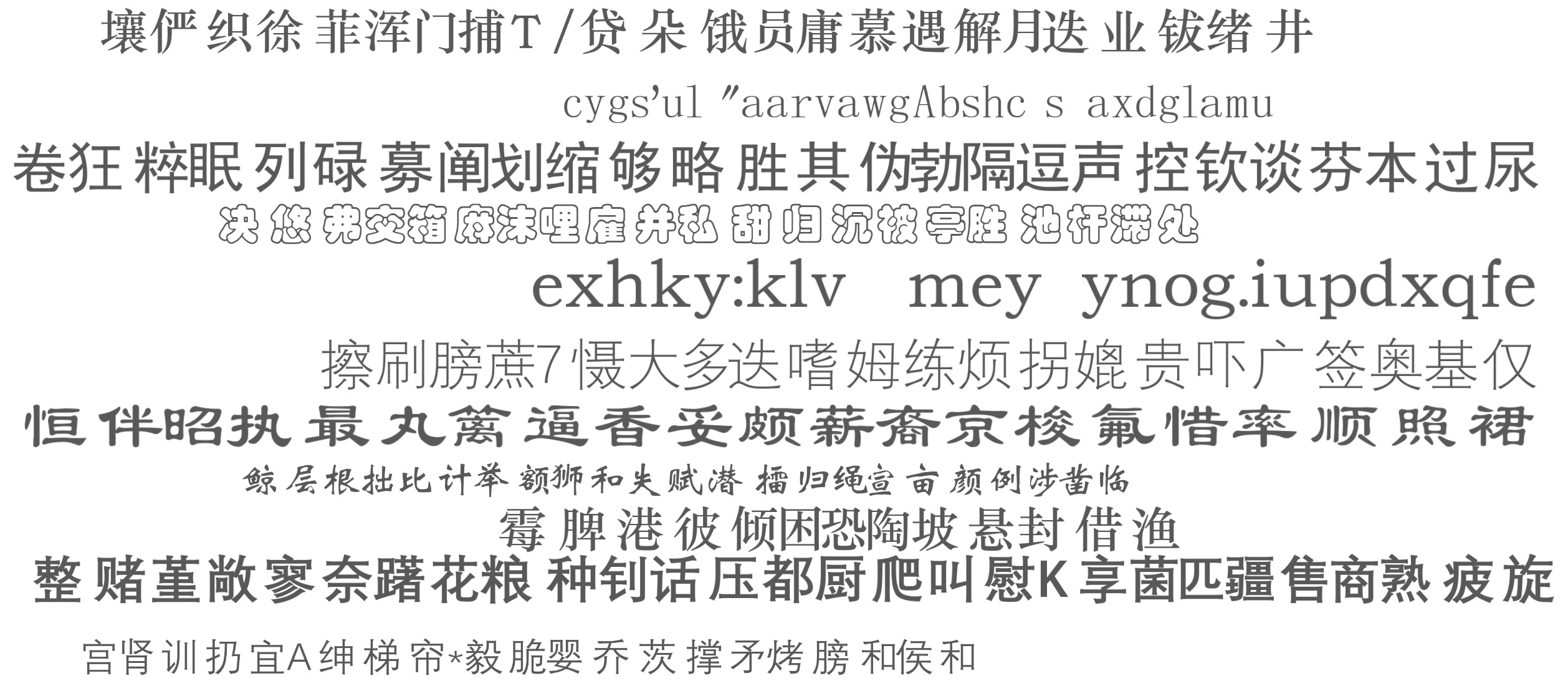}
	\end{minipage}}
	\caption{Example images from CASIA-SR, MTH-SFB, HCCDoc-SFB, and JS-SF.}
	\label{Fig_Exp_Dataset_Syn}
\end{figure}

\noindent\textbf{Synthetic Dataset.} As shown in Fig. \ref{Fig_Exp_Dataset_Syn}, we synthesize four datasets, namely CASIA-SR, MTH-SFB, HCCDoc-SFB, and JS-SF. 
``SR'' and ``SF'' denote synthesizing using character samples from real isolated character databases and font files, respectively. 
The datasets whose names end with ``B'' are synthesized using background images rather than a white background. We adopt 101 font files and 32 background images that are downloaded from the Internet {\bl without using knowledge about real datasets. Specifically, the font files are randomly selected from the free fonts of the FounderType website\footnote{\bl http://www.foundertype.com/}. The background images are chosen from the pictures obtained by searching for ``paper'' on the Internet because our work is aimed at document recognition.} For the CASIA-SR dataset,
single character samples from CASIA-HWDB1.0-1.2 are used. All synthetic datasets have full annotations, i.e., line-level transcripts and bounding boxes of characters. Despite the different layouts of the real datasets, all four synthetic datasets follow a simple synthesis procedure. First, we synthesize text lines with randomly selected characters and obliquities. Note that no corpus is used when synthesizing these text lines. Afterwards, multiple text lines are combined to form a page. There is also no perspective transformation or illumination applied to the synthetic image.

\subsection{Training Strategy}
\label{sec_training_strategy}
\vspace{-0.2cm}
Directly training the randomly initialized PageNet using the proposed weakly supervised learning framework will lead to unsatisfactory performance. This is because of the low accuracy of the model and the lack of pseudo-labels during early iterations. Therefore, an improved training strategy is proposed, which consists of pretraining, initializing, and training stages. First, in the pretraining stage, the model is pretrained using synthetic samples. Then, in the initializing stage, the procedure is the same as Fig. \ref{Fig_WSL_Overview} but without the optimization, which means that the pseudo-labels are not used to train the model. However, a part of the pseudo-labels is initialized and updated during the initializing stage. Finally, the training stage is exactly the procedure shown in Fig. \ref{Fig_WSL_Overview}.

\subsection{Implementation Details}
\label{sec_imp}
\subsubsection{Network Architecture}
\label{sec_network_arch}
\begin{figure}[t]
	\centering
	\includegraphics[width=1.0\columnwidth]{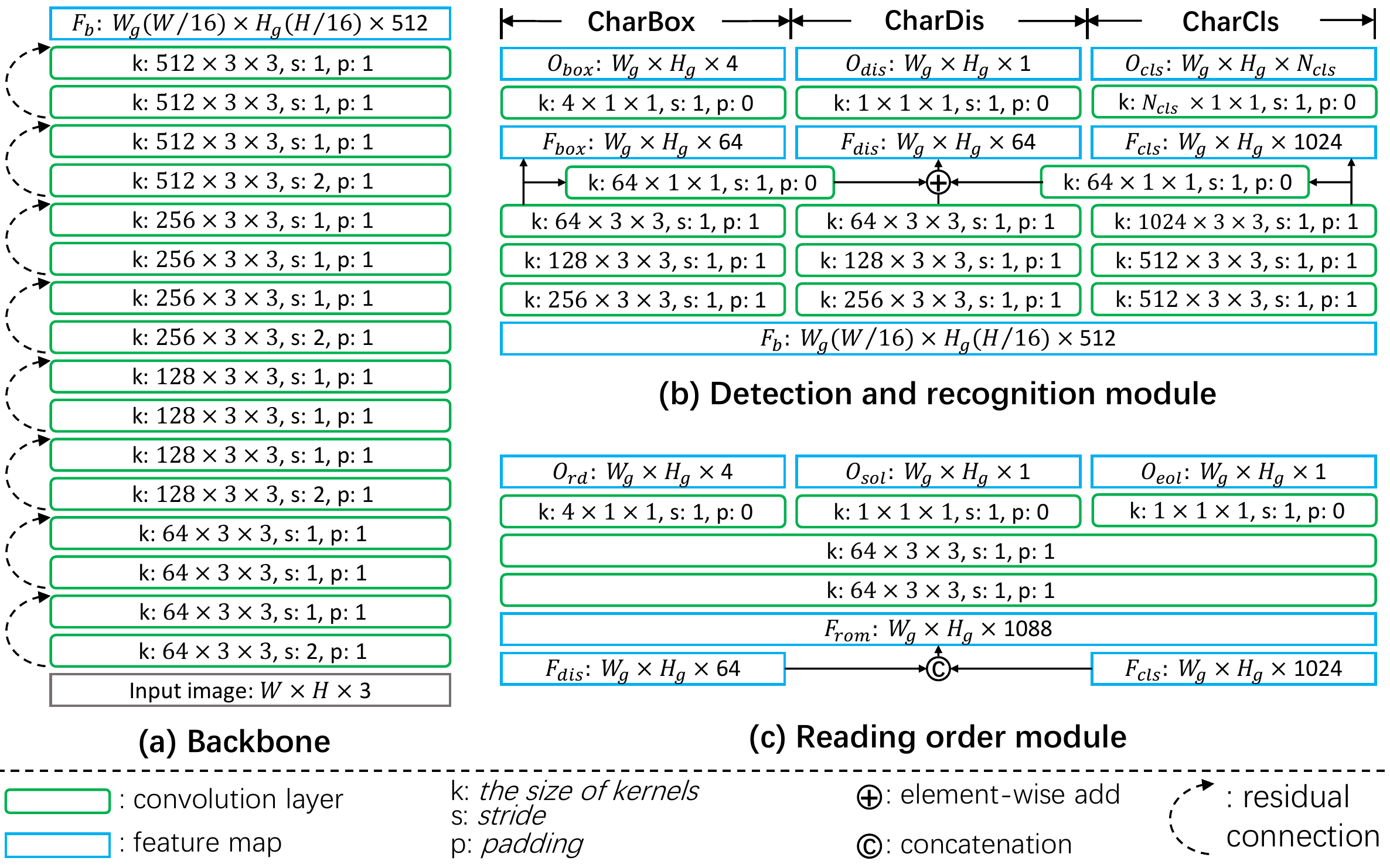}
	{\bl \caption{Detailed network architecture.}\label{Fig_Exp_Arch}}
\vspace{-0.2cm}
\end{figure}

The detailed network architecture is illustrated in Fig. \ref{Fig_Exp_Arch}. {\bl The architecture of the backbone follows the previous work \citep{D_Peng_A_Fast}, which is verified to be effective for HCTR. It can also be easily changed to a standard backbone, such as ResNet \citep{he2016deep}.} In the detection and recognition module, the interaction between the three branches is the same as that in the segmentation and recognition module \citep{D_Peng_A_Fast}. First, the feature $F_{{\bl box}}$ from the {\bl CharBox branch} and the feature $F_{cls}$ from the {\bl CharCls branch} each go through a $1 \times 1$ convolution layer. Then, the two output features and the feature from the last convolution layer of the {\bl CharDis branch} are element-wise added, yielding the feature $F_{{\bl dis}}$.

\subsubsection{Graph-based Decoding Algorithm}
\noindent\textbf{Score of Bounding Boxes.} In Sec. \ref{sec_char_seg_rec}, NMS is used to remove redundant bounding boxes. The confidence in the character distribution $O_{{\bl dis}}$ can be used as the score of bounding boxes. However, following the segmentation and recognition module \citep{D_Peng_A_Fast}, semantic information is integrated into the score of bounding boxes. Specifically, the score is the weighted sum of the character distribution confidence and maximum classification probability. 
The weight of the character distribution confidence is set to 0.8 following \citet{D_Peng_A_Fast}.

\noindent\textbf{Edges.} As shown in the search path of Fig. \ref{Fig_Model_GDA}, 
the next node should be at the next grid of the final grid in the search path.
However, this is too strict and the 4-directional reading order prediction $O_{rd}$ must be very accurate. Therefore, the next node is only required to be in the 4-neighborhoods of the final grid in the search path. Furthermore, we limit the maximum number of steps in a search path for acceleration. 

\noindent\textbf{Start-of-Line and End-of-Line.} In Sec. \ref{Model_GDA_RO}, a node is identified as the start-of-line or the end-of-line if the corresponding confidence in $O_{sol}$ or $O_{eol}$ is greater than 0.9. 

\noindent\textbf{Special Property of Graph.} In page-level documents, a character has at most one previous character and one next character. Therefore, for a node in the graph, we must ensure that there is at most one edge in and one edge out. If there are multiple nodes in the 4-neighborhoods of the final grid in the search path and the direction of the final grid does not point to any one of them, the node whose corresponding bounding box has the highest score is selected. If there are multiple edges ending at the same node, the edge whose slope is closest to the slopes of the previous edges in the path is maintained. 
\subsubsection{Weakly Supervised Learning}
The threshold $th_{AR}$ in Algorithm \ref{Alg_Line_Matching}, $th_{IoU}$ in spatial matching, and the scale factor $\epsilon$ in Algorithm \ref{Alg_Update} are set to 0.3, 0.5 and 10, respectively.

\subsubsection{Experiment Settings}
\label{sec_exp_setting}
We implement our method with PyTorch and conduct experiments using an NVIDIA RTX 2080ti GPU with 11GB of memory. Stochastic gradient descent with a batch size of 1 is used to optimize the network. Both the pretraining and training stages contain 300,000 iterations, and the learning rate is initialized to 0.01 and multiplied by 0.1 after 100,000, 200,000, and 275,000 iterations. The initializing stage contains 75,000 iterations and the learning rate is set to 0.0001. 
During the initializing and training stages, the probabilities of loading real samples and synthetic samples are 0.7 and 0.3, respectively. {\bl In the training stage, following existing methods \citep{C_Xie_High,Y_Baek_Character}, we use synthetic samples in addition to real samples to increase the diversity of training data and improve the stability of training.}
No validation set is adopted. 
All the training and testing images are resized to normalize their widths while maintaining their aspect ratios. The pixel value of the input image is normalized to the range of [0, 1]. Gaussian noise with a mean of 0 and a variance of 0.01 is applied to the synthetic images. {\bl Other settings for specific experiments are listed as follows, where the image width is estimated based on the number of characters on a page. 
Because Chinese characters are composed of complicated strokes, we should ensure that the characters are recognizable with the given input size, as well as consider training efficiency.
}

\noindent\textbf{ICDAR13.} The model is trained using 5,091 real samples from CASIA-HWDB2.0-2.2 and 20,000 synthetic samples from CASIA-SR. The testing is conducted on 300 samples from ICDAR13. The width of the input image is normalized to 1,920 pixels.

\noindent\textbf{MTHv2.} We use the train set of MTHv2 and 10,000 samples from MTH-SFB to train the model and test it on the test set of MTHv2. The width of the input image is normalized to 2,960 pixels.

\noindent\textbf{SCUT-HCCDoc.} The model is trained using the training images from SCUT-HCCDoc and 20,000 synthetic images from HCCDoc-SFB. The test set of SCUT-HCCDoc is used to evaluate the model. The width of the input image is normalized to 1,600 pixels. Note that we adopt $4\times$ iterations in the training stage of this experiment owing to the more complex scenarios and larger scale of SCUT-HCCDoc.

\noindent\textbf{JS-SCUT PrintCC.} The model is trained using the training samples from JS-SCUT PrintCC and 10,000 synthetic samples from JS-SF. The trained model is evaluated on the test samples of JS-SCUT PrintCC. The width of the input image is normalized to 2,080 pixels.
\subsection{Evaluation Metrics}
\label{Sec_Eval}
{\bl Our method requires only line-level transcripts to be annotated for real data.}
However, there is no metric to evaluate the performance when only line-level transcripts are annotated. To this end, we propose two evaluation metrics termed accurate rate* (AR*) and correct rate* (CR*). First, a matching algorithm, which is the same as Algorithm \ref{Alg_Line_Matching} but without filtering out poor recognition results at line 5, is executed, yielding line matches $M_l^*$. 
Moreover, we define $S^*_{R}$ and $S^*_{A}$ as the indices of unpaired lines in the results and annotations, respectively.
Then AR* and CR* are given by

\begin{small}
	\begin{align}
		&N_{Ie}^* = \sum\nolimits_{(p, q) \in M_l^*} IE(L^{(p)}, A^{(q)}) + \sum\nolimits_{i \in S^*_R} N^{(i)}_{ch}, \\
		&N_{De}^* = \sum\nolimits_{(p, q) \in M_l^*} DE(L^{(p)}, A^{(q)}) + \sum\nolimits_{i \in S^*_A} \hat{N}^{(i)}_{ch}, \\
		&N_{Se}^* = \sum\nolimits_{(p, q) \in M_l^*} SE(L^{(p)}, A^{(q)}), \\
		&N_{total}^* = \sum\nolimits_{i} \hat{N}^{(i)}_{ch}, \\
		&AR^* = (N_{total}^* - N_{Ie}^* - N_{De}^* - N_{Se}^*)/N^*_{total}, \\
		&CR^* = (N_{total}^* - N_{De}^* - N_{Se}^*)/N^*_{total},
	\end{align}
\end{small}
where the functions $IE$, $DE$, and $SE$ compute the number of insertion, deletion, and substitution errors between the two input sequences, respectively. 
The errors between every matching pair are accumulated, and all the characters of the unpaired lines in the results and annotations are viewed as insertion and deletion errors, respectively. {\bl Compared with the vanilla accurate rate (AR) and correct rate (CR) \citep{Q_Wang_Handwritten} for line-level text recognition which indicates only text recognition performance, the proposed AR* and CR* for page-level text recognition considers both text detection and recognition.}

\begin{table*}[htb]
	\centering
	\caption{Comparison with existing page-level methods on ICDAR13}
	\label{TBL_ICDAR13}
	\begin{tabular*}{\hsize}{@{}@{\extracolsep{\fill}}llllllllll@{}}
		\hline
		\multirow{2}*{Supervision} & \multirow{2}*{Method} & \multicolumn{5}{l}{End-to-End Recognition} & \multicolumn{3}{l}{Text Line Detection} \\
		\cline{3-7} \cline{8-10}
		& & AR* & CR* & AR & CR & NED & Precision & Recall & F-measure \\
		\hline
		\multicolumn{1}{c}{\multirow{3}*{Full}} & Det + Recog {\bl \citep{K_He_Mask,C_Xie_High}} & 88.36 & 89.09 & 88.39 & 89.08 & 88.27 & 99.54 & 99.88 & 99.71\\
		\multicolumn{1}{c}{} & Mask TextSpotter \citep{P_Lyu_Mask} & 49.48 & 57.95 & 50.60 & 58.29 & 50.40 & 91.21 & 96.77 & 93.91 \\
		\multicolumn{1}{c}{} & FOTS \citep{X_Liu_FOTS} & 67.20 & 67.75 & 67.32 & 67.82 & 65.89 & 98.64 & 97.29 & 97.96\\
		\hline
		\multicolumn{1}{c}{\multirow{3}*{Weak}} & Start-Follow-Read \citep{C_Wigington_Start} & 82.60 & 83.42 & 82.91 & 83.55 & 82.35 & \textbf{99.88} & 98.89 & 99.39\\
		\multicolumn{1}{c}{} & OrigamiNet \citep{M_Yousef_OrigamiNet} & - & - & 5.99 & 5.99 & - & - & - & - \\
		\multicolumn{1}{c}{} & \textbf{PageNet (Ours)} & \textbf{92.83} & \textbf{93.23} & \textbf{92.86} & \textbf{93.24} & \textbf{92.49} & 99.56 & \textbf{99.94} & \textbf{99.75} \\
		\hline
	\end{tabular*}
\end{table*}

\subsection{Line-level Detection and Recognition}
\subsubsection{Performance on ICDAR13 Dataset}
In Table \ref{TBL_ICDAR13}, we compare the line-level detection and recognition results of our approach with existing page-level methods on the ICDAR13 dataset. The page-level methods include fully supervised approaches such as Det + Recog, Mask TextSpotter \citep{P_Lyu_Mask,M_L_Mask}, and FOTS \citep{X_Liu_FOTS}, as well as weakly supervised approaches such as Start-Follow-Read \citep{C_Wigington_Start} and OrigamiNet \citep{M_Yousef_OrigamiNet}. The method denoted as Det + Recog is the combination of two independently trained models which are a Mask R-CNN \citep{K_He_Mask} (for text line detection) and a recognizer \citep{C_Xie_High} (for text line recognition). The recognizer \citep{C_Xie_High} achieves state-of-the-art performance on the text line recognition task of ICDAR13, as shown in Table \ref{TBL_ICDAR13_Line}. The fully supervised methods are trained using CASIA-SR and fully annotated CASIA-HWDB2.0-2.2, whereas the weakly supervised methods are trained using CASIA-SR and weakly annotated CASIA-HWDB2.0-2.2.

In addition to the proposed AR* and CR*, other evaluation metrics adopted in Table \ref{TBL_ICDAR13} are as follows.
(1) Because there is only one paragraph on each page of ICDAR13, page-level transcript annotations and recognition results can be easily obtained. 
Thus, we calculate the page-level AR and CR in Table \ref{TBL_ICDAR13}.
(2) The normalized edit distance (NED) is calculated following the evaluation protocol of task 4 in \citep{R_Zhang_RECTS}, which considers both text line detection and recognition. Because the results of our method and the annotations of ICDAR13 only provide the bounding boxes of characters, the bounding box of a text line is calculated as the rotated rectangle with the minimum area enclosing the characters of this text line.
(3) Precision, recall, and f-measure are used to evaluate the performance of text line detection with an IoU threshold of 0.5.\\
\indent As shown in Table \ref{TBL_ICDAR13}, compared with existing page-level methods including three fully supervised methods, the proposed PageNet with weak supervision achieves state-of-the-art performance in terms of both end-to-end recognition and text line detection. For Det + Recog, although text line detection seems to be accurate in terms of f-measure, it is common that the bounding box of one text line contains the noise from other text lines and does not entirely cover the characters at both ends, which affects the accuracy of recognition.
For the two end-to-end methods, namely Mask TextSpotter and FOTS, the large number of categories and the diversity of writing styles of Chinese texts make recognition a heavy burden for model optimization and their feature sharing mechanism. 
OrigamiNet performs page-level text recognition by unfolding 2-dimensional features to 1-dimensional. However, in contrast to English texts, each Chinese character itself is a complex 2-dimensional structure, which may make this mechanism difficult to work.
\begin{table*}[t]
	\centering 
	\caption{Comparison with existing page-level methods on MTHv2, SCUT-HCCDoc, and JS-SCUT PrintCC datasets}
	\label{TBL_Other_Page}
	\begin{tabular*}{\hsize}{@{}@{\extracolsep{\fill}}llllllll@{}}
		\hline 
		\multirow{2}*{Supervision} & \multirow{2}*{Method} & \multicolumn{2}{l}{MTHv2} & \multicolumn{2}{l}{SCUT-HCCDoc} & \multicolumn{2}{l}{JS-SCUT PrintCC} \\
		\cline{3-4} \cline{5-6} \cline{7-8}
		& & AR* & CR* & AR* & CR* & AR* & CR* \\
		\hline
		\multirow{2}*{Full} & Det + Recog {\bl \citep{K_He_Mask,B_Shi_CRNN}} & \textbf{94.50} & \textbf{95.29} & \textbf{83.44} & \textbf{87.97} & 94.68 & 95.03 \\
		& FOTS \citep{X_Liu_FOTS} & 87.97 & 89.25 & 66.61 & 70.01 &  94.19 & 94.31 \\
		\hline 
		\multirow{3}*{Weak} & Start-Follow-Read \citep{C_Wigington_Start} & 69.54 & 73.11 & 56.29 & 61.26 & 81.36 & 82.47 \\
		& OrigamiNet \citep{M_Yousef_OrigamiNet} & 9.72 & 9.83 & 30.55 & 30.97 & 44.09 & 45.72\\
		& \textbf{PageNet (Ours)} & 93.76 & 95.23 & 77.95 & 82.15 & \textbf{97.25} & \textbf{98.19} \\
		\hline
	\end{tabular*}
\end{table*}
\subsubsection{Performance on Other Datasets}
The quantitative results of our method and existing page-level methods on MTHv2, SCUT-HCCDoc, and JS-SCUT PrintCC are listed in Table \ref{TBL_Other_Page}, where Det + Recog is the combination of a Mask R-CNN \citep{K_He_Mask} (for text line detection) and a CTC-based recognizer \citep{B_Shi_CRNN} (for text line recognition). The fully supervised methods are trained using both synthetic data and real fully annotated data, whereas the weakly supervised methods are trained using both synthetic data and real weakly annotated data.

The proposed metrics AR* and CR* are reported in Table \ref{TBL_Other_Page}, as they are verified to be effective by Table \ref{TBL_ICDAR13} and require fewer annotations compared with other metrics. In particular, for the JS-SCUT PrintCC dataset that only provides line-level transcript annotations, only AR* and CR* can be calculated. Other metrics, such as page-level AR and CR, are not applicable to these three datasets because the annotations do not provide the reading order between text lines. However, because OrigamiNet can only be trained with page-level transcripts, we concatenate the line-level transcripts in the annotations based on the spatial location and obtain fake page-level transcripts. Therefore, the results of OrigamiNet are actually page-level AR and CR.

Compared with the fully supervised methods, our method can achieve competitive performance. Specifically, compared with Det + Recog, our method achieves lower accuracy on MTHv2 and SCUT-HCCDoc but performs better on JS-SCUT PrintCC. This is because MTHv2 and SCUT-HCCDoc contain significantly more complex layouts than JS-SCUT PrintCC, which is a big challenge for the weakly supervised learning. In contrast, the text line detection part of Det + Recog can be trained significantly better under full supervision.

Compared with the weakly supervised methods, our method achieves the best performance. For Start-Follow-Read, the complex layouts lead to the failure of its line follower and end-of-line determination. However, our method can still maintain a relatively high accuracy owing to the bottom-up design and the effectiveness of the reading order module and the graph-based decoding algorithm.

Unlike other datasets, JS-SCUT PrintCC has three unique characteristics: (1) it contains printed documents, (2) 30\% of the text lines are totally in English, and (3) the training set of real data is much smaller than other datasets. Therefore, the best result achieved on JS-SCUT PrintCC verifies the capability of our method on printed documents, multilingual texts, and few training samples. 

\subsubsection{Comparison with Line-level Methods}

In Table \ref{TBL_ICDAR13_Line}, we compare our method with existing line-level methods on ICDAR13 which directly recognize the text line images cropped from the full pages based on the detection annotations. Although more stringent AR* and CR* take both text line detection and recognition into consideration, our method still achieves the best performance without language model compared with the results reported in previous literature.

\begin{table}[t]
	\centering
	\caption{Comparison with existing line-level methods on ICDAR13 (LM: language model)}
	\label{TBL_ICDAR13_Line}
	\begin{tabular*}{\hsize}{@{}@{\extracolsep{\fill}}lllll@{}}
		\hline 
		\multicolumn{1}{l}{\multirow{2}*{Method}} & \multicolumn{2}{l}{Without LM} & \multicolumn{2}{l}{With LM} \\
		\cline{2-3} \cline{4-5}
		\multicolumn{1}{l}{} & AR & CR & AR & CR \\ 
		\hline
		\citet{F_Yin_ICDAR13} & - & - & 86.73 & 88.76 \\
		\citet{R_Messina_Segmentation} & 83.50 & - & 89.40 & - \\
		\citet{Y_Wu_Handwritten} & 86.64 & 87.43 & 90.38 & - \\
		\citet{J_Du_Deep} & 83.89 & - & 93.50 & - \\
		\citet{S_Wang_Deep} & 88.79 & 90.67 & 94.02 & 95.53 \\
		\citet{Y_Wu_Improving} & - & - & 96.20 & 96.32 \\
		\citet{Z_Wang_A_Comprehensive} & 89.66 & - & 96.47 & - \\
		\citet{Z_Wang_Writer} & 91.58 & - & \textbf{96.83} & - \\
		\citet{D_Peng_A_Fast} & 89.61 & 90.52 & 94.88 & 95.51 \\
		\citet{Y_Xiu_A_Handwritten} & 88.74 & - & 96.35 & - \\
		\citet{C_Xie_High} & 91.55 & 92.13 & 96.72 & \textbf{96.99} \\
		\citet{Z_Wang_Weakly} & 87.00 & 89.12 & 95.11 & 95.73 \\
		\citet{Z_Zhu_Attention} & 90.86 & - & 94.00 & - \\
		\hline
		\multirow{2}{*}{\textbf{PageNet (Ours)}} & AR* & CR* & AR* & CR* \\
		\cline{2-3} \cline{4-5}
		 & \textbf{92.83} & \textbf{93.23} & 96.24 & 96.66 \\
		\hline
	\end{tabular*}
\end{table}

\subsubsection{Incorporation with Language Models}
\label{sec_lm}
The proposed graph-based decoding algorithm can also use n-gram language models to improve the recognition performance. For the $p$-th path in the reading order, the grids corresponding to the nodes and the grids in the search paths are concatenated as
\begin{equation}
	\begin{aligned}
		grids^{(p)} = & \{(\alpha^{(p,1)}, \beta^{(p,1)})\} \oplus P_{sch}^{(p, 1)} \oplus ...\\
		& \oplus \{(\alpha^{(p,N_{ch}^{(p)})}, \beta^{(p,N_{ch}^{(p)})})\} \oplus P_{sch}^{(p, N_{ch}^{(p)})},
	\end{aligned}
\end{equation}
where the elements in $grids^{(p)}$ are the coordinates of the grids. {\bl Specifically, as defined in Sec. \ref{sec_final_results} and \ref{sec_def_symbol}, $N_{ch}^{(p)}$ is the number of characters in the $p$-th line, $(\alpha^{(p,m)}, \beta^{(p,m)})$ is the coordinate of the grid corresponding to the $m$-th character of the $p$-th line, and $P_{sch}^{(p,m)}$ contains the coordinates of the grids in the search path starting from the $m$-th character of the $p$-th line.} Every element of $grids^{(p)}$ can be viewed as a time step similar to the decoding process of line-level recognizers. For each $(i,j) \in grids^{(p)}$, the blank probability is $1-O_{{\bl dis}}^{(i,j)}$ and the classification probabilities of $N_{cls}$ categories are $O_{cls}^{(i,j)}$. Then, we use a trigram language model generated from the same corpus as \citep{Z_Xie_Learning} and a decoding algorithm proposed by \citep{A_Graves_Towards} to obtain the recognition result of the $p$-th line. As shown in Table \ref{TBL_ICDAR13_Line}, the language model significantly improves the recognition performance, boosting AR* from 92.83\% to 96.24\%.

\begin{table*}[bp]
	\centering 
	\caption{Comparison of character-level detection and recognition result on ICDAR13}
	\label{TBL_CLSR}
	\begin{tabular*}{\hsize}{@{}@{\extracolsep{\fill}}llllllll@{}}
		\hline 
		\multirow{2}*{Supervision} & \multirow{2}*{Method} & \multicolumn{3}{l}{DetOnly} & \multicolumn{3}{l}{7356C} \\
		\cline{3-5} \cline{6-8}
		& & Precision & Recall & F-measure & Precision & Recall & F-measure \\
		\hline
		\multirow{4}*{Full} & Faster R-CNN (DetOnly) {\bl \citep{S_Ren_Faster}} &\textbf{98.93} & 92.12 & 95.41 & - & - & - \\
		& Faster R-CNN (7356C) {\bl \citep{S_Ren_Faster}} & 95.61 & 89.83 & 92.63 & 88.85 & 83.48 & 86.08 \\
		& YOLOv3 (DetOnly) {\bl \citep{J_Redmon_Yolov3}} & 93.94 & \textbf{98.25} & \textbf{96.05} & - & - & - \\
		& YOLOv3 (7356C) {\bl \citep{J_Redmon_Yolov3}} & 89.56 & 92.16 & 90.84 & 66.32 & 68.24 & 67.26\\
		\hline 
		Weak & \textbf{PageNet (Ours)} & 95.72 & 94.91 & 95.31 & \textbf{90.89} & \textbf{90.12} &\textbf{90.50}\\
		\hline
	\end{tabular*}
\end{table*}

\subsection{Character-level Detection and Recognition}
Because the annotations of ICDAR13 and CASIA-HWDB2.0-2.2 contain the bounding boxes of characters, we can compare the character-level detection and recognition results of our approach with two representative object detection methods, namely Faster R-CNN \citep{S_Ren_Faster} and YOLOv3 \citep{J_Redmon_Yolov3}. These two methods are trained using CASIA-SR and fully annotated CASIA-HWDB2.0-2.2, whereas PageNet is trained using CASIA-SR and weakly annotated CASIA-HWDB2.0-2.2. 

In Table \ref{TBL_CLSR}, there are two versions of Faster R-CNN and YOLOv3. The one marked with DetOnly is trained to only detect characters, whereas the other marked with 7356C needs to classify 7,356 categories of characters in addition. There are also two sets of evaluation metrics. The one denoted as DetOnly evaluates the character detection regardless of the classification, whereas the other denoted as 7356C requires not only the detection is accurate but also the classification is correct. All evaluation metrics are calculated with an IoU threshold of 0.5. 

As shown in Table \ref{TBL_CLSR}, PageNet achieves better DetOnly and 7356C performances than Faster R-CNN (7356C) and YOLOv3 (7356C). Even compared with Faster R-CNN (DetOnly) and YOLOv3 (DetOnly), PageNet can still achieve comparable DetOnly performance. Note that Faster R-CNN and YOLOv3 are trained with full annotations (containing character bounding boxes), whereas our method is trained under weak supervision (without bounding box annotations). It can be concluded that the decoupled three-branch design of the detection and recognition module can handle the task of character-level detection and recognition very well, especially when the number of categories is very large. Moreover, the proposed weakly supervised learning framework can effectively train the model using only transcript annotations.

\subsection{Visualizations}
\label{sec_vis}
\begin{figure*}[htb]
	\centering 
	\subfigure{
		\begin{minipage}[t]{0.5\columnwidth}
			\centering
			\includegraphics[height=0.63\columnwidth]{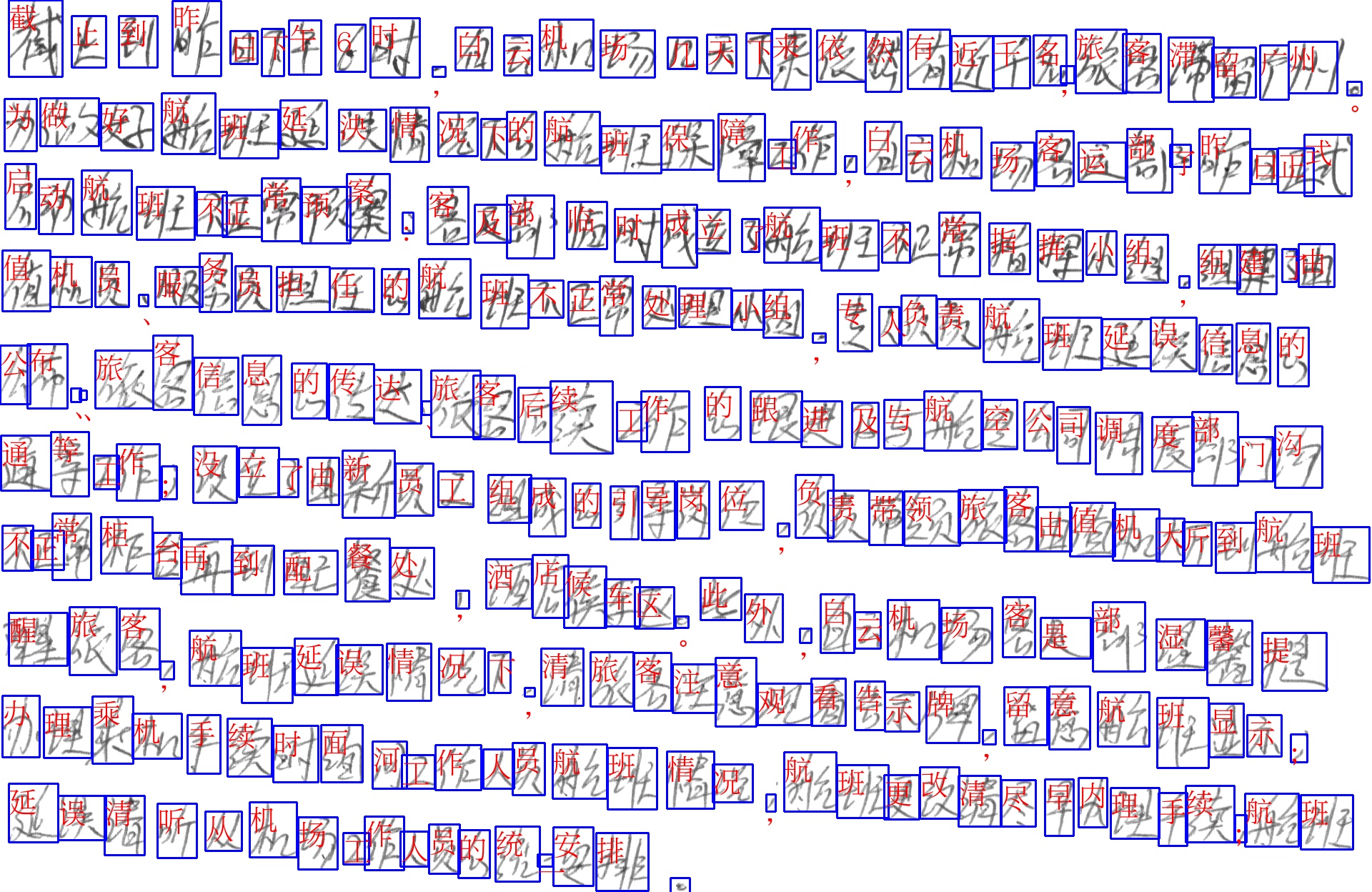}
	\end{minipage}}%
	\subfigure{
		\begin{minipage}[t]{0.5\columnwidth}
			\centering
			\includegraphics[height=0.63\columnwidth]{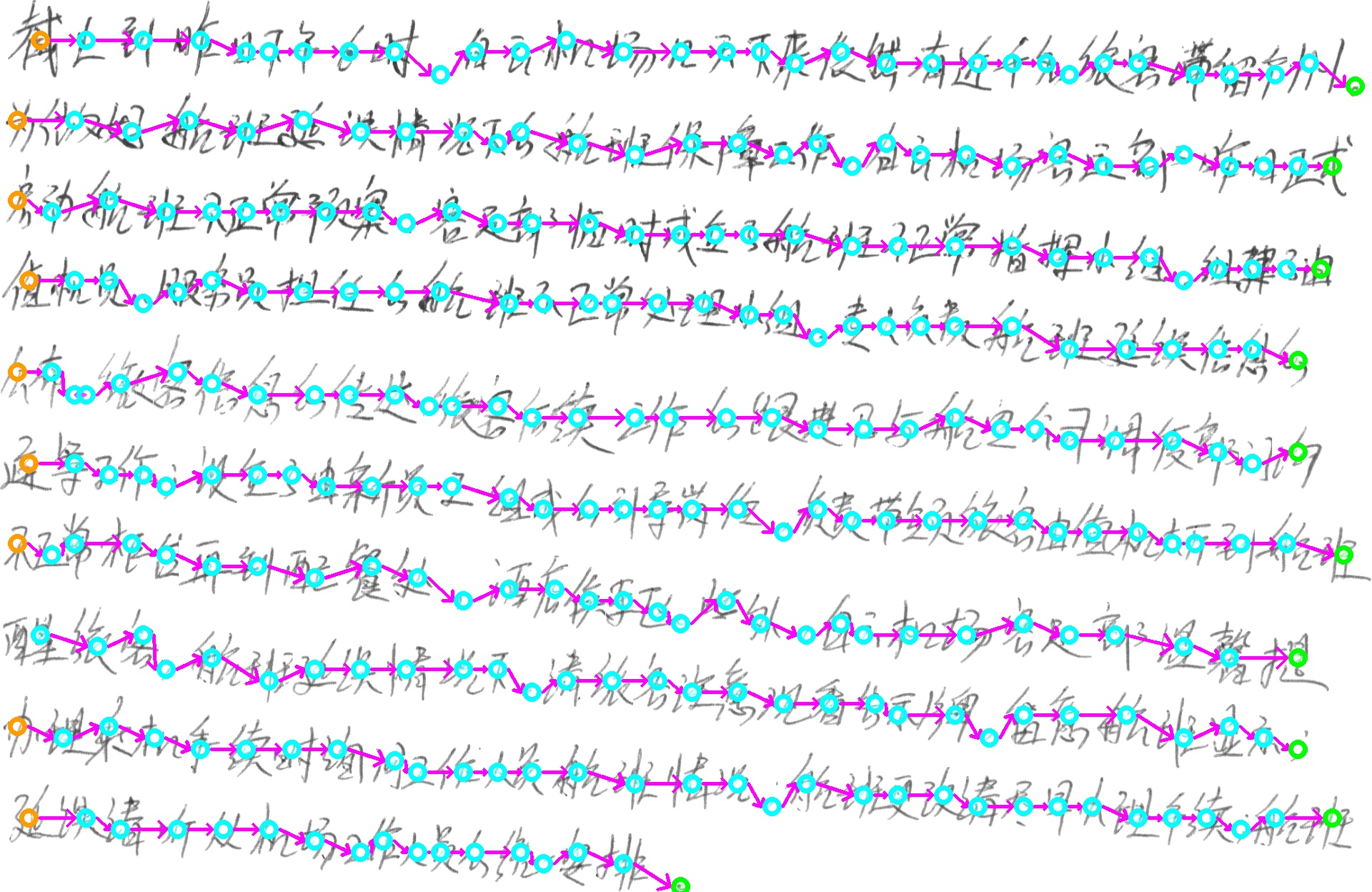}
	\end{minipage}}%
	\subfigure{
		\begin{minipage}[t]{0.5\columnwidth}
			\centering
			\includegraphics[height=0.63\columnwidth]{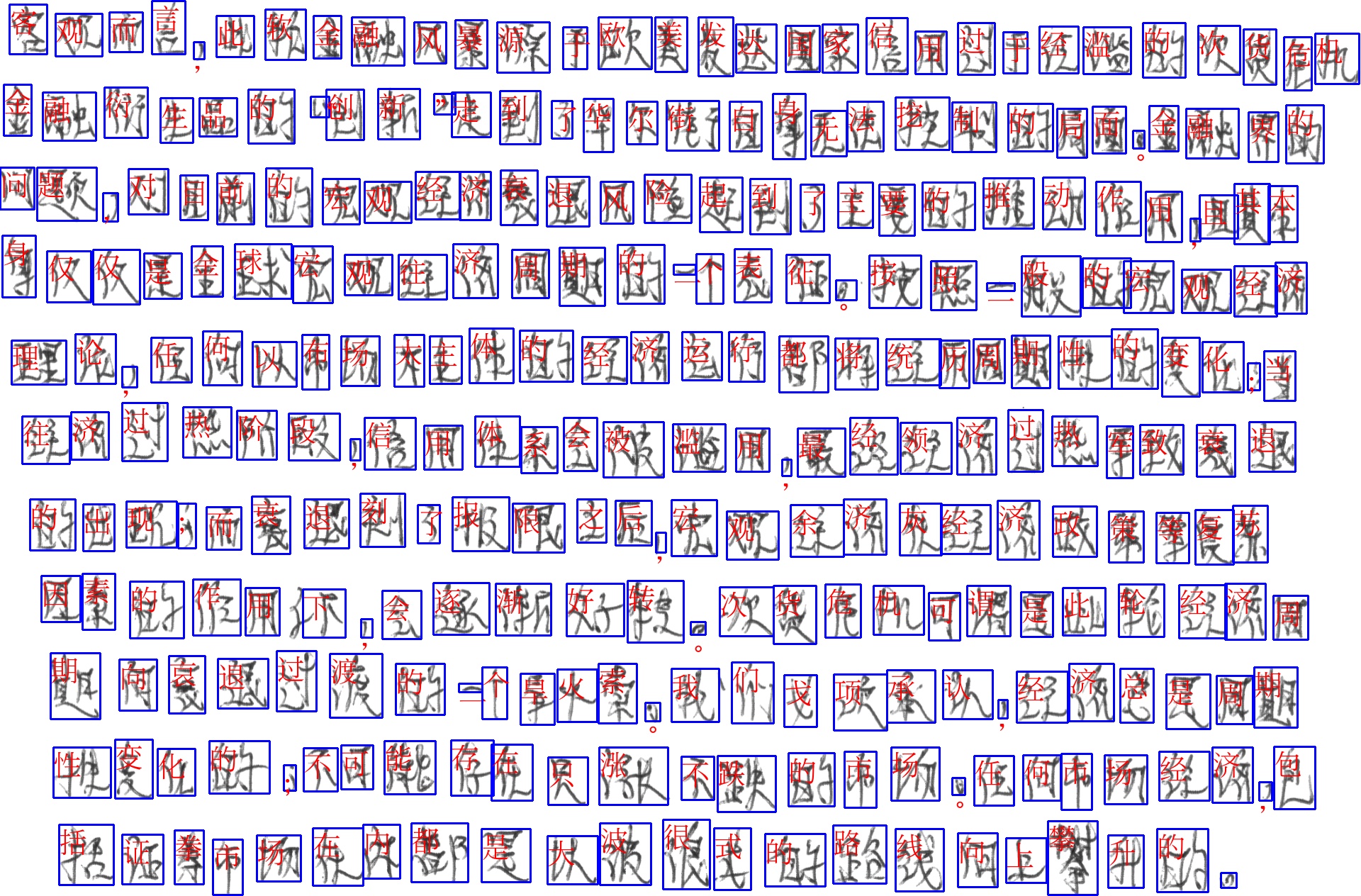}
	\end{minipage}}%
	\subfigure{
		\begin{minipage}[t]{0.5\columnwidth}
			\centering
			\includegraphics[height=0.63\columnwidth]{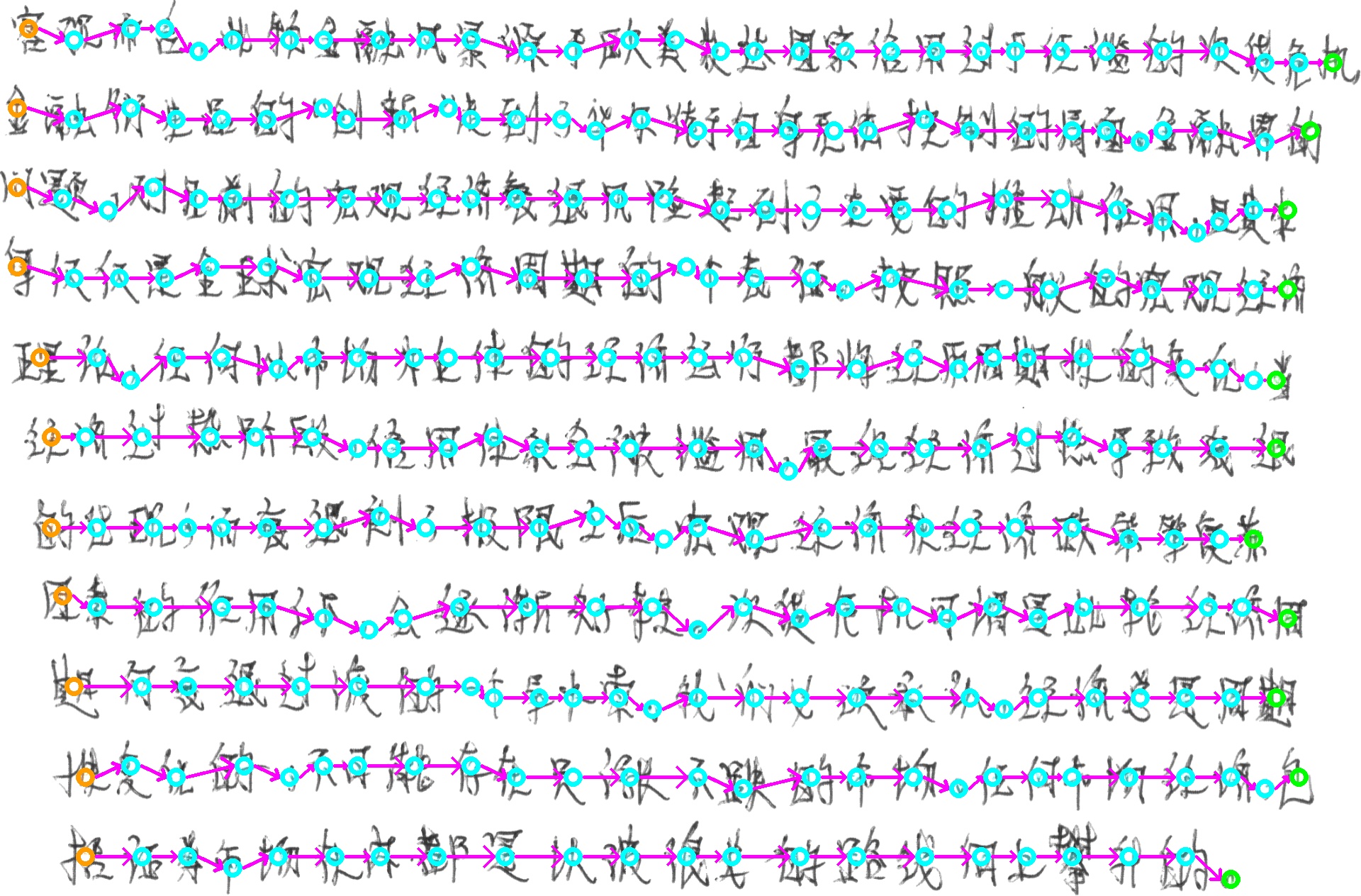}
	\end{minipage}}
	\subfigure{
		\begin{minipage}[t]{0.5\columnwidth}
			\centering
			\includegraphics[height=0.63\columnwidth]{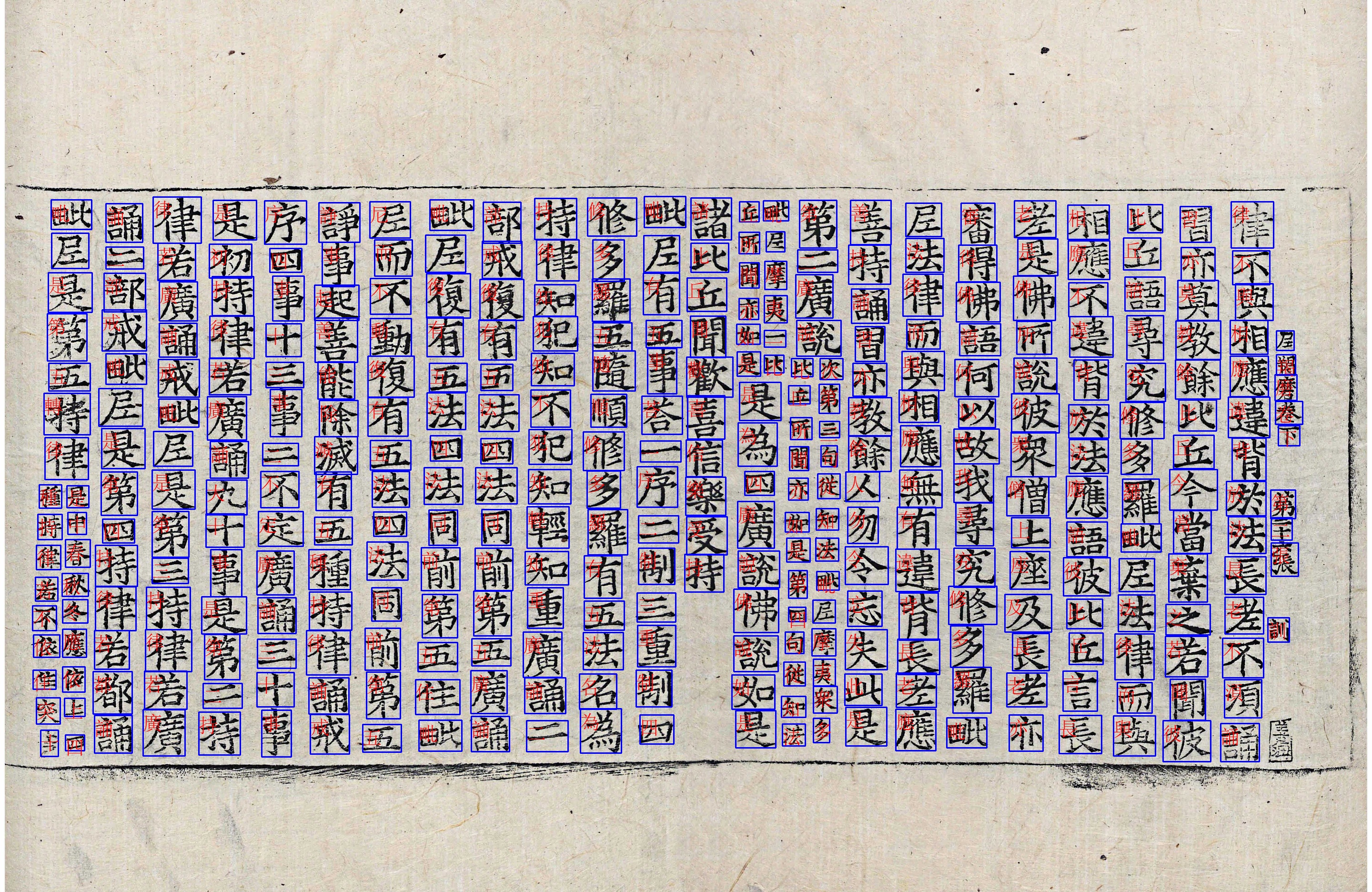}
	\end{minipage}}%
	\subfigure{
		\begin{minipage}[t]{0.5\columnwidth}
			\centering
			\includegraphics[height=0.63\columnwidth]{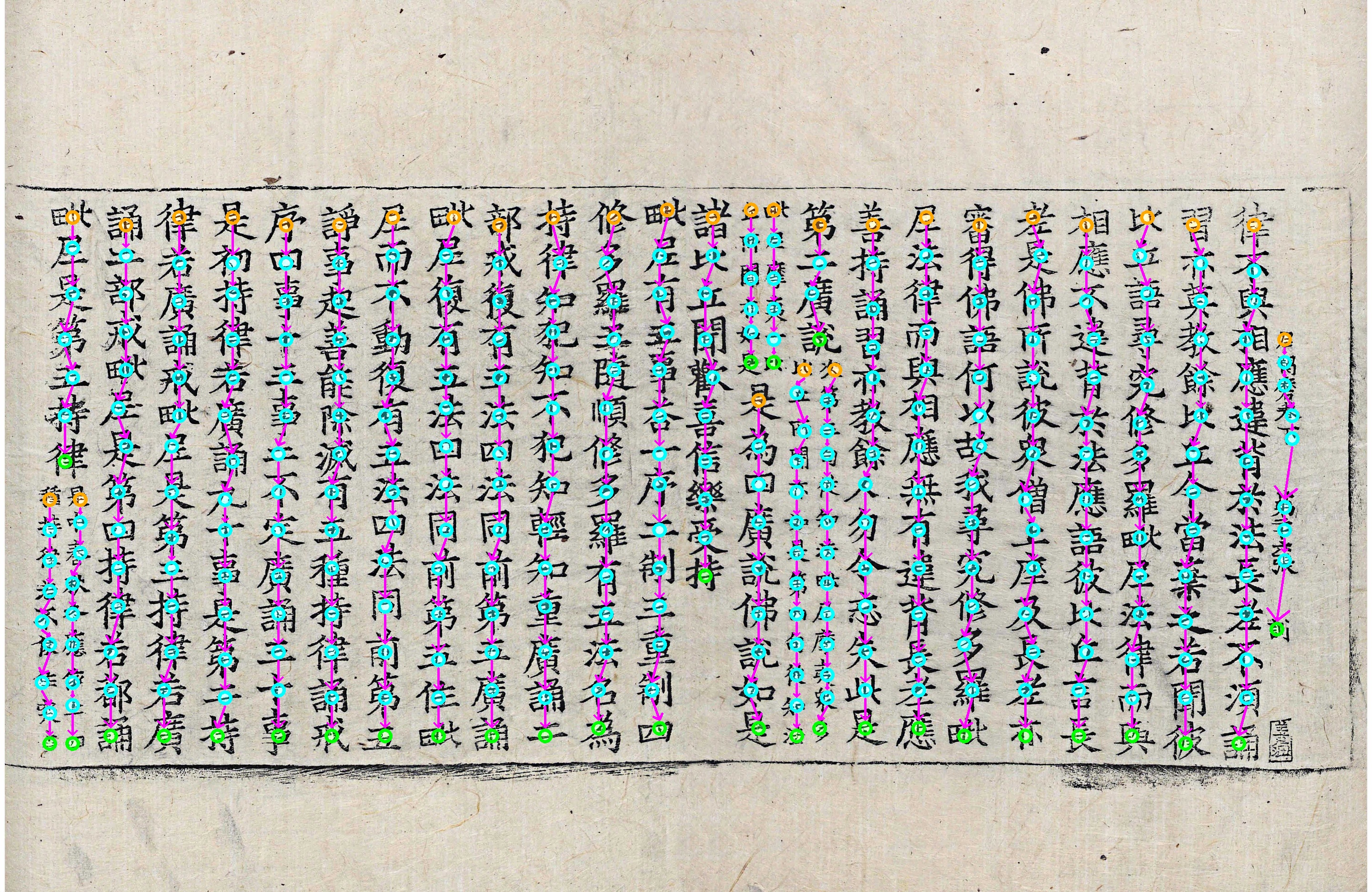}
	\end{minipage}}%
	\subfigure{
		\begin{minipage}[t]{0.25\columnwidth}
			\centering
			\includegraphics[height=1.3\columnwidth]{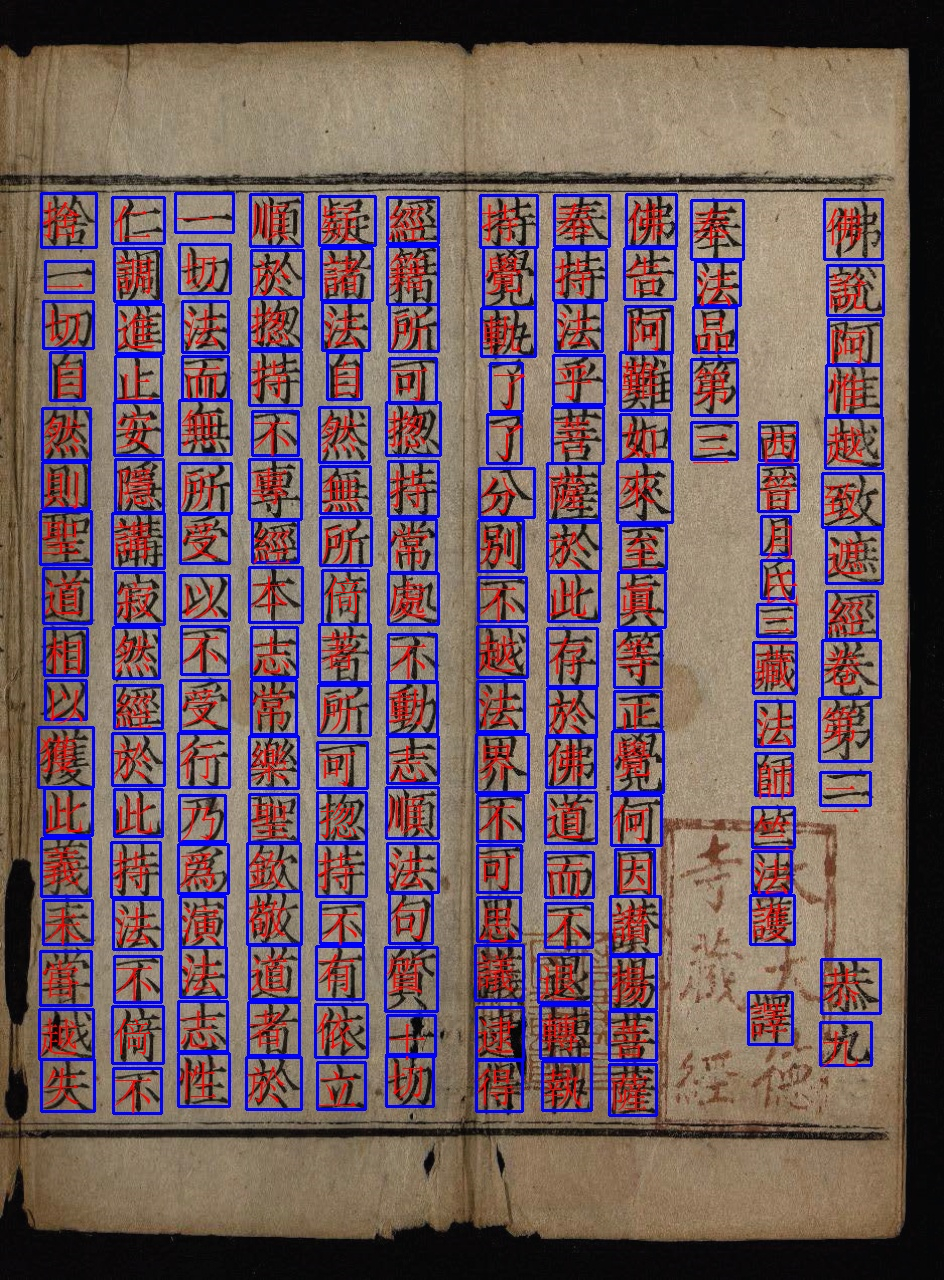}
	\end{minipage}}%
	\subfigure{
		\begin{minipage}[t]{0.25\columnwidth}
			\centering
			\includegraphics[height=1.3\columnwidth]{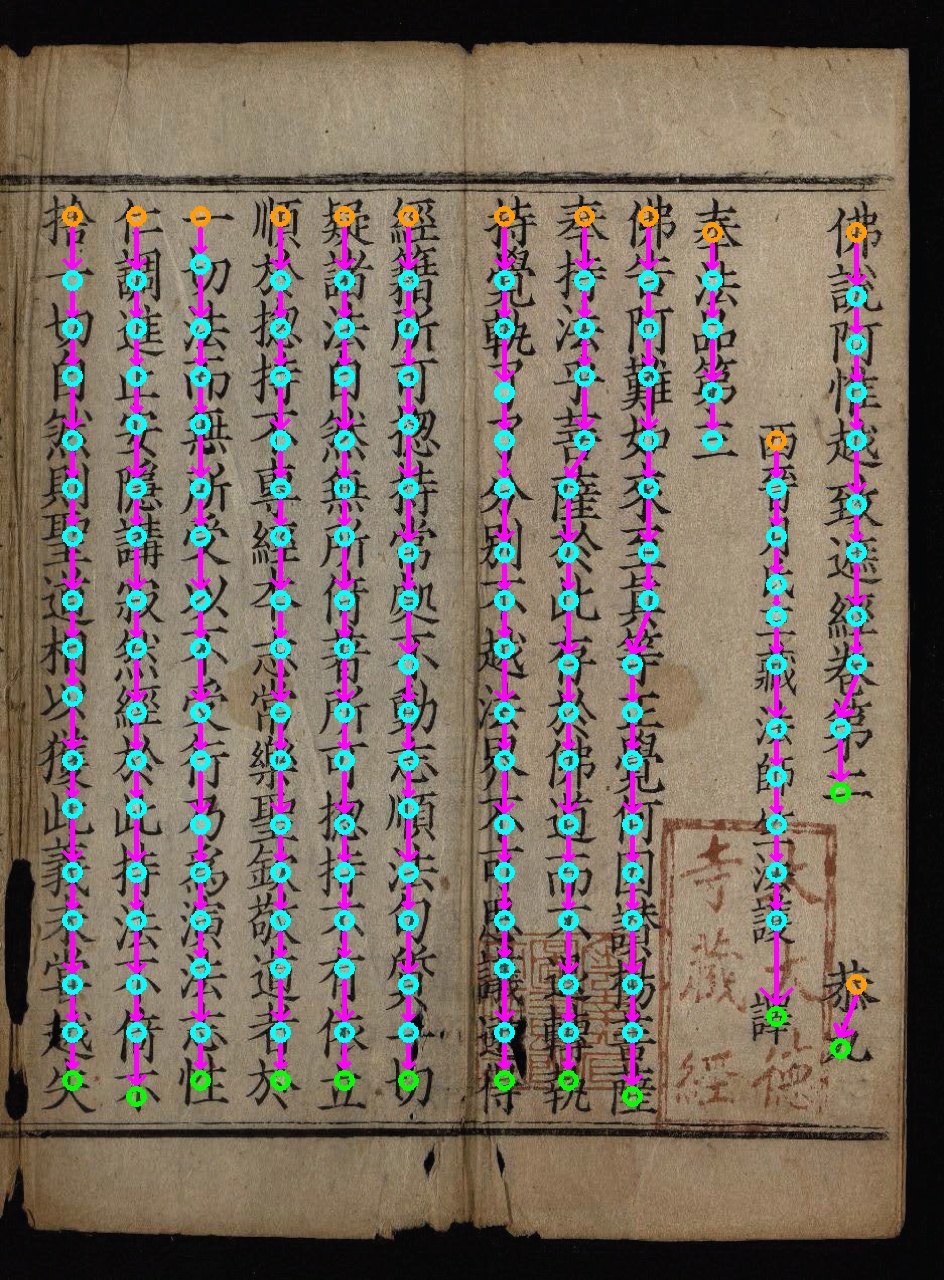}
	\end{minipage}}%
	\subfigure{
		\begin{minipage}[t]{0.25\columnwidth}
			\centering
			\includegraphics[height=1.3\columnwidth]{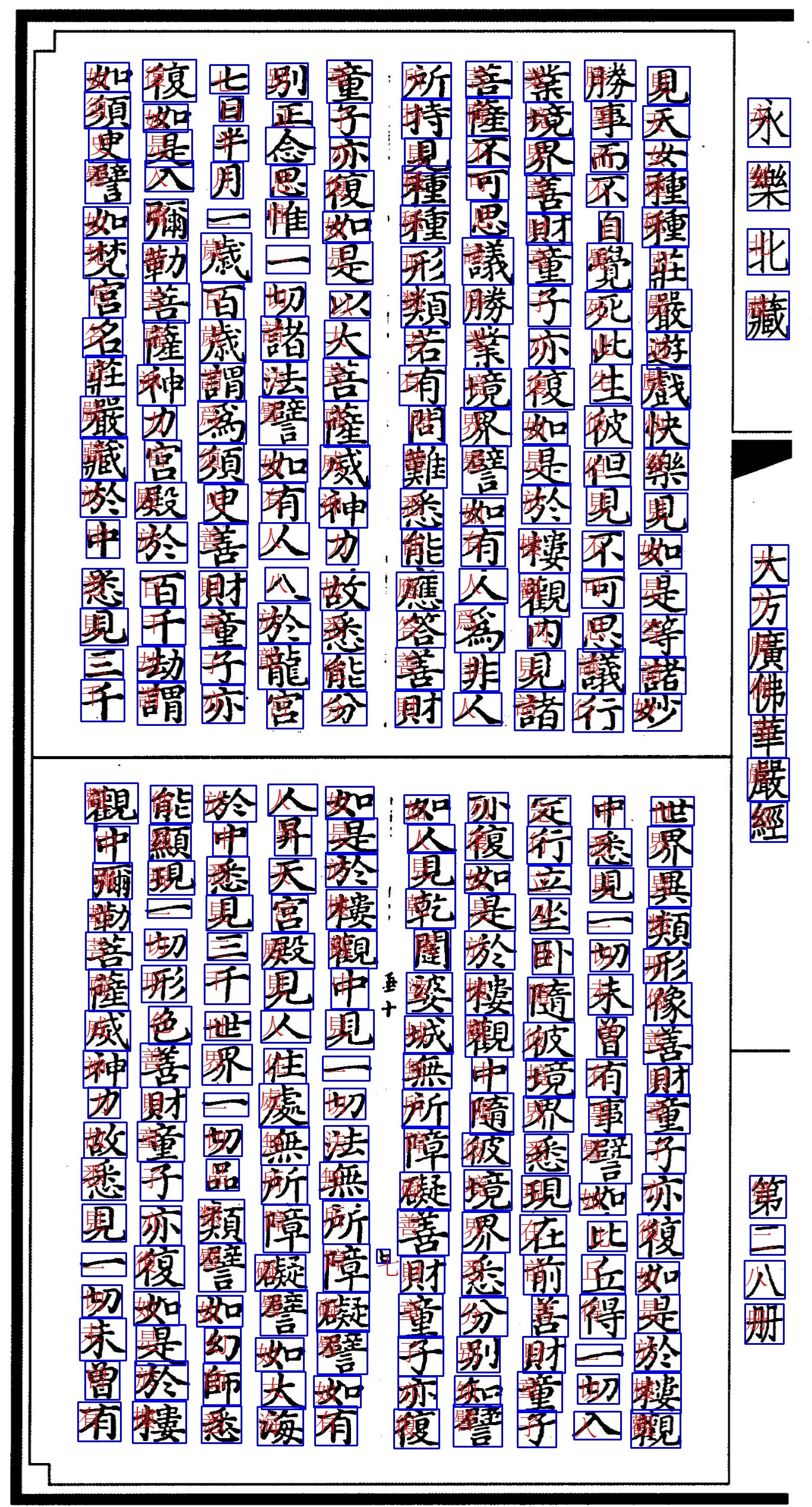}
	\end{minipage}}%
	\subfigure{
		\begin{minipage}[t]{0.25\columnwidth}
			\centering
			\includegraphics[height=1.3\columnwidth]{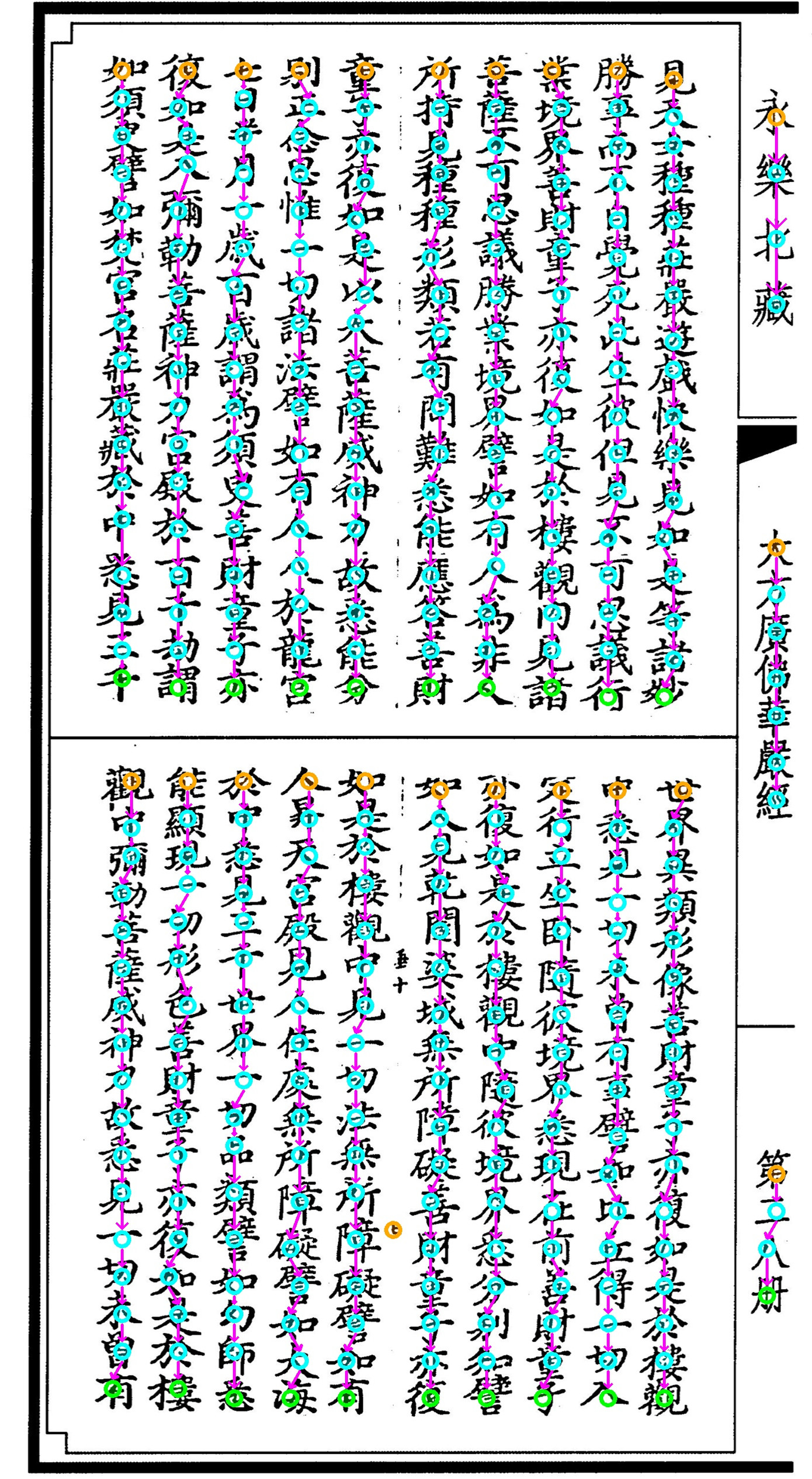}
	\end{minipage}}
	\subfigure{
		\begin{minipage}[c]{0.33\columnwidth}
			\centering
			\includegraphics[height=0.95\columnwidth]{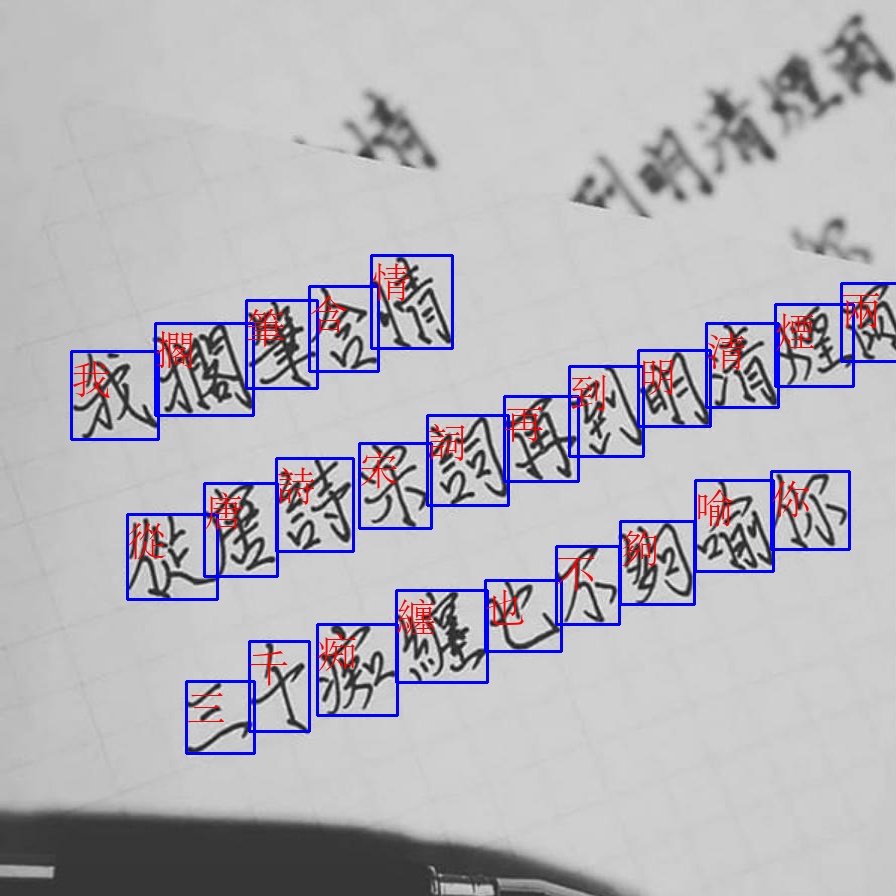}
	\end{minipage}}%
	\subfigure{
		\begin{minipage}[c]{0.33\columnwidth}
			\centering
			\includegraphics[height=0.95\columnwidth]{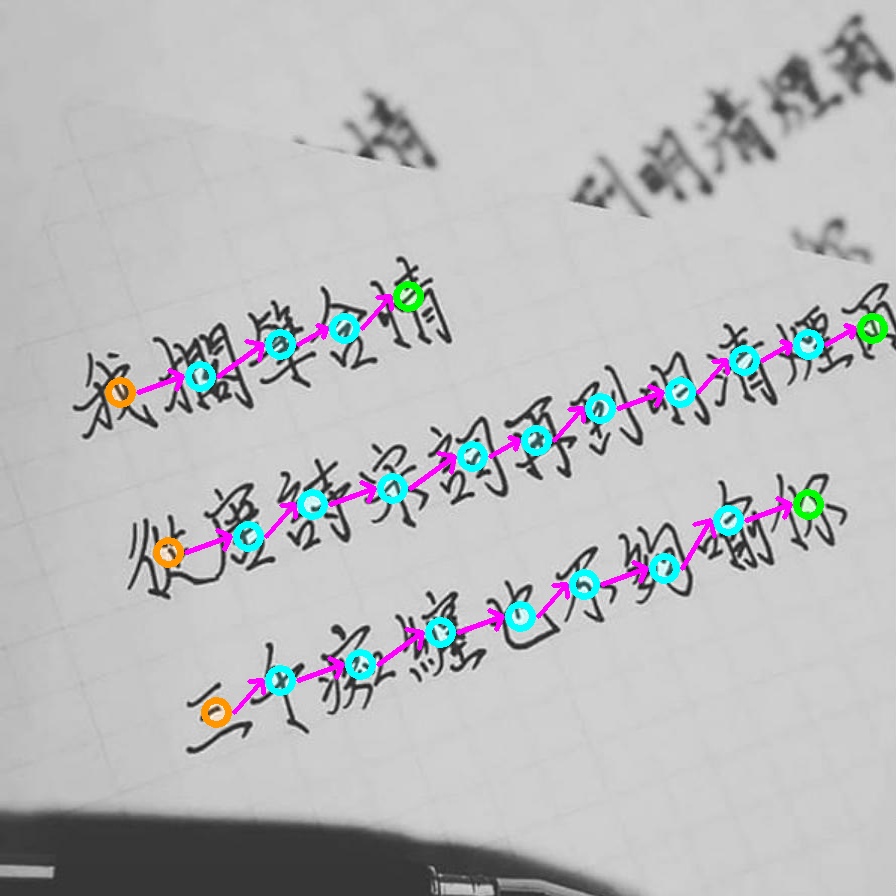}
	\end{minipage}}%
	\subfigure{
		\begin{minipage}[c]{0.33\columnwidth}
			\centering
			\includegraphics[height=1.3\columnwidth]{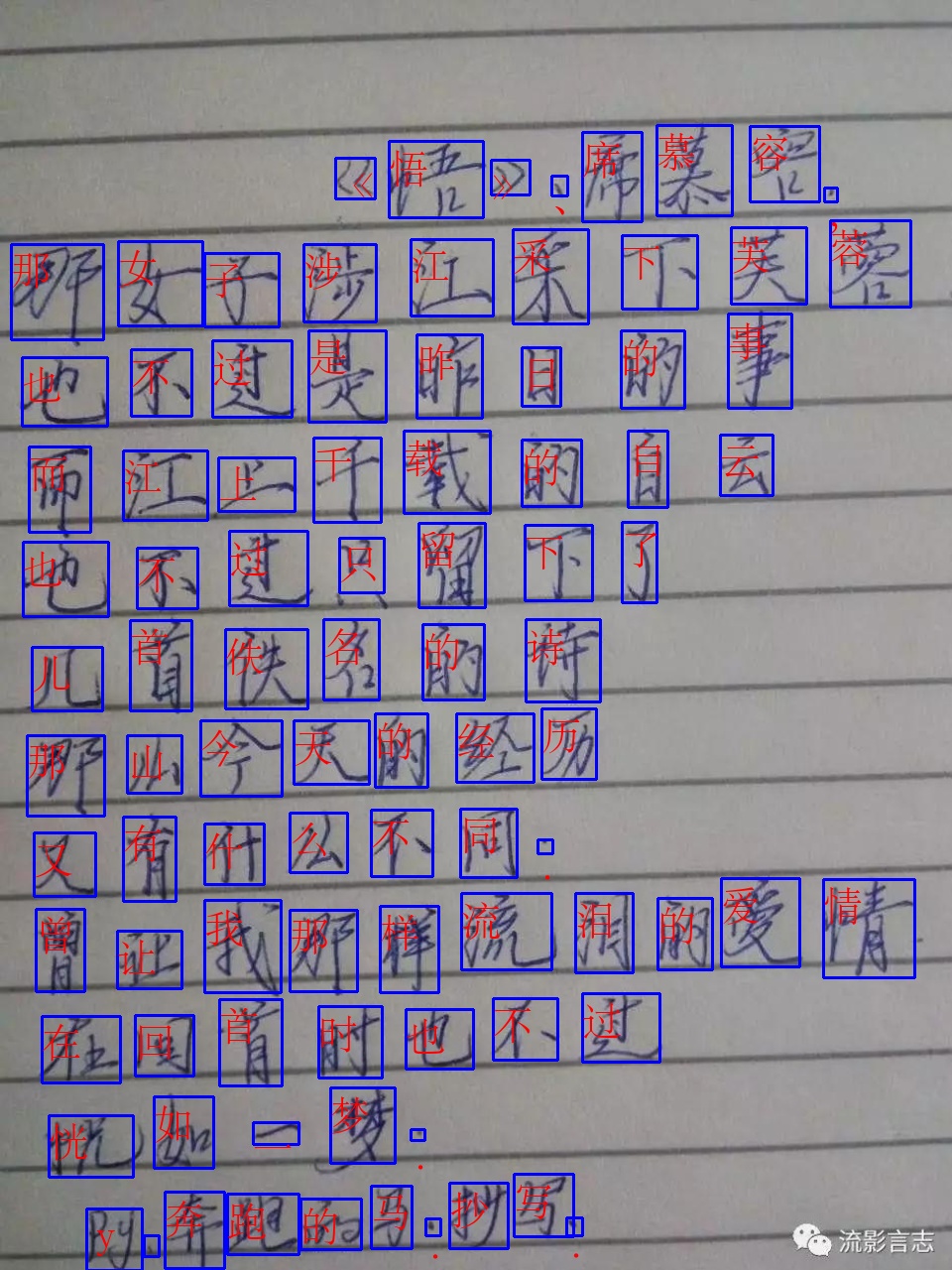}
	\end{minipage}}%
	\subfigure{
		\begin{minipage}[c]{0.33\columnwidth}
			\centering
			\includegraphics[height=1.3\columnwidth]{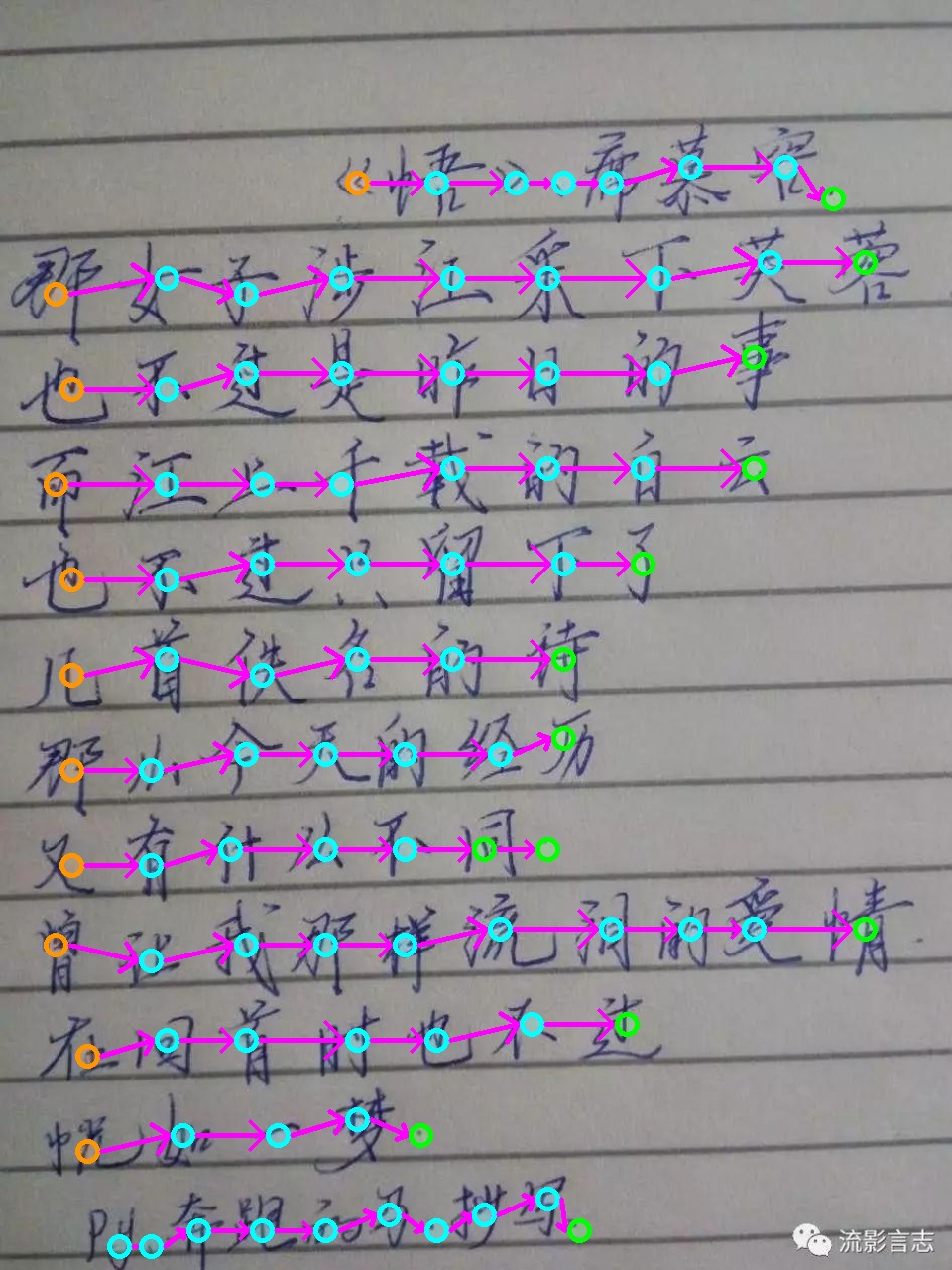}
	\end{minipage}}%
	\subfigure{
		\begin{minipage}[c]{0.33\columnwidth}
			\centering
			\includegraphics[height=1.3\columnwidth]{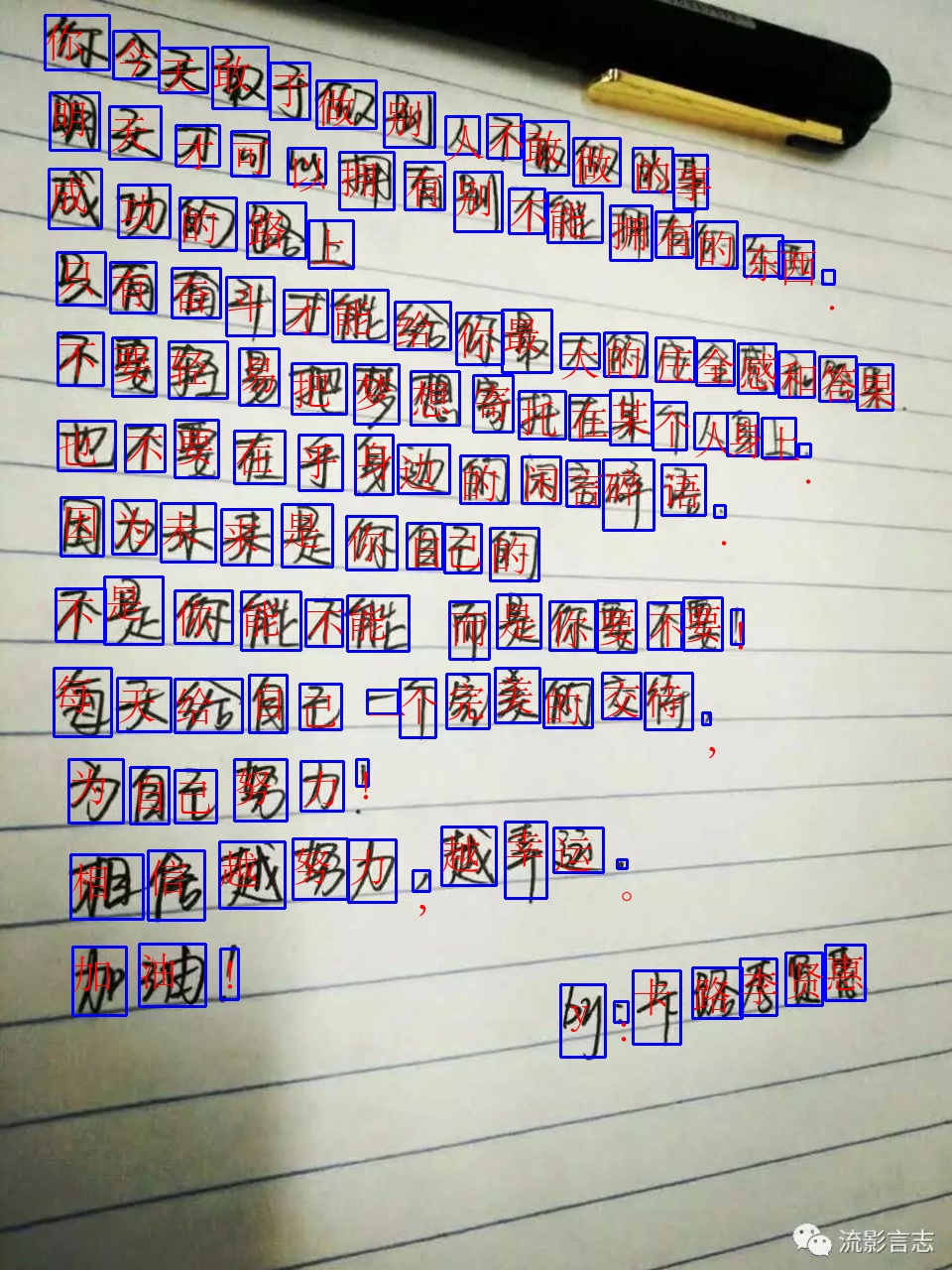}
	\end{minipage}}%
	\subfigure{
		\begin{minipage}[c]{0.33\columnwidth}
			\centering
			\includegraphics[height=1.3\columnwidth]{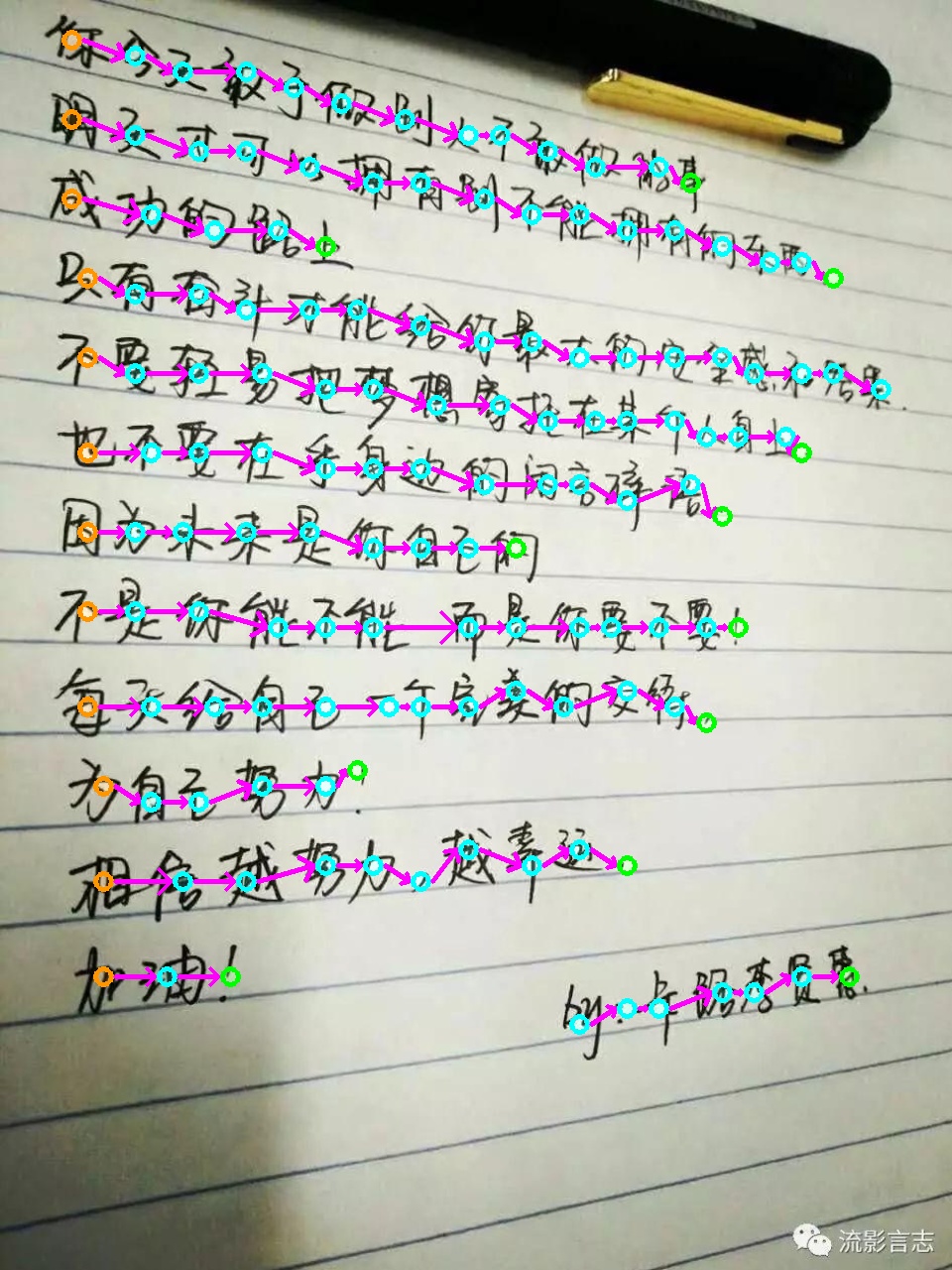}
	\end{minipage}}
	\subfigure{
		\begin{minipage}[t]{0.33\columnwidth}
			\centering
			\includegraphics[height=1.75\columnwidth]{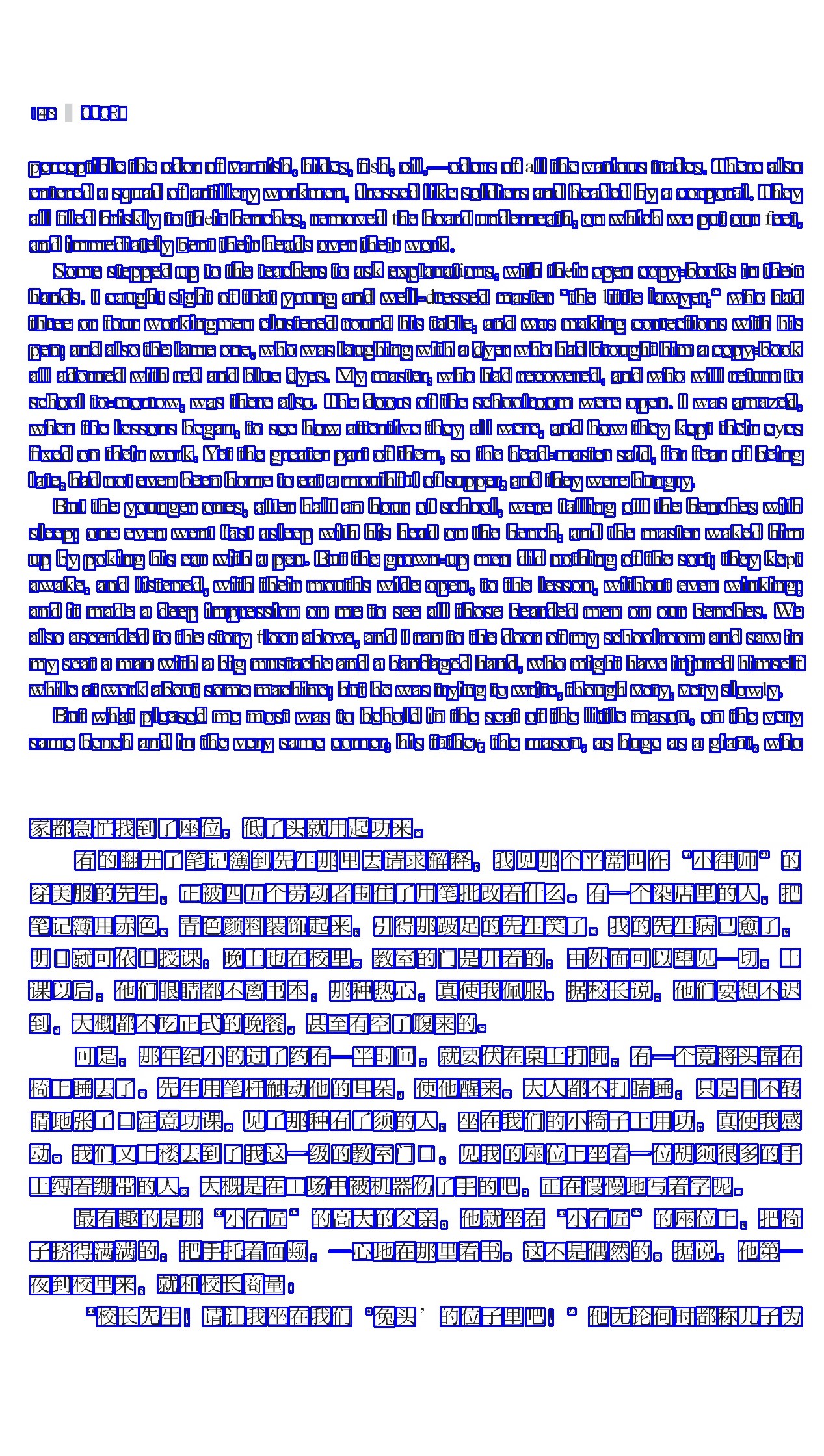}
	\end{minipage}}%
	\subfigure{
		\begin{minipage}[t]{0.33\columnwidth}
			\centering
			\includegraphics[height=1.75\columnwidth]{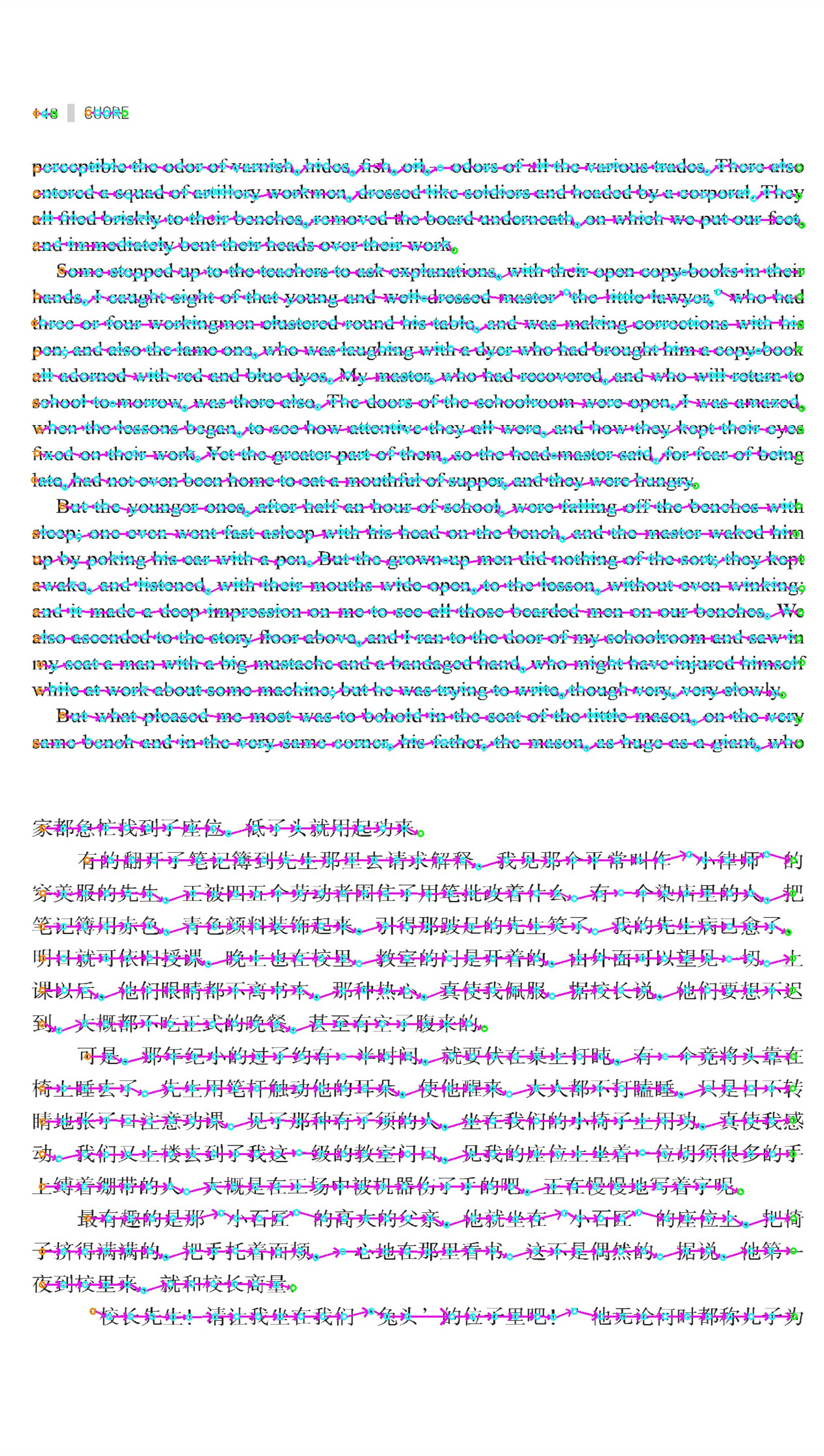}
	\end{minipage}}%
	\subfigure{
		\begin{minipage}[t]{0.33\columnwidth}
			\centering
			\includegraphics[height=1.75\columnwidth]{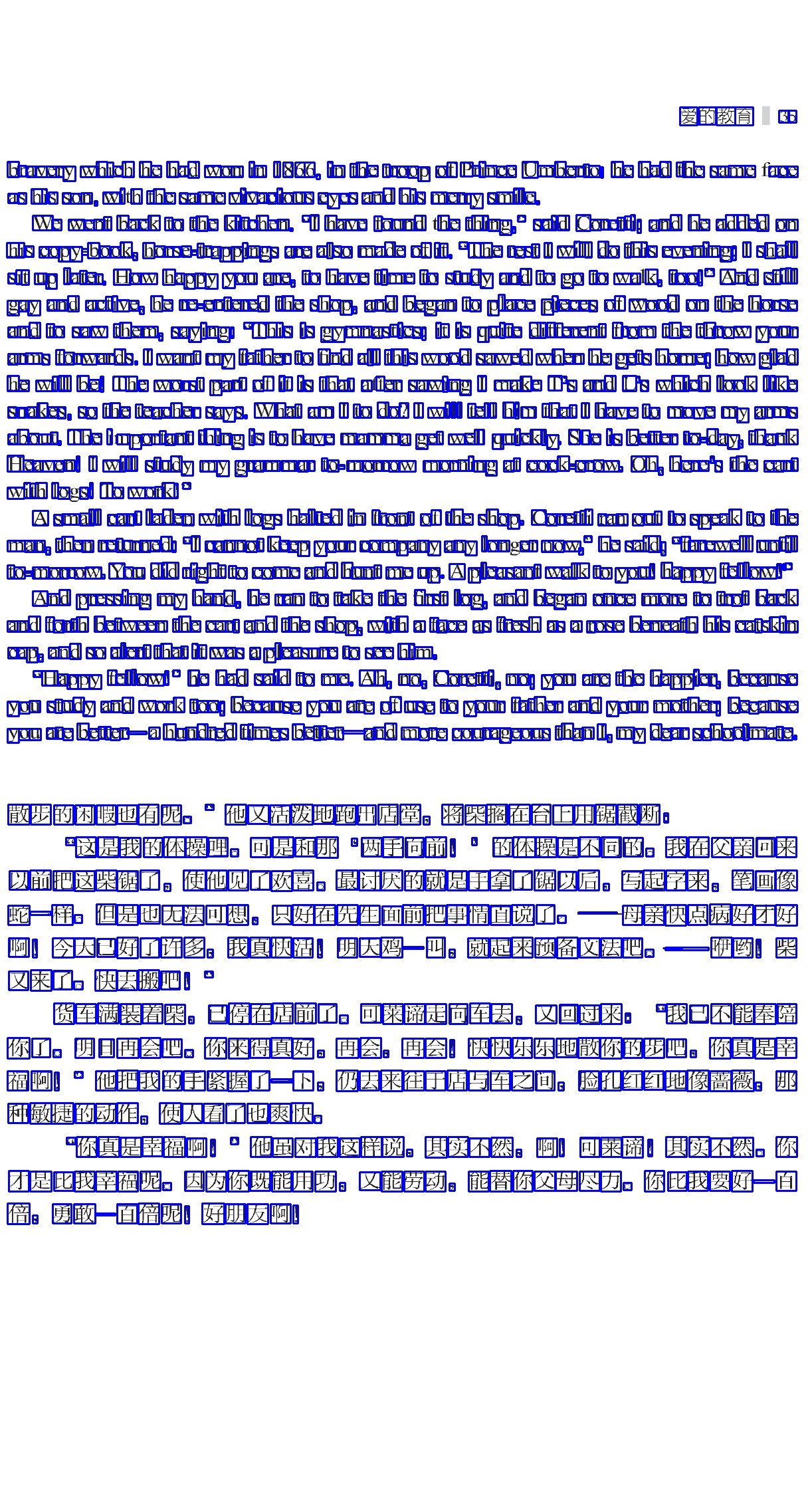}
	\end{minipage}}%
	\subfigure{
		\begin{minipage}[t]{0.33\columnwidth}
			\centering
			\includegraphics[height=1.75\columnwidth]{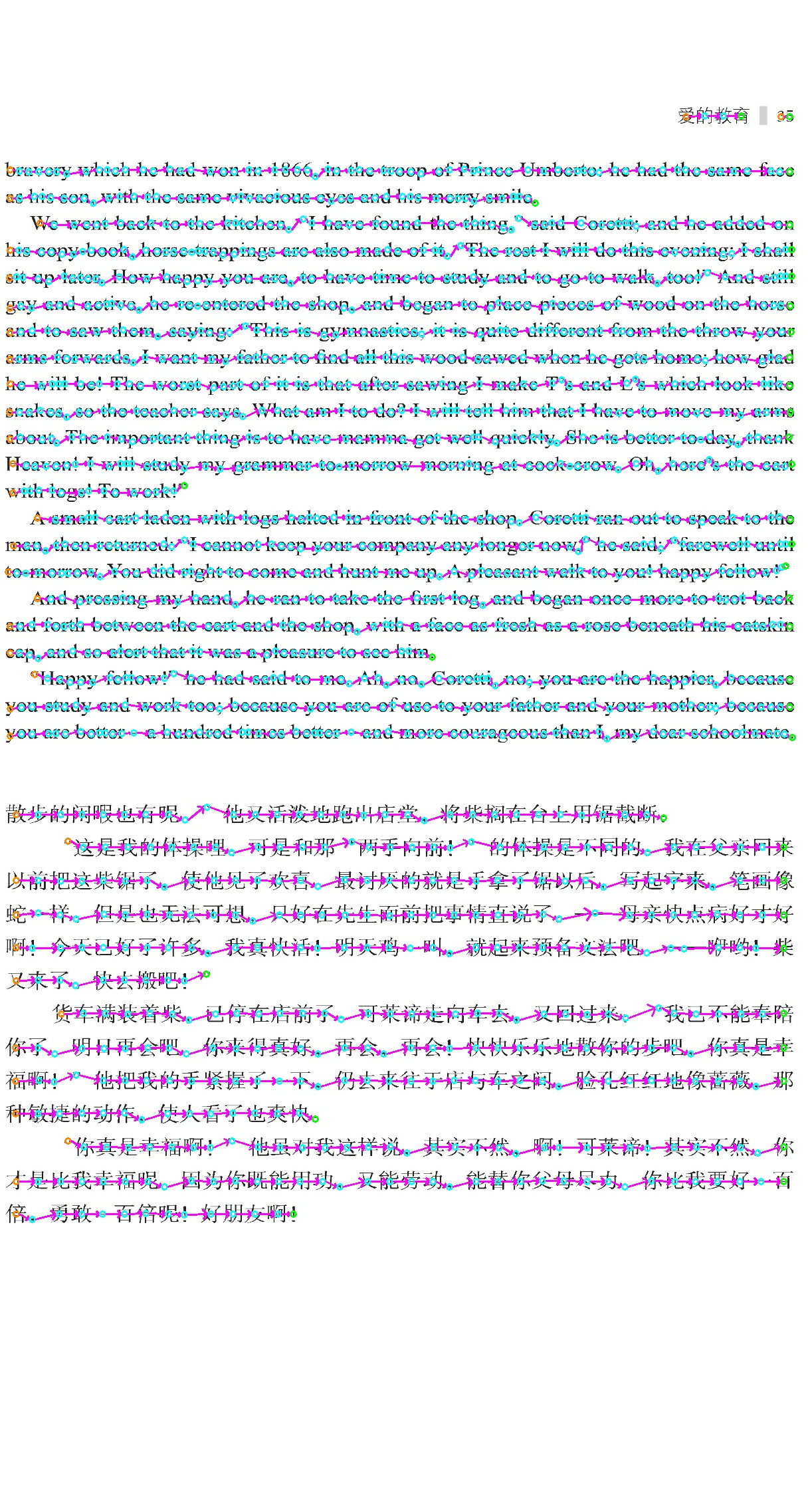}
	\end{minipage}}%
	\subfigure{
		\begin{minipage}[t]{0.33\columnwidth}
			\centering
			\includegraphics[height=1.75\columnwidth]{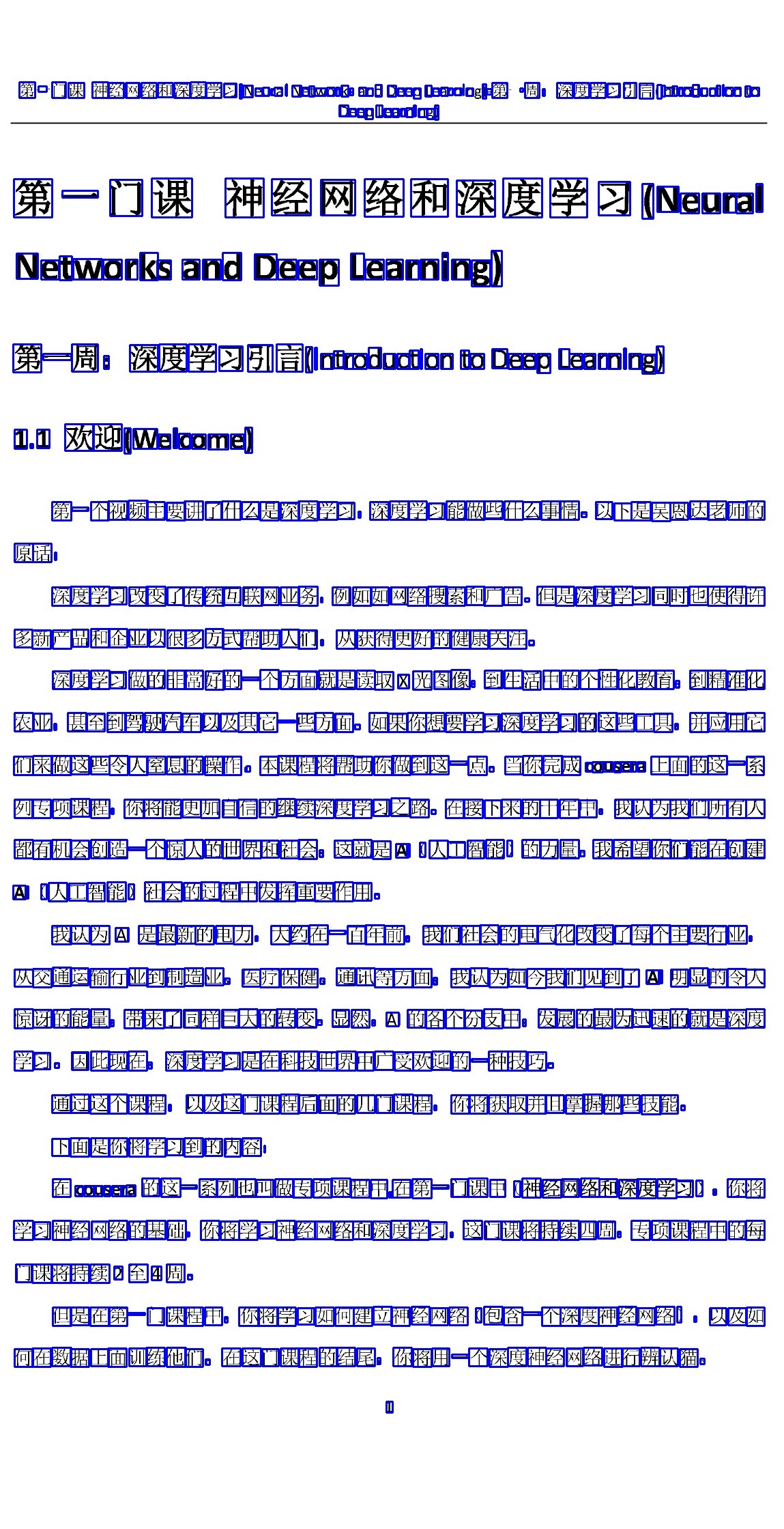}
	\end{minipage}}%
	\subfigure{
		\begin{minipage}[t]{0.33\columnwidth}
			\centering
			\includegraphics[height=1.75\columnwidth]{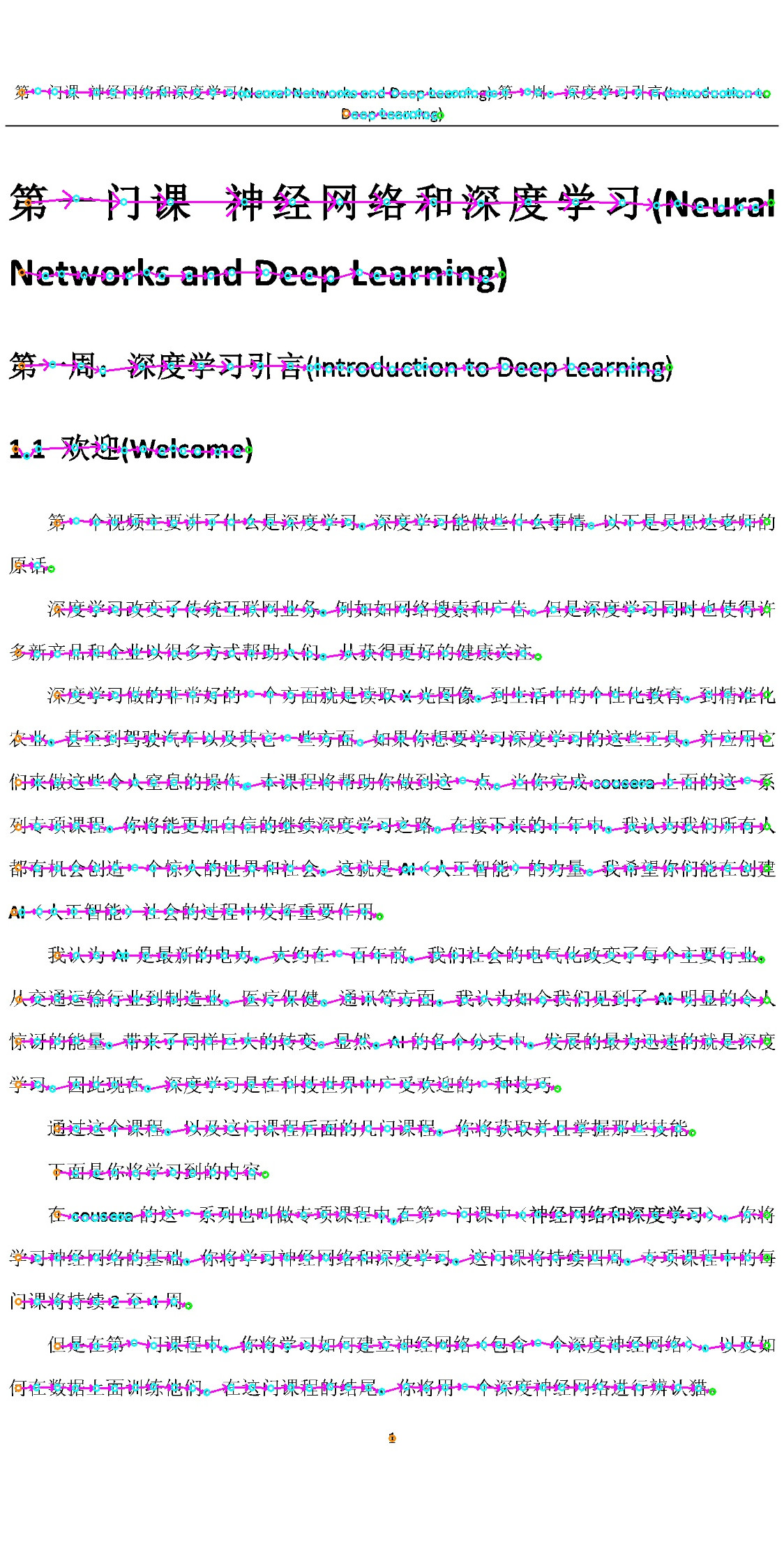}
	\end{minipage}}%
	{\bl \caption{Visualization results of PageNet. For each pair of images, the left is the character detection and recognition results and the right is the predicted reading order. In the visualization of the reading order, each circle represents a character ({\bl orange} circle: start-of-line; green circle: end-of-line). From the first row to the fourth row, the images are from ICDAR13, MTHv2, SCUT-HCCDoc, and JS-SCUT PrintCC, respectively. In the last row, the character recognition results of the images from JS-SCUT PrintCC are not visualized, because the characters are too small and densely distributed. Zoom in for a better view.}	\label{Fig_Exp_Vis}}
\end{figure*}

The visualization results are shown in Fig. \ref{Fig_Exp_Vis}. 
It can be seen that the character detection and recognition results and the reading order can be accurately predicted, although there is no bounding box annotation {\bl for real data}.

The synthetic images of MTHv2 and SCUT-HCCDoc are synthesized by simply placing characters from font files on simple backgrounds. However, the visualization results demonstrate that our method learns to handle diverse handwriting styles, complex backgrounds and layouts, various perspective transformations, and uneven illuminations in real samples through the proposed weakly supervised learning framework.
As shown in the visualization results from JS-SCUT PrintCC, our method can also process multilingual texts including both Chinese and English.

\begin{table*}[t]
	\centering 
	\caption{Comparison of different matching and updating algorithms on ICDAR13}
	\label{TBL_Abla_MUF}
	\begin{tabular*}{\hsize}{@{}@{\extracolsep{\fill}}llll@{}}
		\hline
		Part & Algorithm & AR* & CR* \\
		\hline 
		\multirow{2}*{Semantic Matching} & Page-level Length {\bl \citep{L_Xing_Convolutional,Y_Baek_Character}} & 15.18 & 15.54 \\
		& Line-level Length {\bl \citep{L_Xing_Convolutional,Y_Baek_Character}} & 41.90 & 44.70 \\
		\hline 
		Spatial Matching & No Spatial Matching & 91.62 & 92.11 \\
		\hline 
		\multirow{3}*{Updating} & Replace {\bl \citep{L_Xing_Convolutional,Y_Baek_Character,C_Wigington_Start}} & 12.24 & 64.53 \\
		& Average & 91.58 & 92.28 \\
		& Fixed Ratio & 91.51 & 92.17 \\
		\hline 
		\multicolumn{2}{c}{\textbf{Ours}} & \textbf{92.83} & \textbf{93.23} \\
		\hline 
	\end{tabular*}
\end{table*}

\subsection{Experiments on Weakly Supervised Learning}
\label{sec_exp_WSL}

\subsubsection{Effectiveness of Semantic Matching}

In Sec. \ref{Sec_Matching}, semantic matching is proposed to find reliable character-level results, which consists of line matching based on AR and character matching based on edit distance. However, some existing methods \citep{L_Xing_Convolutional,Y_Baek_Character} simply use the length of predictions to determine the reliable results. 
Therefore, we compare our semantic matching with two algorithms following their ideas, which are \textit{Page-level Length} and \textit{Line-level Length}.

Specifically, in \textit{Page-level Length}, the predicted results are regarded as reliable if the total number of characters on a page in the predictions is equal to that in the annotations. For \textit{Line-level Length}, because the bounding box annotations of text lines required by \citep{L_Xing_Convolutional,Y_Baek_Character} are not used in our method, we first perform line matching using Algorithm \ref{Alg_Line_Matching}. Then, a line-level result is viewed as reliable if the lengths of it and the matched line-level transcript are equal.

The results of our approach and the other two algorithms are presented in Table \ref{TBL_Abla_MUF}. It can be seen that our semantic matching outperforms both \textit{Page-level Length} and \textit{Line-level Length} by a large margin.

\begin{figure}[b]
	\centering
	\subfigure[Image]{
		\label{Fig_Exp_Filter_Ori}
		\begin{minipage}[t]{0.33\columnwidth}
			\centering
			\includegraphics[width=0.8\columnwidth]{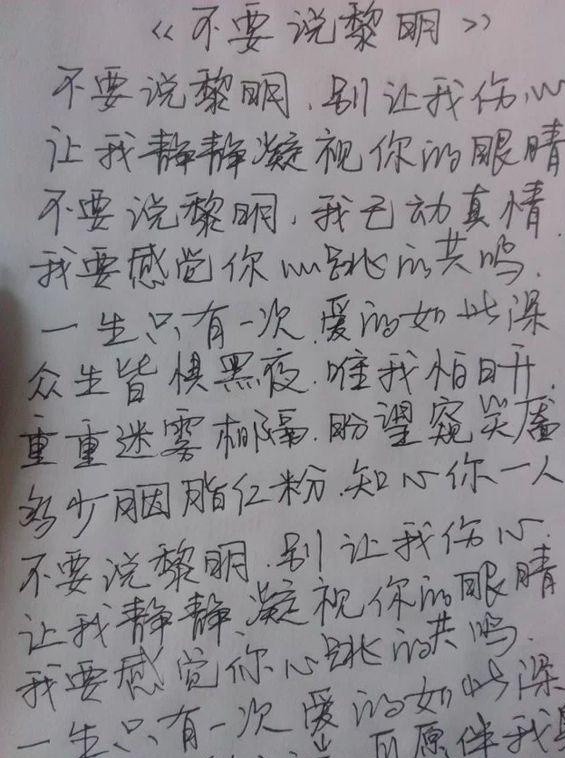}
	\end{minipage}}%
	\subfigure[No spatial matching]{
		\label{Fig_Exp_Filter_No_Filtering}
		\begin{minipage}[t]{0.33\columnwidth}
			\centering
			\includegraphics[width=0.8\columnwidth]{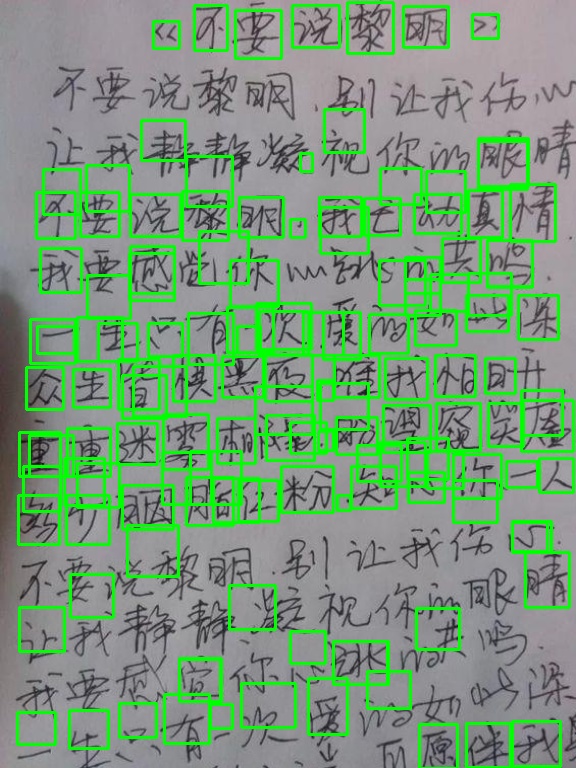}
	\end{minipage}}%
	\subfigure[Spatial matching]{
		\label{Fig_Exp_Filter_Filtering}
		\begin{minipage}[t]{0.33\columnwidth}
			\centering
			\includegraphics[width=0.8\columnwidth]{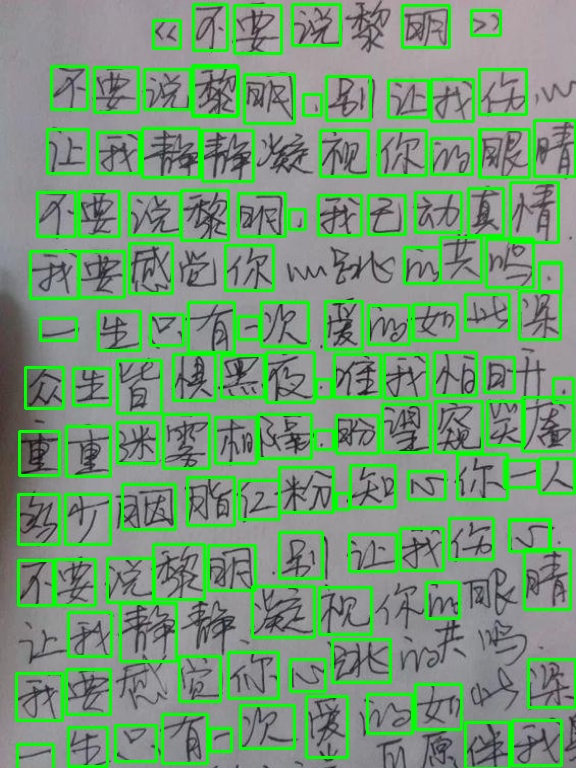}
	\end{minipage}}%
	\caption{(a): Original image. (b): The visualization of pseudo-labels without spatial matching. (c): The visualization of pseudo-labels with spatial matching.}
	\label{Fig_Exp_Filter}
\end{figure}

\subsubsection{Effectiveness of Spatial Matching}

Spatial matching is proposed to solve the matching ambiguity of semantic matching as described in Sec. \ref{Sec_Matching}. As shown in Table \ref{TBL_Abla_MUF}, the performance decreases from 92.83\% to 91.62\% when spatial matching is removed.

We also provide qualitative results of spatial matching in Fig. \ref{Fig_Exp_Filter}. The page in Fig. \ref{Fig_Exp_Filter_Ori} is from SCUT-HCCDoc and contains several lines with identical or similar contents. When conducting semantic matching, owing to the lack of location information in the annotations, one line-level result at the same location may be matched to different line-level transcripts with similar contents at different iterations. Then, the updating algorithm will cause inaccurate pseudo-labels, as shown in Fig. \ref{Fig_Exp_Filter_No_Filtering}. Spatial matching can prevent incorrect matches and result in accurate pseudo-labels, as shown in Fig. \ref{Fig_Exp_Filter_Filtering}.

\subsubsection{Effectiveness of Updating}

To verify the effectiveness of our updating algorithm, we compare it with three algorithms denoted as \textit{Replace}, \textit{Average}, and \textit{Fixed Ratio}, which replace lines 6-8 of Algorithm \ref{Alg_Update} with Eq. (\ref{Equ_replace}), (\ref{Equ_average}), and (\ref{Equ_fix}), respectively.
\begin{gather}
A_{ps}^{(q,n)} = (x^{(p,m)}, y^{(p,m)}, w^{(p,m)}, h^{(p,m)}), \label{Equ_replace}\\
A_{ps}^{(q,n)}\! =\! \frac{k}{k\!+\!1}\! *\! A_{ps}^{(q,n)}\! +\! \frac{1}{k\!+\!1}\! *\! (x^{(p,m)}\!, y^{(p,m)}\!, w^{(p,m)}\!, h^{(p,m)})\!, \label{Equ_average} \\
A_{ps}^{(q,n)}\! =\! 0.9\! *\! A_{ps}^{(q,n)}\! +\! 0.1\! *\! (x^{(p,m)}\!, y^{(p,m)}\!, w^{(p,m)}\!, h^{(p,m)}), \label{Equ_fix} 
\end{gather}
where $k$ is the number of times $A_{ps}^{(q,n)}$ has been updated. 

\begin{figure}[b]
	\centering 
	\includegraphics[width=1.0\columnwidth]{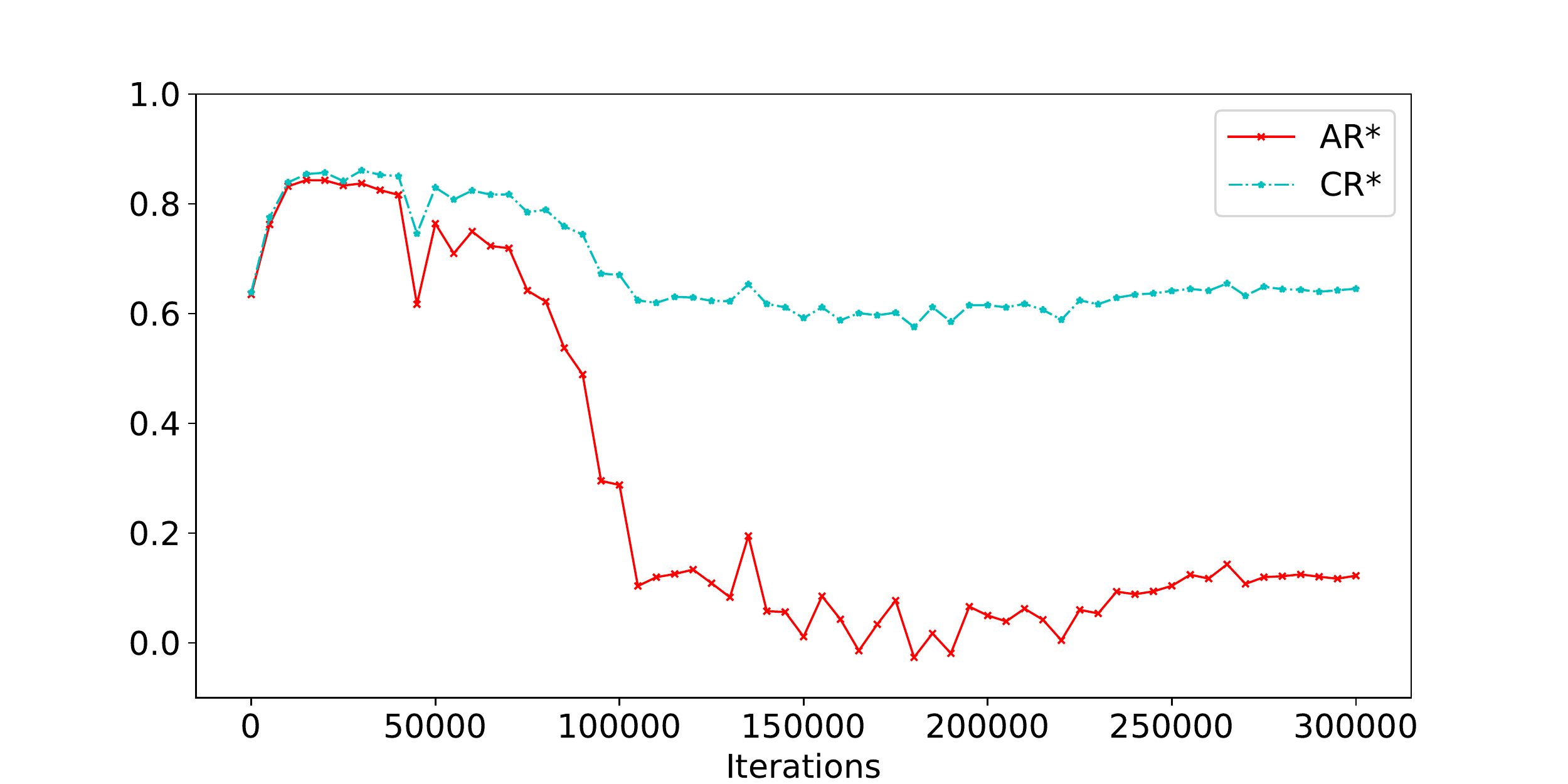}
	\caption{Curves of AR* and CR* on ICDAR13 during the training stage when using the \textit{Replace} algorithm.}
	\label{Fig_Exp_NoIterUpdate}
\end{figure}

The results in Table \ref{TBL_Abla_MUF} show that our updating algorithm achieves the best performance. Especially compared with the \textit{Replace} algorithm which is commonly adopted by previous methods \citep{L_Xing_Convolutional,Y_Baek_Character,C_Wigington_Start}, our updating algorithm exhibits a significant improvement in performance. The \textit{Replace} algorithm directly copies the new matched bounding boxes from the results as the updated pseudo-labels, which makes the model training easily interfered with by poor predictions. Fig. \ref{Fig_Exp_NoIterUpdate} shows the curves of AR* and CR* on ICDAR13 during the training stage when using the \textit{Replace} algorithm. It can be seen that as the training progresses, the performance decreases, eventually converging at 12.24\% AR*.

\subsubsection{Comparison with Fully Supervised Learning}

\begin{table}[t]
	\centering 
	\caption{Performance of PageNet under different supervision}
	\label{TBL_Full}
	\begin{tabular*}{\hsize}{@{}@{\extracolsep{\fill}}lllll@{}}
		\hline
		\multirow{2}*{Supervision} & \multicolumn{2}{l}{ICDAR13} & \multicolumn{2}{l}{MTHv2} \\
		\cline{2-3} \cline{4-5}
		& AR* & CR* & AR* & CR* \\
		\hline
		Full & 91.37 & 91.88 & \textbf{93.81} & \textbf{95.54} \\
		Weak & \textbf{92.83} & \textbf{93.23} & 93.76 & 95.23 \\ 
		\hline
	\end{tabular*}
\end{table}

\begin{figure}[t]
	\centering 
	\includegraphics[width=1.0\columnwidth]{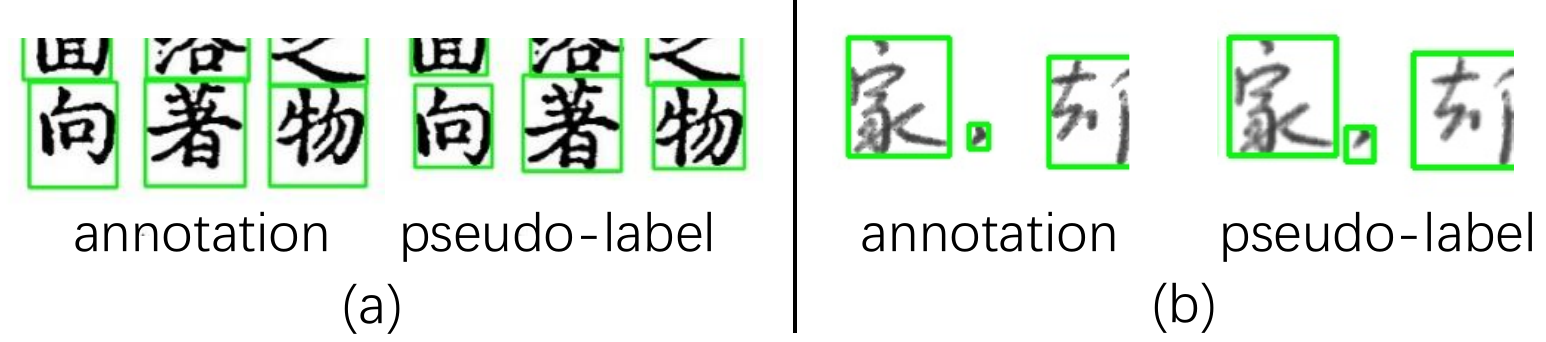}
	\caption{Visualizations of bounding boxes from annotations and pseudo-labels.}
	\label{Compare_WS_Anno_}
\end{figure} 

Because the CASIA-HWDB2.0-2.2 and MTHv2 datasets provide character-level bounding box annotations, we present the results of our method under full supervision in Table \ref{TBL_Full}. Compared with the weakly supervised PageNet, the fully supervised counterpart uses annotated bounding boxes of real samples for loss calculation instead of leveraging the proposed weakly supervised learning framework. As for other details, the models under different supervision share the same settings. 

As shown in Table \ref{TBL_Full}, compared with the fully supervised counterpart, the weakly supervised PageNet achieves better performance on ICDAR13 and comparable performance on MTHv2, 
which demonstrates the success of the proposed weakly supervised learning framework. The automatically generated pseudo-labels can avoid inaccurate and inappropriate bounding boxes in manual annotations. In Fig. \ref{Compare_WS_Anno_}(a), the pseudo-labels are more accurate than the annotations. In Fig. \ref{Compare_WS_Anno_}(b), the bounding box annotations of punctuations in CASIA-HWDB2.0-2.2 are usually very small, which makes it difficult for the detection part to converge, but the pseudo-labels can avoid such inappropriate annotations. Furthermore, the iteratively updated pseudo-labels may serve as data augmentation. Although the images remain unchanged, the annotations are changeable at different iterations, which can improve the robustness of the model.

\subsubsection{Ablation Experiments on Training Strategy}
\label{Sec_Exp_Weakly}
\begin{table}[t]
	\centering 
	\caption{Ablation experiments on the training strategy (evaluated on ICDAR13)}
	\label{TBL_Abla_Pipe_WS}
	\begin{tabular*}{\hsize}{@{}@{\extracolsep{\fill}}lllll@{}}
		\hline 
		Pretraining & Initializing & Training & AR* & CR* \\
		\hline 
		\checkmark & & & 64.94 & 65.27 \\
		& & \checkmark & 80.17 & 85.60 \\
		\checkmark & & \checkmark & 91.57 & 92.15 \\
		\checkmark & \checkmark & \checkmark & \textbf{92.83} & \textbf{93.23} \\
		\hline
	\end{tabular*}
\end{table}
In Table \ref{TBL_Abla_Pipe_WS}, ablation experiments on the training strategy are conducted using ICDAR13. The results show that the proposed training strategy, which consists of pretraining, initializing, and training stages as described in Sec. \ref{sec_training_strategy}, achieves the best performance. {\bl The pretraining and initializing stages are aimed at improving the efficiency of the training stage. Specifically, if the pretraining stage is not adopted, the model at early iterations cannot correctly detect and recognize any characters; thus, the real samples loaded at early iterations are wasted because there is no pseudo-label generated. If the initializing stage is not adopted, the early iterations of the training stage will be wasted on assigning the first round of pseudo-labels to the real samples.}

\begin{table*}[pt]
	\centering 
	\caption{Performance of PageNet on multi-directional texts}
	\label{TBL_MDT}
	\begin{tabular*}{\hsize}{@{}@{\extracolsep{\fill}}lllllllllll@{}}
		\hline
		\multirow{2}*{Supervision} & \multicolumn{2}{l}{0\degree} & \multicolumn{2}{l}{90\degree} & \multicolumn{2}{l}{180\degree} & \multicolumn{2}{l}{270\degree} & \multicolumn{2}{l}{Total} \\
		\cline{2-3} \cline{4-5} \cline{6-7} \cline{8-9} \cline{10-11}
		& AR* & CR* & AR* & CR* & AR* & CR* & AR* & CR* & AR* & CR*  \\
		\hline
		Full & 89.41 & 90.29 & 89.50 & 90.38 & 89.43 & 90.32 & 89.39 & 90.32 & 89.43 & 90.33 \\
		Weak & 89.48 & 90.46 & 89.55 & 90.52 & 89.55 & 90.52 & 89.49 & 90.50 & 89.52 & 90.50 \\
		\hline
	\end{tabular*}
\end{table*}
\subsection{Experiments on Reading Order}
\label{sec_reading_order}

\subsubsection{Multi-directional Reading Order}
\label{sec_exp_multidir}

Most existing methods simply arrange the recognized characters in a text line from left to right, ignoring the complex reading order in the real world. However, taking advantage of the proposed reading order module and graph-based decoding algorithm, our method is able to recognize pages with multi-directional reading order.

\begin{figure}[b]
	\centering 
	\subfigure[0\degree]{
		\begin{minipage}[t]{0.5\columnwidth}
			\centering
			\includegraphics[width=0.9\columnwidth]{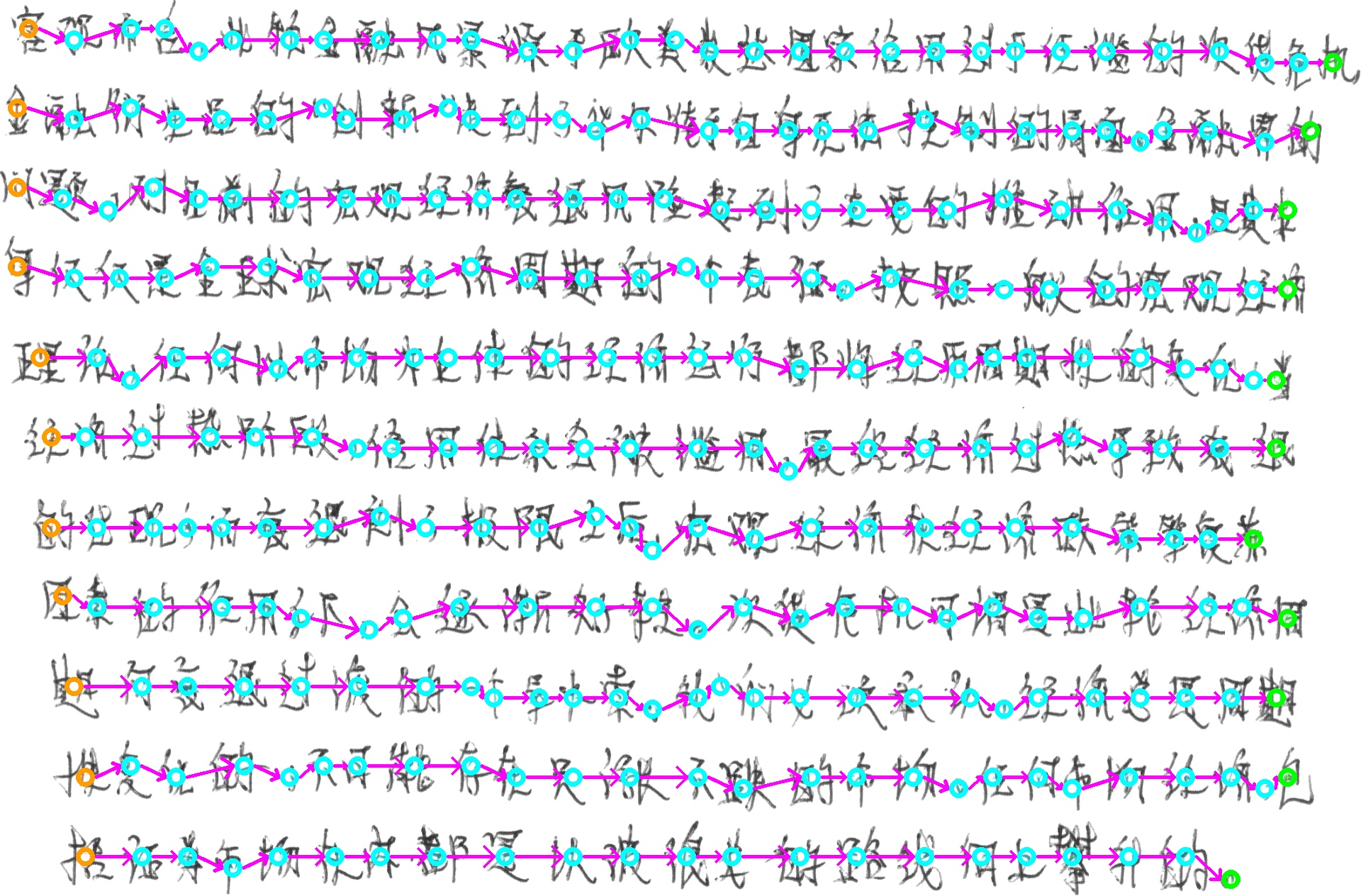}
	\end{minipage}}%
	\subfigure[180\degree]{
		\begin{minipage}[t]{0.5\columnwidth}
			\centering
			\includegraphics[width=0.9\columnwidth]{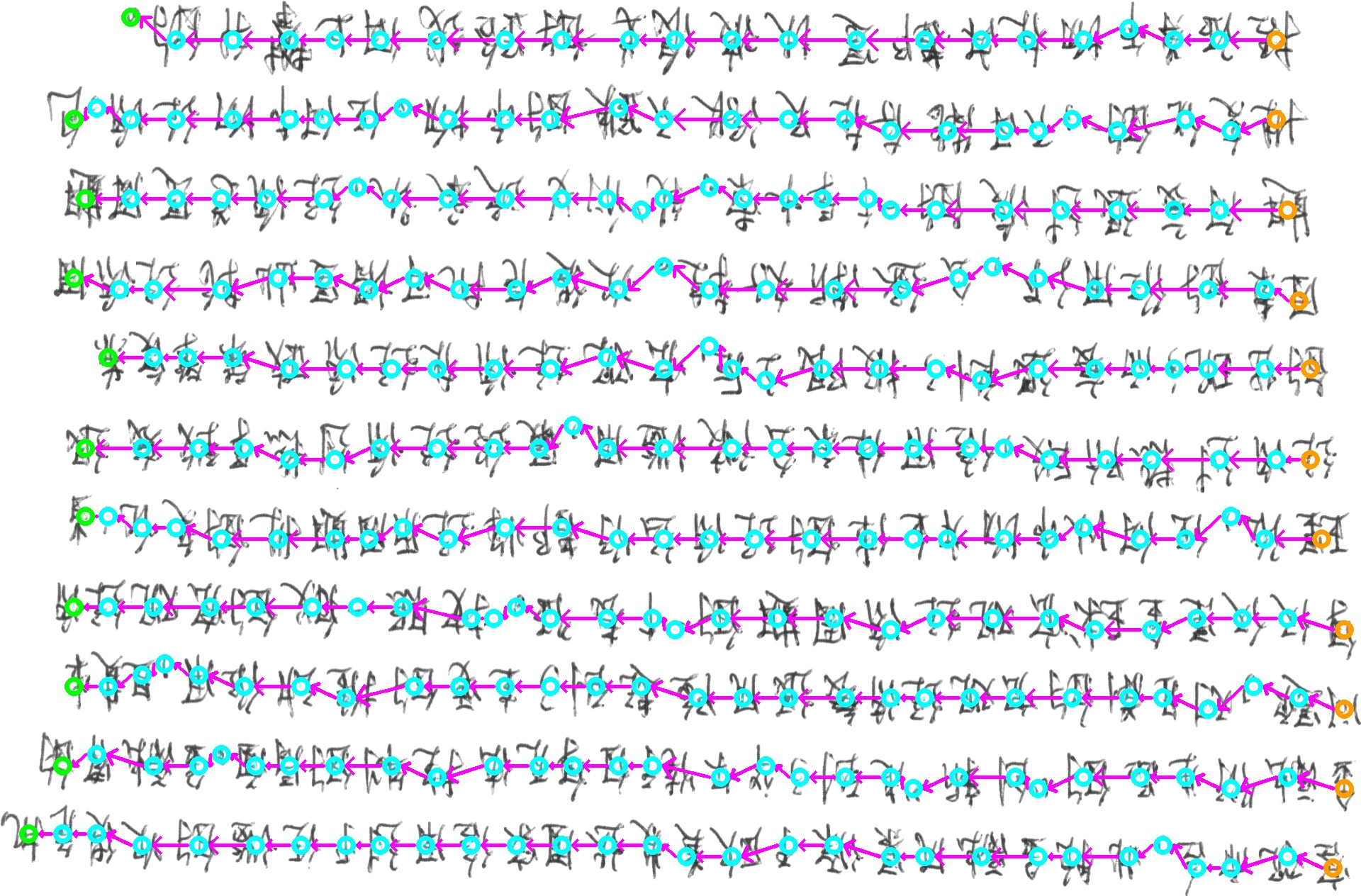}
	\end{minipage}}
	\subfigure[90\degree]{
		\begin{minipage}[t]{0.5\columnwidth}
			\centering
			\includegraphics[width=0.56\columnwidth]{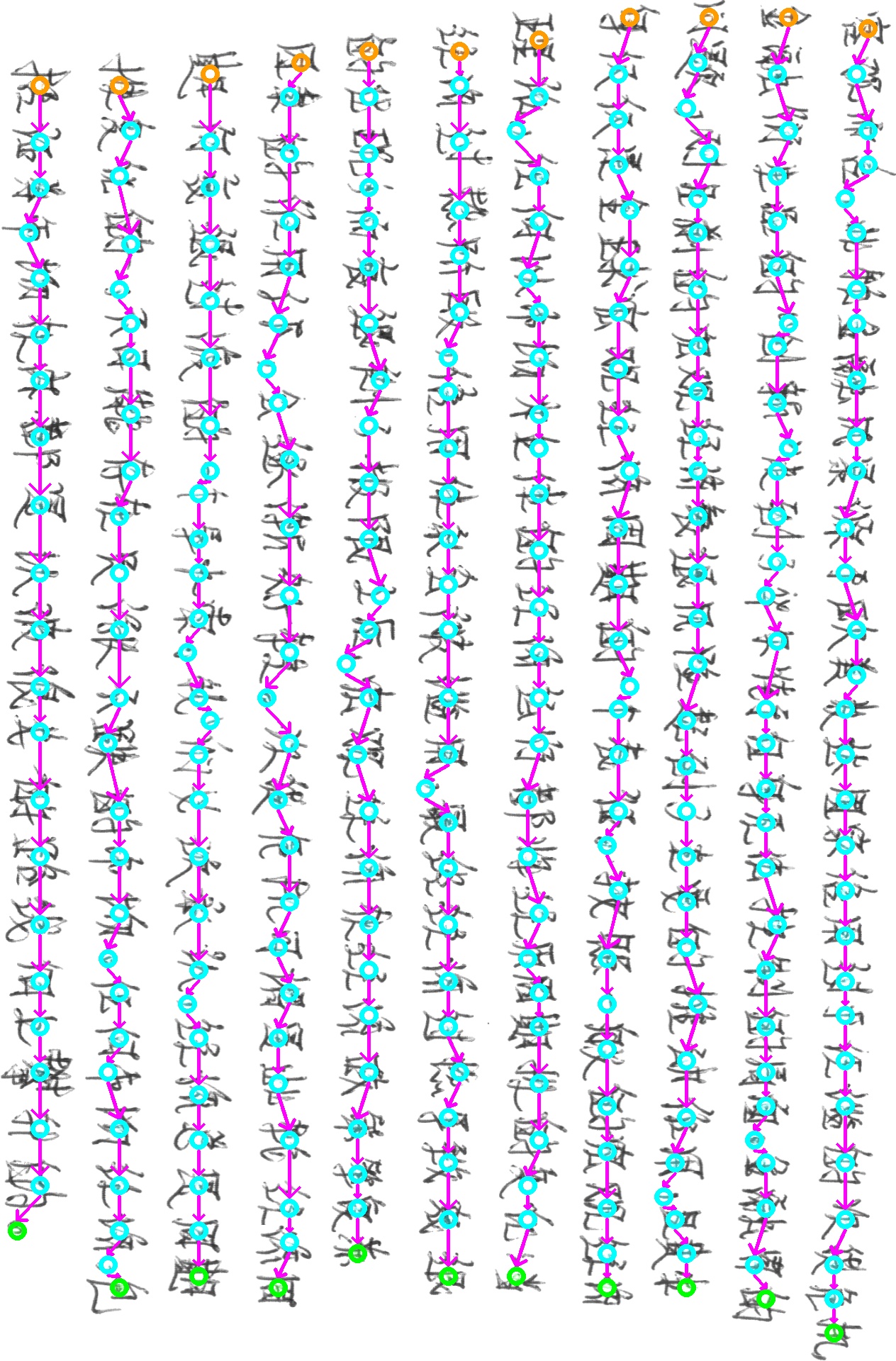}
	\end{minipage}}%
	\subfigure[270\degree]{
		\begin{minipage}[t]{0.5\columnwidth}
			\centering
			\includegraphics[width=0.56\columnwidth]{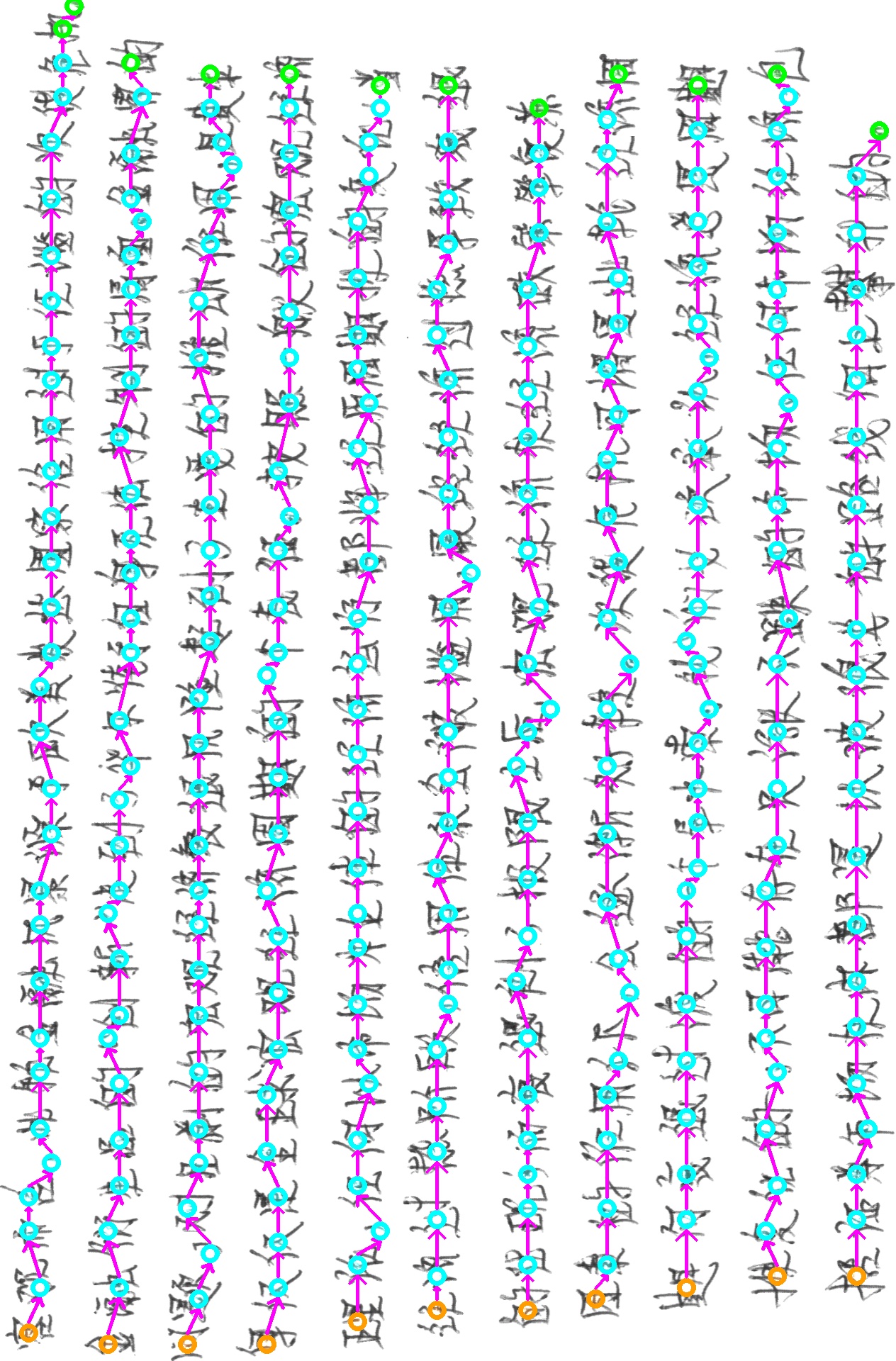}
	\end{minipage}}%
	{\bl \caption{Reading orders predicted by PageNet of the same image rotated by different degrees{\bl , where each circle represents a character (orange circle: start-of-line; green circle: end-of-line).}}\label{Fig_Exp_MultiDir}
	}
	
\end{figure}

For further verification, a multi-directional reading order experiment is conducted on ICDAR13. All the training and testing images are rotated clockwise by 90\degree, 180\degree, and 270\degree, but the reading order in the annotations remains unchanged. Using the original data and three rotated versions, we train and evaluate two models that are under full and weak supervision, respectively. {\bl Note that each of the two models is trained for all the four directions.}

The experimental results are presented in Table \ref{TBL_MDT}. The performance of one model on the rotated data is comparable to that on the unrotated data, and the total performance still maintains a high accuracy. The predicted reading orders of the same image rotated by different angles are visualized in Fig. \ref{Fig_Exp_MultiDir}. It can be seen that the reading order is correctly predicted for all directions.

\subsubsection{Curved Text Lines}
\label{sec_curved_text_line}

\begin{figure}[b]
	\centering 
	\subfigure{
		\begin{minipage}[t]{0.5\columnwidth}
			\centering
			\includegraphics[width=1.0\columnwidth]{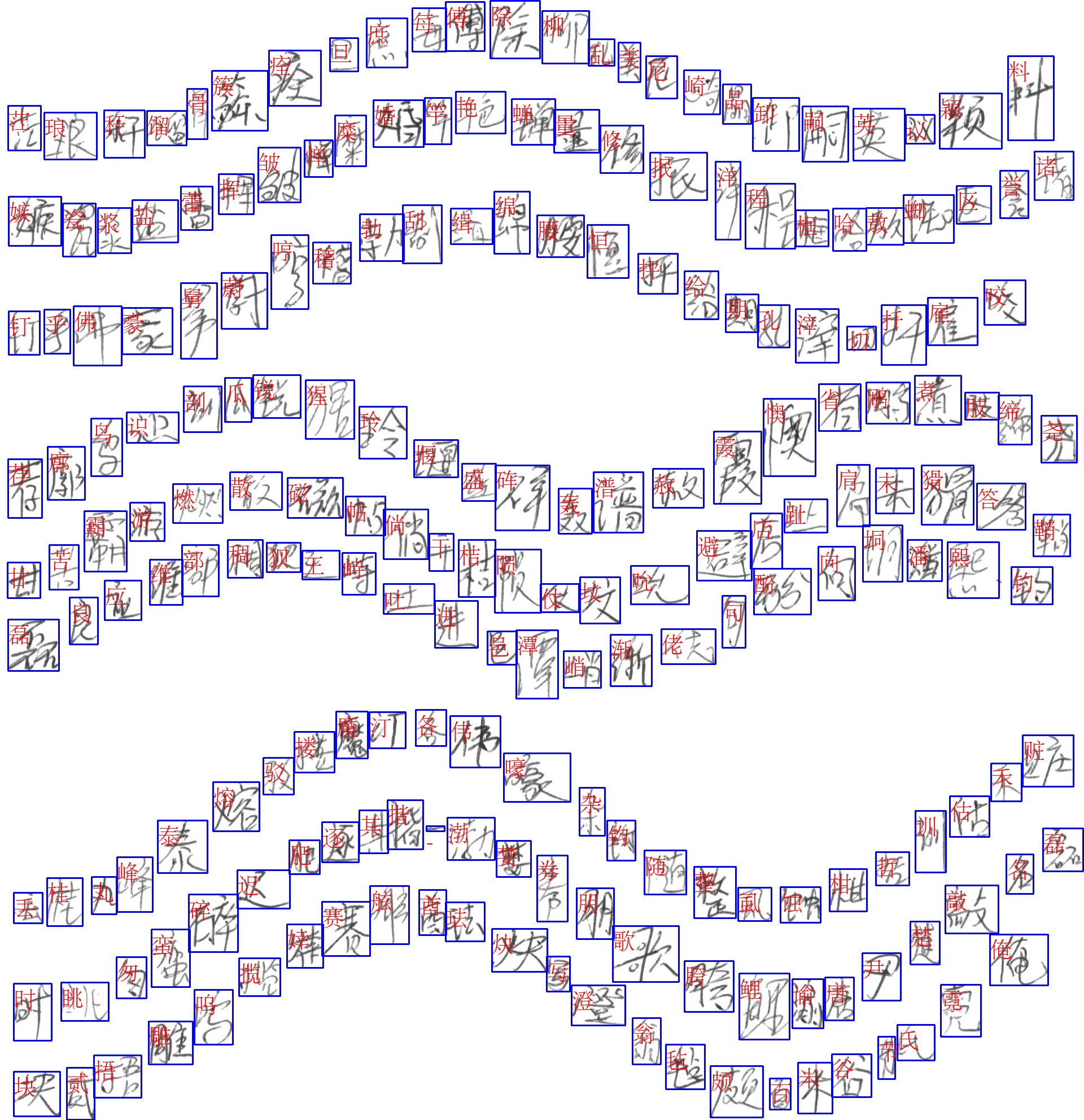}
	\end{minipage}}%
	\subfigure{
		\begin{minipage}[t]{0.5\columnwidth}
			\centering
			\includegraphics[width=1.0\columnwidth]{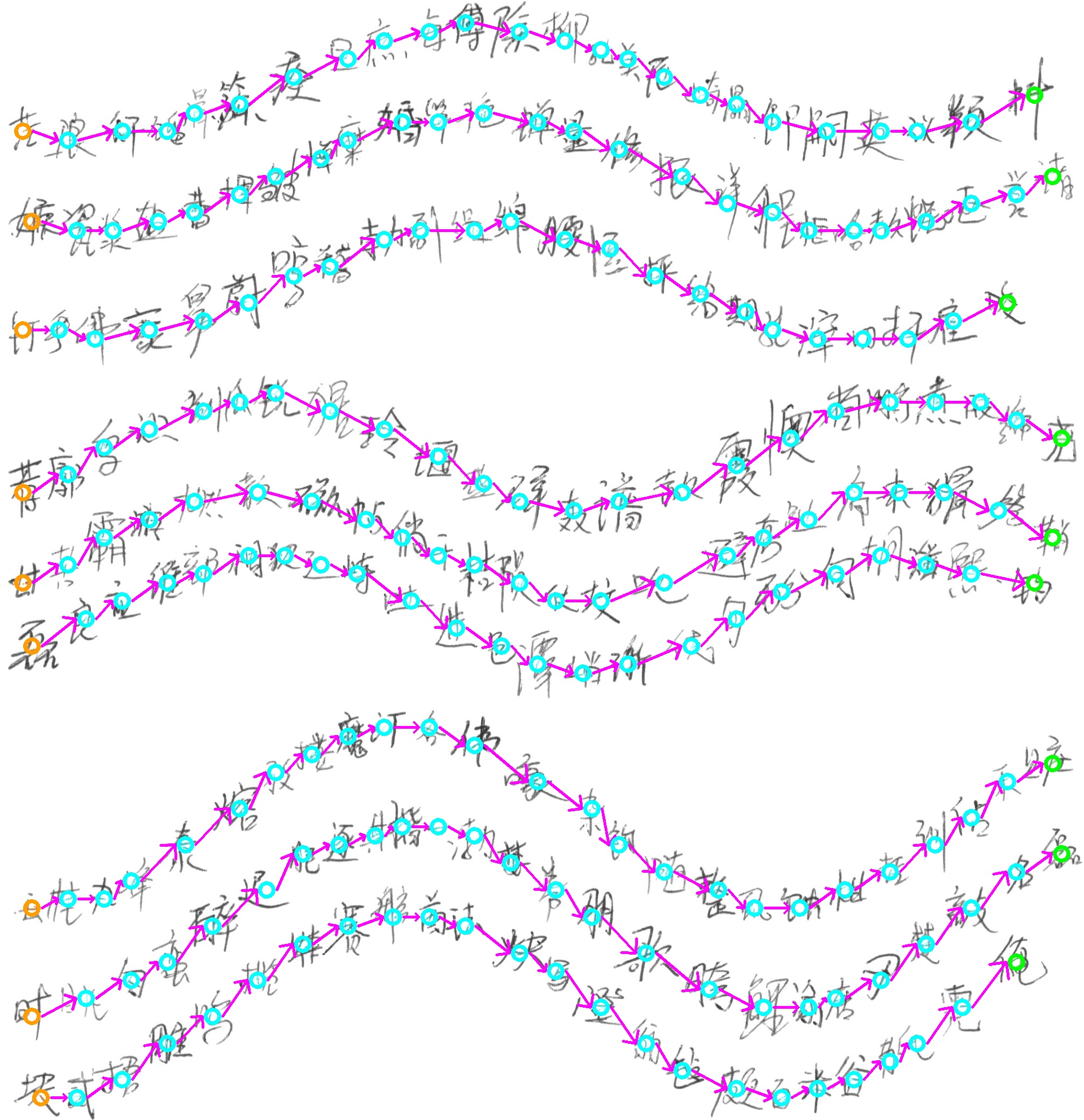}
	\end{minipage}}%
	{\bl \caption{Visualization results of a page containing curved text lines. The two images present character-level detection and recognition results (left) and reading order (right). {\bl In the visualization of reading order, each circle represents a character (orange circle: start-of-line; green circle: end-of-line).}}\label{Fig_Exp_Curved}}
\end{figure}

We also conduct experiments on curved text lines. Both the training and testing sets are synthetic pages containing curved text lines, where the $\mathrm{y}$ coordinate of the character in a text line is a sine function of the $\mathrm{x}$ coordinate. To avoid overfitting,
the training set uses the character samples from CASIA-HWDB1.0-1.2, while the testing set uses the character samples from ICDAR13-SC.
Because the synthetic data has full annotations, the model is fully supervised. The AR* and CR* on the testing set are 94.04\% and 94.36\%, respectively. In addition, Fig. \ref{Fig_Exp_Curved} shows the visualization results. Owing to the bottom-up design and strong reading order predicting mechanism, curved text lines can be effectively recognized by our method.

\subsection{Experiments on Decoding Algorithm}
In Table \ref{TBL_Decoding}, we compare the graph-based decoding algorithm with the rule-based algorithm. The rule-based algorithm groups the characters based on their vertical coordinates and reorders the characters in each group according to their horizontal coordinates.
It can be seen that the proposed graph-based decoding algorithm achieves significantly better performance owing to the effective design of the reading order prediction.
\begin{table}[t]
	\centering 
	\caption{Comparison of different decoding algorithms}
	\label{TBL_Decoding}
	\begin{tabular*}{\hsize}{@{}@{\extracolsep{\fill}}lllll@{}}
		\hline
		\multirow{2}*{Algorithm} & \multicolumn{2}{l}{ICDAR13} & \multicolumn{2}{l}{SCUT-HCCDoc}\\
		\cline{2-3} \cline{4-5}
		& AR* & CR* & AR* & CR* \\
		\hline
		Rule-based & 75.28 & 82.62 & 68.49 & 77.45 \\
		Ours & \textbf{92.83} & \textbf{93.23} & \textbf{77.95} & \textbf{82.15} \\
		\hline
	\end{tabular*}
\end{table}

\subsection{Automatic Labeling}
\label{sec_auto_label}
\begin{figure}[b]
	\centering 
	\subfigure{
		\begin{minipage}[t]{0.5\columnwidth}
			\centering
			\includegraphics[width=0.95\columnwidth]{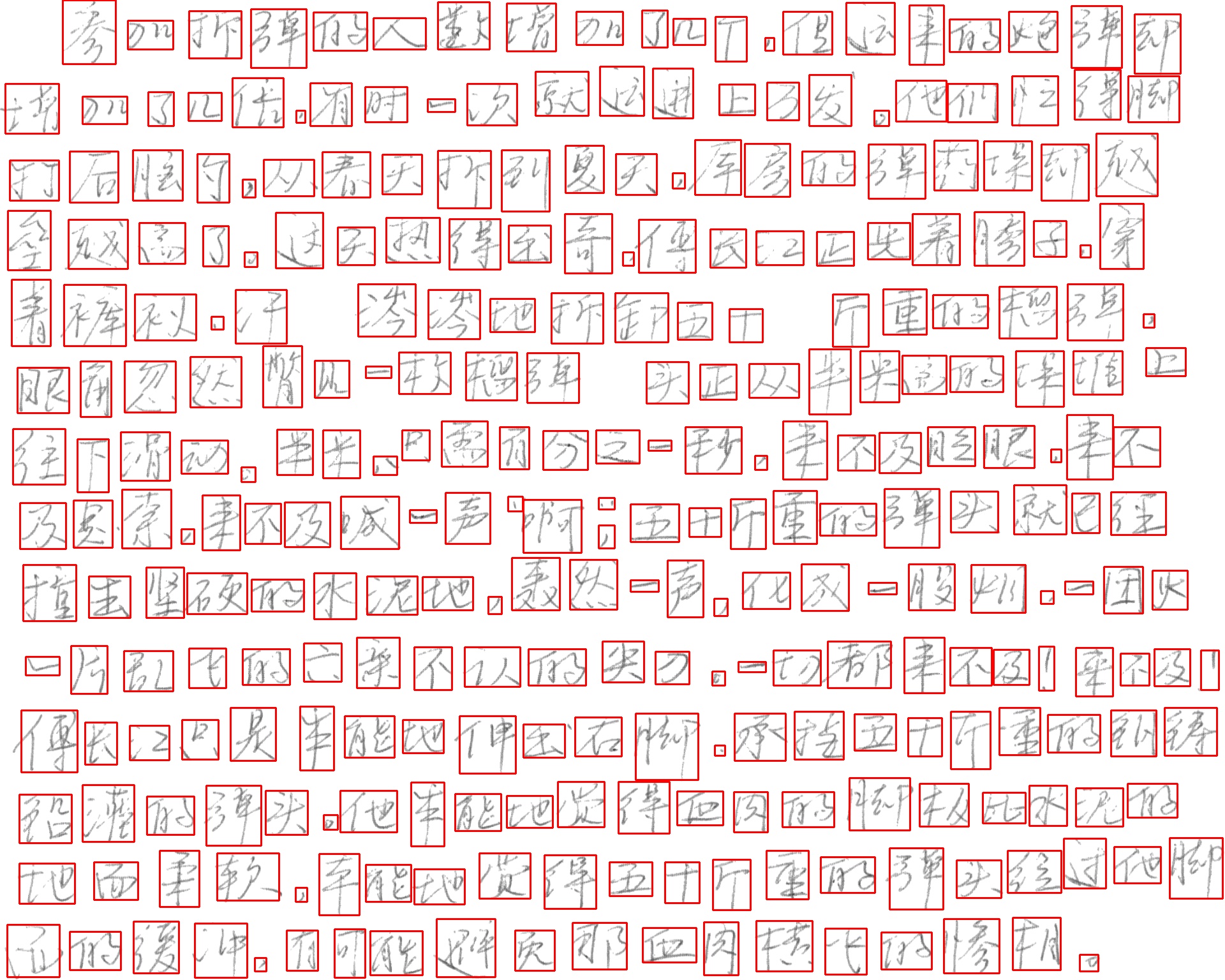}
	\end{minipage}}%
	\subfigure{
		\begin{minipage}[t]{0.5\columnwidth}
			\centering
			\includegraphics[width=0.95\columnwidth]{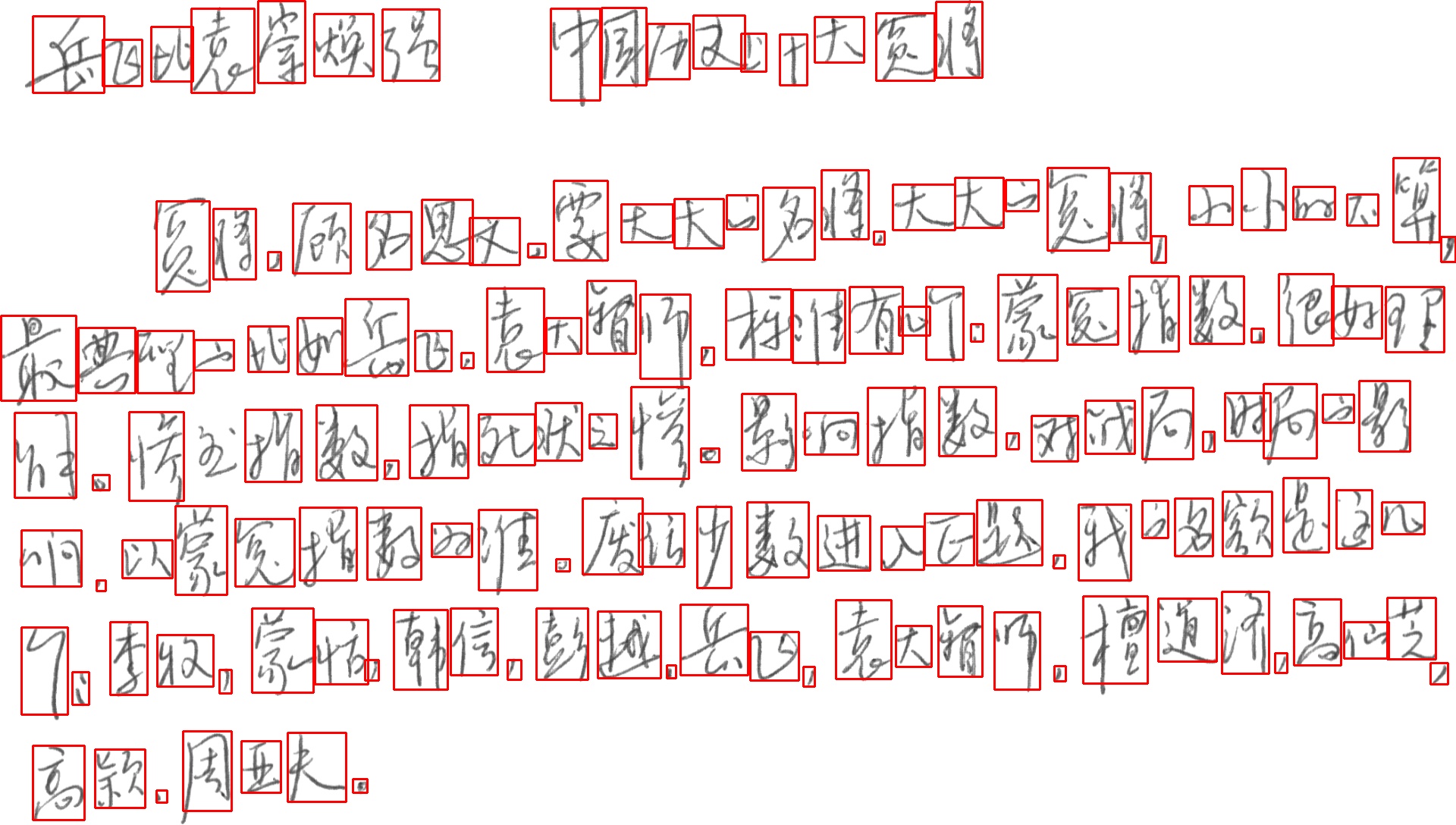}
	\end{minipage}}
	{\bl\caption{Automatically generated annotations for CASIA-HWDB2.0-2.2.}\label{Fig_Exp_AutoLabel_Vis}}
\end{figure}

Another potential application of the proposed weakly supervised learning framework is automatic labeling. Given the line-level transcripts of a page, the bounding boxes of characters can be automatically annotated by the pseudo-labels generated by our method.

Fig. \ref{Fig_Exp_AutoLabel_Vis} illustrates the automatically generated annotations for CASIA-HWDB2.0-2.2. After the training stage, 98.75\% of the characters in CASIA-HWDB2.0-2.2 have corresponding pseudo-labels, and the average IoU between the pseudo-labels and ground-truth bounding boxes is 86.45\%, which indicates the high quality of automatic labeling.

\begin{table}[t]
	\bl
	\centering 
	\caption{Effectiveness of automatically labeled annotations. This table presents the character detection performance on ICDAR13 when Faster R-CNN is trained using different annotations. The performance using original annotations is copied from Table \ref{TBL_CLSR}.}
	\label{tbl_exp_auto_label}
	\begin{tabular*}{\hsize}{@{}@{\extracolsep{\fill}}llll@{}}
		\hline
		Annotations & Precision & Recall & F-measure \\
		\hline
		Original & 98.93 & 92.12 & 95.41\\
		Automatically labeled & 97.70 & 91.78 & 94.64 \\
		\hline	
	\end{tabular*}		
\end{table}

To verify the applicability of automatically generated annotations, we replace the original annotations of CASIA-HWDB2.0-2.2 with automatically labeled bounding boxes. Then, a Faster R-CNN \citep{S_Ren_Faster} is trained to detect characters using CASIA-HWDB2.0-2.2 with new annotations and CASIA-SR, and is tested on ICDAR13. {\bl As shown in Table \ref{tbl_exp_auto_label}, compared with the Faster R-CNN using original annotations, the counterpart using automatically labeled annotations achieves comparable performance on character detection.}

\begin{table}[b]
	\centering 
	\caption{Effects of different synthetic data (evaluated on ICDAR13)}
	\label{TBL_Synthetic_data}
	\begin{threeparttable}
		\begin{tabular*}{\hsize}{@{}@{\extracolsep{\fill}}llll@{}}
			\hline
			Method & Synthetic Data & AR* & CR* \\
			\hline
			\multirow{3}*{Det + Recog} & No synthetic data & 86.27 & 87.37 \\
			& CASIA-SR & 88.36 (2.09$\uparrow$) & 89.09 (1.72$\uparrow$)  \\
			& CASIA-SF & 86.14 (0.13$\downarrow$) & 87.30 (0.07$\downarrow$) \\
			\hline 
			\multirow{3}*{PageNet} & No synthetic data\tnote{1} & 87.03 & 87.63 \\
			& CASIA-SR & 92.83 (5.80$\uparrow$) & 93.23 (5.60$\uparrow$)\\
			& CASIA-SF & 89.56 (2.53$\uparrow$)& 90.52 (2.89$\uparrow$)\\
			\hline
		\end{tabular*}
		\begin{tablenotes}
			\scriptsize\item[1] PageNet without synthetic data is trained under full supervision.
		\end{tablenotes}
	\end{threeparttable}
\end{table}

\subsection{Effects of Synthetic Data}
\label{sec_effect_synthetic_data}

Recently, in order to improve the performance of text detection and recognition, many methods \citep{M_Jaderberg_Synthetic,A_Gupta_Synthetic,F_Zhan_Verisimilar} have been proposed for data synthesis. However, taking advantage of the proposed weakly supervised learning framework, the performance of our method does not rely heavily on the quality of synthetic data. For our method, the synthetic data are synthesized following a simple and unified procedure for all real datasets, which greatly reduces the labor required to design specific synthesis methods for different scenarios.

\begin{table*}[b]
	\centering 
	\caption{Improvement of the final model compared with the pretrained model}
	\label{TBL_Improvement}
	\begin{tabular*}{\hsize}{@{}@{\extracolsep{\fill}}lllllllllll@{}}
		\hline 
		\multirow{3}*{Model} & \multicolumn{4}{l}{ICDAR13} & \multicolumn{2}{l}{\multirow{2}*{MTHv2}} &  \multicolumn{2}{l}{\multirow{2}*{SCUT-HCCDoc}} & \multicolumn{2}{l}{\multirow{2}*{JS-SCUT PrintCC}} \\
		\cline{2-5}
		& \multicolumn{2}{l}{CASIA-SR} & \multicolumn{2}{l}{CASIA-SF}  & & & & & & \\
		\cline{2-3} \cline{4-5} \cline{6-7} \cline{8-9} \cline{10-11}
		& AR* & CR* & AR* & CR* & AR* & CR* & AR* & CR* & AR* & CR* \\
		\hline 
		Pretrained Model & 63.62 & 64.00 & 38.70 & 39.01 & 54.30 & 61.61 & 29.93 & 32.94 & 75.99 & 79.51 \\
		Final Model & 92.83 & 93.23 & 89.56 & 90.52 & 93.76 & 95.23 & 77.95 & 82.15 & 97.25 & 98.19 \\
		Improvement & 45.92\% & 45.67\% & 131.42\% & 132.04\% & 72.67\% & 54.57\% & 160.44\% & 149.39\%& 27.98\% & 23.49\%\\
		\hline
	\end{tabular*}
\end{table*}

We analyze the effects of different synthetic data in Table \ref{TBL_Synthetic_data}. Compared with CASIA-SR, CASIA-SF is synthesized using character samples from font files rather than real character samples. Obviously, CASIA-SR is more similar to the real data than CASIA-SF. Compared with the performance without synthetic data, Det + Recog performs better using CASIA-SR but worse using CASIA-SF. However, PageNet with CASIA-SR and CASIA-SF both achieve significantly better results than the one without synthetic data. 

Furthermore, in Table \ref{TBL_Improvement}, we compare the performances of the pretrained models using synthetic data and the final models after the training stage. Despite the performances of the pretrained models, the final results are greatly improved by the weakly supervised learning.

Based on the above results, we can conclude that the proposed weakly supervised learning framework can effectively learn usable information from synthetic data and adapt to different scenarios, which makes it less dependent on the quality of synthetic data.

{\bl\subsection{Discussion}
\label{sec_discussion}
Generalization ability is an important issue in real-world applications. In the following, we discuss the generalization ability of our method based on the above methodology and experimental results. 

As described in Sec. \ref{Sec_Model}, our method is formulated in a general manner without using prior knowledge of any specific dataset. Each component of our method is designed based on the general properties of the documents rather than considering only specific scenarios.

Extensive experiments are conducted using five datasets that cover most document scenarios. Specifically, CASIA-HWDB2.0-2.2 and ICDAR13 contain scanned documents with cursive handwritten characters and diverse writing styles, MTHv2 contains historical documents with severe degradation, SCUT-HCCDoc contains camera-captured handwritten documents with various illuminations, perspectives, and backgrounds, and JS-SCUT PrintCC contains multilingual printed documents including English and Chinese. The experimental results (Tables \ref{TBL_ICDAR13}, \ref{TBL_Other_Page}, and \ref{TBL_ICDAR13_Line}) and visualizations (Fig. \ref{Fig_Exp_Vis}) demonstrate that our method achieves promising performance on all these datasets.

Additional experiments in Sec. \ref{sec_reading_order} verify the effectiveness of our method on multi-directional reading order and arbitrarily curved text lines. Furthermore, the experiments in Sec. \ref{sec_effect_synthetic_data} demonstrate that our method is less dependent on the quality of synthetic data. Although all synthetic samples are synthesized in a very simple manner as described in Sec. \ref{Sec_Dataset} instead of using advanced synthesis approaches that are specifically designed, our method can work well on all the benchmark datasets. 

Therefore, it may be safe to say that our method can be easily generalized to different scenarios. Nevertheless, the method in this paper is mainly for document-based text recognition. It may be less effective for other complex scenarios such as end-to-end scene text recognition, which is still an open problem that deserves further study.
}

\section{Conclusion}
\label{Sec_Conclusion}

In this paper, we propose PageNet for solving end-to-end weakly supervised page-level handwritten Chinese text recognition from a new perspective. With only line-level transcripts annotated {\bl for real data}, PageNet is able to end-to-end predict detection and recognition results at both the character and line levels, as well as the important reading order of each text line. Extensive experiments on five datasets, including CASIA-HWDB, ICDAR2013, MTHv2, SCUT-HCCDoc, and JS-SCUT PrintCC, demonstrate that our method can achieve state-of-the-art performance, even when compared with fully supervised methods. We further show that the proposed PageNet can surpass the line-level methods of handwritten Chinese text recognition which directly recognize the pre-supplied cropped text line images. It is worth mentioning that our method can serve as an automatic annotator that can produce highly accurate character-level bounding boxes. As there are thousands of web images with only transcript labels on the Internet, the powerful generalization ability of PageNet exhibits its promising potential in real-world applications. We hope that this work opens up new possibilities for end-to-end weakly supervised page-level text recognition.

\section*{Acknowledgement}
This research is supported in part by NSFC (Grant No.: 61936003), GD-NSF (no.2017A030312006, No.2021A1515 \ 011870), and the Science and Technology Foundation of Guangzhou Huangpu Development District (Grant 2020GH17).
%
%


\bibliographystyle{spbasic}


\end{document}